\documentclass[preprint,12pt]{elsarticle}

\usepackage{amssymb}
\usepackage{amsmath}

\usepackage{lineno}
\usepackage{natbib} 
\usepackage{booktabs}       
\usepackage{amsfonts}       
\usepackage{nicefrac}       
\usepackage{microtype}      
\usepackage{xcolor}         
\usepackage{enumitem}
\usepackage{graphicx}
\usepackage{caption}
\usepackage{subcaption}
\usepackage{multirow}
\usepackage{gensymb}
\usepackage{rotating}

\usepackage{mathtools}

\usepackage{algorithm}
\usepackage{algpseudocode}
\newcommand{\HL}[1]{\textcolor{blue}{{#1}}}
\algnewcommand{\LineComment}[1]{\State \(\triangleright\) #1}
\makeatletter
\newenvironment{breakablealgorithm}
  {
     \refstepcounter{algorithm}
     \hrule height.8pt depth0pt \kern2pt
     \renewcommand{\caption}[2][\relax]{
      {\raggedright\textbf{\ALG@name~\thealgorithm} ##2\par}%
      \ifx\relax##1\relax 
         \addcontentsline{loa}{algorithm}{\protect\numberline{\thealgorithm}##2}%
      \else 
         \addcontentsline{loa}{algorithm}{\protect\numberline{\thealgorithm}##1}%
      \fi
      \kern2pt\hrule\kern2pt
     }
  }{
     \kern2pt\hrule\relax
  }
\makeatother

\newcommand{\revise}[1]{\textcolor{black}{{#1}}}

\newcommand{\pjn}{{DGBA}}

\journal{Neurocomputing}

\begin{document}

\begin{frontmatter}



\title{Attacking All Tasks at Once Using Adversarial Examples in Multi-Task Learning}


\author[umass,amazon]{Lijun Zhang} 
\author[umass]{Xiao Liu}
\author[uri]{Kaleel Mahmood}
\author[umn]{Caiwen Ding}
\author[umass,amazon]{Hui Guan}

\affiliation[umass]{organization={University of Massachusetts Amherst}, 
            city={Amherst},
            postcode={01003}, 
            state={MA},
            country={USA}}
\affiliation[amazon]{organization={Amazon}, 
            city={Seattle},
            postcode={98109}, 
            state={WA},
            country={USA}}
\affiliation[uri]{organization={ University of Rhode Island}, 
            city={Kingston},
            postcode={02881}, 
            state={RI},
            country={USA}}
\affiliation[umn]{organization={University of Minnesota - Twin Cities}, 
            city={Minneapolis},
            postcode={55455}, 
            state={MN},
            country={USA}}


\begin{abstract}
Visual content understanding frequently relies on multi-task models to extract robust representations of a single visual input for multiple downstream tasks. 
However, in comparison to extensively studied single-task models, the adversarial robustness of multi-task models has received significantly less attention and many questions remain unclear: 
1) How robust are multi-task models to single task adversarial attacks, 
2) Can adversarial attacks be designed to simultaneously attack all tasks in a multi-task model, 
and 3) How does parameter sharing across tasks affect multi-task model robustness to adversarial attacks?
This paper aims to answer these questions through careful analysis and rigorous experimentation. 
First, we analyze the inherent drawbacks of two commonly-used adaptations of single-task white-box attacks in attacking multi-task models. 
We then propose a novel attack framework, Dynamic Gradient Balancing Attack (\pjn{}). 
Our framework poses the problem of attacking all tasks in a multi-task model as an optimization problem that can be efficiently solved through integer linear programming.
Extensive evaluation on two popular MTL benchmarks, NYUv2 and Tiny-Taxonomy, demonstrates the effectiveness of \pjn{} compared to baselines in attacking both clean and adversarially trained multi-task models. 
Our results also reveal a fundamental trade-off between improving task accuracy via parameter sharing across tasks and undermining model robustness due to increased attack transferability from parameter sharing.
\end{abstract}


\begin{highlights}
\item Existing attempts on attacking multi-task models presents inherent drawbacks.
\item Formulate the MTL adversarial attack as an optimization problem. 
\item Develop DGBA to efficiently solve the multi-task attack problem.
\end{highlights}


\end{frontmatter}


\section{Introduction}
\label{sect:intro}
Visual content understanding often employs multi-task learning (MTL) to tackle diverse downstream tasks such as image segmentation and depth estimation. 
MTL is favored for its high resource efficiency and task accuracy~\cite{ruder2017overview, zhang2022automtl, zhang2022tree, yao2020survey}. 
It involves constructing a unified \textit{multi-task model} that extracts feature representations from visual inputs by enabling \textit{parameter sharing} across tasks. 
%
As multi-task models are an inseparable component in many applications with high-security requirements, such as autonomous driving and robotics~\cite{leang2020dynamic,kokkinos2017ubernet,arcari2023bayesian}, it is important to understand and assess their {adversarial robustness} to ensure the effective adoption in real applications. 


Machine learning models are susceptible to adversarial attacks where small, imperceptible perturbations to input data can lead to incorrect prediction~\cite{goodfellow2014explaining}. 
Although adversarial attacks have been extensively studied on single-task models~\cite{Delving, BARZ, xu2022securing}, related work on multi-task models is scarce. 
A pioneering study~\cite{mao2020multitask} claimed that the adversarial robustness of deep neural networks (DNN) increases as the number of tasks increases when attacking the sum of the tasks' loss, but its conclusion is overturned by a following study~\cite{ghamizi2022adversarial}. 
Multi-Task Attack~\cite{guo2020multi} tries to develop attacks in the MTL setting; however, the adversarial samples generated are specific to each task and thus fail to attack all tasks simultaneously.
\revise{Some other related works~\cite{gurulingan2021uninet,sobh2021adversarial} start from generating adversarial examples by attacking one task at a time, which does not consider the multiple tasks simultaneously. 
UniNet \cite{gurulingan2021uninet} goes one step further by evaluating a straightforward multi-task attack that targets the sum of all task losses; yet this naïve strategy still mainly degrades subset of tasks, leaving others comparatively robust.}
Motivated by these limitations, we systematically investigate two critical security research questions (\textbf{RQ}) for MTL in this work: 
\setlist{nolistsep}
\begin{itemize}[noitemsep]
\item \textbf{RQ1}: \textit{How robust are multi-task models to conventional single task adversarial attacks?} 
\item \textbf{RQ2}: \textit{Can adversarial attacks be designed to simultaneously attack all tasks in a multi-task model?}

\end{itemize} 

To answer \textbf{RQ1}, we first categorize existing attempts on attacking multi-task models into two strategies, both relying on adapting existing single-task white-box attack approaches to multi-task models. 
We refer to these existing approaches as ``\textit{adapted multi-task attacks}'' throughout the paper.
We then analyze their inherent drawbacks that limit their attack effectiveness on multi-task models and motivate our proposed approach. (Section~\ref{sec:background})

To answer \textbf{RQ2}, we propose a novel multi-task attack method, \pjn{} (\underline{D}ynamic \underline{G}radient \underline{B}alancing Multi-task \underline{A}ttack) to generate adversarial samples effective in attacking all tasks at once in a multi-task model. 
The central principle of \pjn{} is to harness established single-task attack techniques, thus avoiding redundant efforts, and at the same time, overcome the inherent limitations of existing adaptation strategies, thereby generating effective adversarial samples. 
The key insight is to dynamically balance gradients from multiple tasks in a multi-task model when creating adversarial samples that work across all tasks. 
(Section~\ref{sect:DGBA})

We conduct thorough experiments using common MTL benchmarks and branched multi-task models with different levels of parameter sharing. 
Experiments demonstrate that \pjn{} achieves up to 80.41\% higher attack effectiveness on clean multi-task models and up to 18.65\% higher attack effectiveness on adversarially trained multi-task models compared to baselines. \pjn{} remains the most effective attack approach in most cases regardless of the levels of parameter sharing and the attack strengths. (Section~\ref{sect:evaluation})  

\revise{Our results in Section~\ref{sect:evaluation} also demonstrate that a higher degree of parameter sharing is associated with increased adversarial vulnerability of multi-task models, from which we raise the third research question \textbf{RQ3}: \textit{How does parameter sharing across tasks affect multi-task model robustness to adversarial attacks?}.}
To answer \textbf{RQ3}, we empirically study the reason behind the tradeoff of parameter sharing and model robustness and find that 
improving a task's accuracy and efficiency by sharing its parameters with other related tasks can increase the task's vulnerability to adversarial attacks designated for these related tasks.  
This underlines the importance of balancing accuracy and robustness in multi-task model design.
(Section~\ref{sect:exp-sharing}) 

We summarize our contributions as follows. 
\setlist{nolistsep}
\begin{itemize}
    \item We categorize existing attempts on attacking multi-task models and analyze their inherent drawbacks. 
    
    \item 
    We formulate the MTL adversarial attack as an optimization problem and develop \pjn{} to efficiently solve it. 
    
    
    \item We empirically evaluate the effectiveness of \pjn{} on multi-task models with various levels of parameter sharing and demonstrate that \pjn{} performs best for 7 out of 8 models on NYUv2~\cite{silberman2012indoor} and 6 out of 8 on Tiny-Taskonomy~\cite{zamir2018taskonomy}.  

        \item 
    We incorporate adversarial examples into the training of multi-task models to defend against adversarial attacks. 
    The robustness of the models improves markedly, measured by the decreased task performance drop after attacking from $46.65-105.74\%$ to $5.97-29.26\%$. 
    When attacking these adversarially trained models, \pjn{} still outperforms baselines by up to $18.65\%$.
    
    \item 
    We empirically demonstrate that parameter sharing can undermine model robustness due to increased attack transferability.

\end{itemize}

\section{Background and Motivation}
\label{sec:background}
\revise{\textbf{RQ1}: \textit{How robust are multi-task models to conventional single task adversarial attacks?}
To answer this question, this section first gives background on existing white-box single-task attacks and our adversarial threat model. 
It then summarizes existing attempts on adapting single-task attacks to multi-task models and analyzes their drawbacks that motivate our proposed method \pjn{}. A detailed discussion of related work is in~\ref{sect:related}.}


\subsection{Single Task White-Box Attacks}
In general, adversarial attacks can be formulated as follows~\cite{madry2018towards}.
Let $(x,y)$ represent a clean input and its corresponding label. An attacker adds an adversarial perturbation $\delta$ to the input $x$, to maximize the value of a loss function $\mathcal{L}$:
\begin{equation}\label{equ:op_ori}
    \max_\delta \mathcal{L}(x+\delta,y;\theta), \quad s.t. \quad \lVert \delta \rVert_p \leq \epsilon,
\end{equation}
where $\theta$ denotes the parameters of the trained model under attack, and $\epsilon$ represents the maximum amount the adversary can perturb the input according to a given $p$-norm. For notational simplicity, we omit $\theta$ in our future derivations.
\revise{Note that, although the input $x$ and the perturbation $\eta$ are matrices in vision tasks (defined at the pixel level), we keep matrix-valued variables in italic following the prevailing convention in adversarial-learning literature~\cite{croce2020reliable,xu2022securing}. A complete notation table is provided in~\ref{sect:notation}.}

\textbf{Threat Model}: In this paper, we focus on the untargeted white-box adversarial threat model~\cite{carlini2019evaluating} as this represents one of the strongest and most widely used adversarial machine learning formulations~\cite{carlini2017towards, dong2018boosting, croce2020reliable}. In this setup, the attacker has knowledge of the model structure, trained model parameters $\theta$ and the corresponding loss function $\mathcal{L}$. In terms of bounds on the adversarial perturbation, we use one of the most widely used norms, $p=\infty$, in line with previous works~\cite{guo2020multi,xu2022securing,rathbun2022game}.

\textbf{Single Task Attacks}: In the white-box setting, one of the most prevalent strategies for generating the adversarial perturbation $\delta$ is to maximize the loss function $\mathcal{L}$ by following the gradient ascent direction. This was originally done with the Fast Gradient Sign Method (FGSM) attack proposed in~\cite{ goodfellow2014explaining}. Since the advent of FGSM, numerous improvements to the attack have been proposed. Although enumerating all the improvements in FGSM is beyond the scope of this paper, several important attack updates are worth noting. Updated attacks include the Projected Gradient Descent (PGD) attack~\cite{madry2018towards}, which adds a randomized start and makes FGSM iterative. The Momentum Iterative Momentum (MIM)~ \cite{dong2018boosting} adds momentum to the gradient ascent optimization. More recently, in APGD~\cite{croce2020reliable}, an adaptive step size has been shown to be one of the most effective white-box attacks for a single task, even against adversarial trained models~\cite{mahmood2021robustness}.

\subsection{Adapting Single-Task Attacks to MTL}\label{sect:naive}
We first categorize existing methods on attacking multi-task models into two strategies, \textsc{Single} attack and \textsc{Total} attack. 
Both strategies rely on adapting single-task white-box attacks to multi-task models and thus are referred to as \textit{adapted multi-task attacks} in this paper. 
We then analyze and empirically demonstrate their inherent flaws, which motivate the proposed multi-task attack method \pjn{} in Section~\ref{sect:DGBA}. 

 In the case of a single task, untargeted white-box attack, an adversarial example can be generated~\cite{madry2018towards} iteratively:
\begin{equation}
\label{equ:main}
      x_{adv}^{(k)} = \mathcal{P}_{\mathcal{S}}( x^{(k-1)}_{adv} + F_{\delta}(\epsilon^{(k-1)}, \frac{\partial \mathcal{L}}{\partial x_{adv}^{(k-1)}})),
\end{equation}
where $k$ is the current step, $F_{\delta}$ represents the perturbation function associated with a white-box attack, $\mathcal{L}$ represents the loss function for a single task, $\epsilon^{(k-1)}$ is the magnitude of the perturbation added in the current iteration of the attack and $x_{adv}^{(0)}=x$. 
Lastly, $\mathcal{P}_{S}$ is the projection operation~\cite{croce2020reliable} to bound the adversarial sample within a specified range. 

In MTL, each input $x$ is associated with a set of true labels $\{y_1,\dots,y_n\}$ for tasks $\mathcal{T}=\{t_1,\dots,t_n\}$. Each task $t_i$ has its own task-specific loss function $\mathcal{L}_i(x,y_i)$. 


\textbf{\textsc{Single} Attack} - The first way in which multi-task models can be attacked is by
focusing on only a single task's gradient and ignoring the gradients of the rest tasks. 
This approach is explored in~\cite{sobh2021adversarial, gurulingan2021uninet,li2023sibling}. 
For example, APGD~\cite{croce2020reliable} can be utilized to attack one of the task $t_{i}$ in the task set $\mathcal{T}$:

\begin{equation}
    \begin{aligned}
    x_{adv}^{(k)} = \mathcal{P}_{\mathcal{S}}(&x_{adv}^{(k-1)} \\
    &+\alpha(\mathcal{P}_{\mathcal{S}}(x_{adv}^{(k-1)}+\epsilon^{(k-1)}\text{sign}(\frac{\partial \mathcal{L}_{i}}{x_{adv}^{(k-1)}}))-x_{adv}^{(k-1)}) \\
    &+(1-\alpha)(x_{adv}^{(k-1)}-x_{adv}^{(k-2)})),
    \end{aligned}
\end{equation}
where $\alpha$ is a hyperparameter in APGD that controls the influence of previous update steps on the current update step. 
$\mathcal{L}_{i}$ is the objective function of the task $t_i$ being attacked.
We use \textsc{Single-x} to represent performing \textsc{Single} attack on a specific task \textsc{x}.

\textbf{\textsc{Total} Attack} - The second way in which single task attacks can be converted to attack multi-task models is through totaling all associated task loss functions $\mathcal{L}_i$, via summation. This approach is used in \cite{mao2020multitask,gurulingan2021uninet,ghamizi2022adversarial}. For example, single-task PGD~\cite{madry2018towards} can be adapted to \textsc{Total}-PGD: 
\begin{equation}
\label{equ:totalpgd}
      x_{adv}^{(k)} = \mathcal{P}_{\mathcal{S}}( x^{(k-1)}_{adv} + \epsilon^{(k-1)}\cdot\text{sign}(\sum_{i=1}^{n} \frac{\partial \mathcal{L}_{i}}{\partial x_{adv}^{(k-1)}})). 
\end{equation}
We show the different attack formulations for \textsc{Single-x} with APGD and \textsc{Total} with PGD, but it is important to note that any combination of existing white-box attacks and adaptations can be made.

\begin{figure}
  \centering
    \includegraphics[width=.8\linewidth]{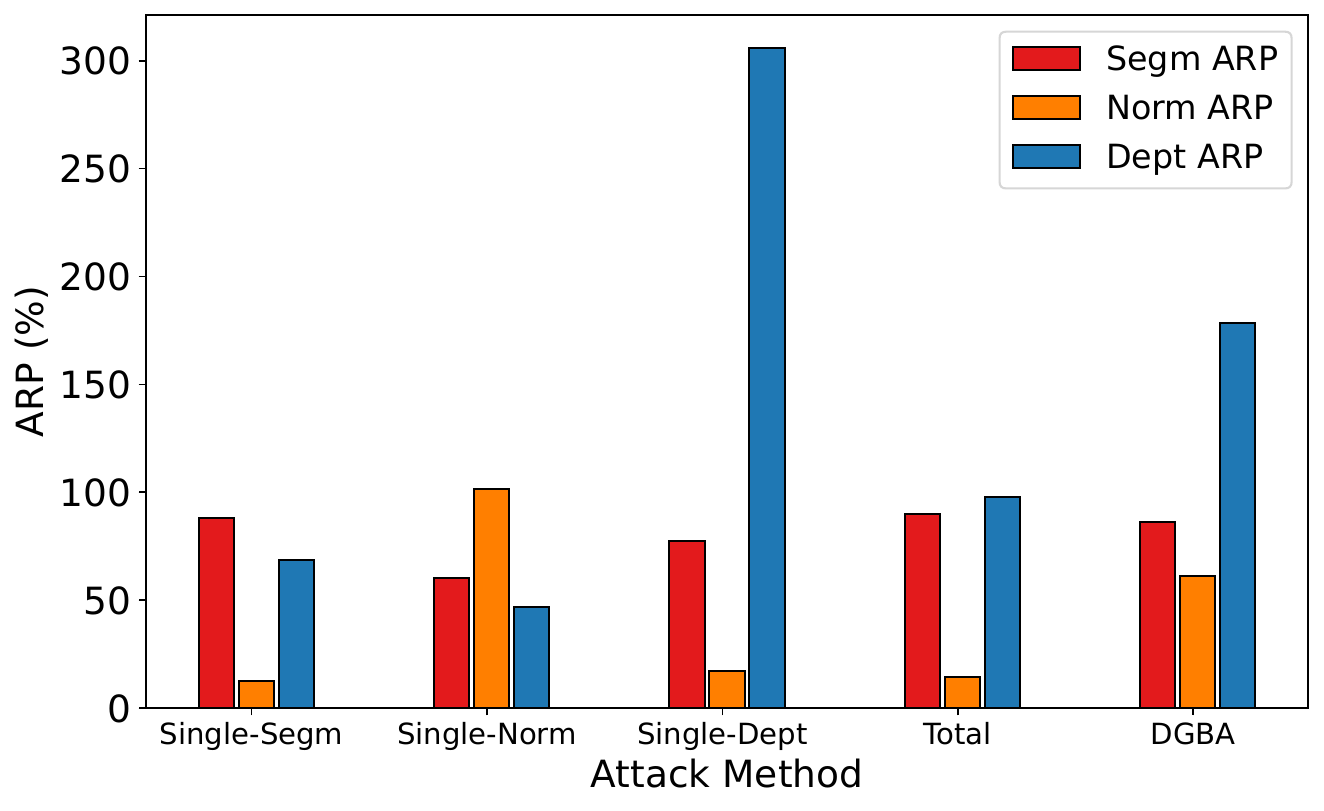}
    \vspace{-5pt}
    \caption{Attack effectiveness in terms of Average Relative Performance (ARP) as defined in Eq.\ref{equ:arp} (y-axis, higher-the-better) for each task when applying \textsc{Single}, \textsc{Total}, and the proposed \pjn{} attacks on NYUv2 (a three-task dataset). ARP is a generic metric that reflects how much the task performance has degraded after an attack regardless of the diverse metrics used in different tasks. The attack variants are built on APGD. Segm: semantic segmentation; Norm: normal prediction; Dept: depth estimation.}\label{fig:transfer}
    \vspace{-14pt}
\end{figure}

\textbf{Limitations of Existing Attacks} - 
Both the \textsc{Single} and \textsc{Total} attacks come with significant drawbacks.
\revise{\textsc{Single} attack is designed to compromise one target task. It may incidentally affect the other tasks through adversarial transferability, but this multi-task effect is typically limited~\cite{mahmood2021robustness}.}
For example, in Figure \ref{fig:transfer}, we show that when attacking the segmentation or the depth estimation task solely (\textsc{Single-Segm} or \textsc{Single-Depth}), the attack effectiveness on the normal prediction task is limited. 

Likewise, the effectiveness of the \textsc{Total} attack is based on an underlying assumption: 
\textit{for an adversarial example to work across all tasks, no one task's gradient should dominate to avoid the limited attack transferability issue that \textsc{Single} faces.}
We denote it as \textit{the non-dominant magnitude} assumption.
However, the issue of gradient dominance is well recognized within the MTL literature and has garnered significant attention in the field of MTL optimizers~\cite{yu2020gradient,navon2022multi}.
Empirically, we observe from Figure \ref{fig:transfer} that the \textsc{Total} attack exhibits a pattern similar to \textsc{Single-Segm}, indicating that the segmentation task dominates the gradient directions. 

We also consider a modified version of \textsc{Total} where we take the sign of the gradients before the summation. In this way, the non-dominant magnitude assumption can be circumvented. We denote this attack as \textsc{SignTotal}. However, we empirically show that this attack is also not effective in Section \ref{sect:exp-overall}, as \textit{completely ignoring task's gradient magnitudes also leads to a suboptimal attack}. 

The limitations due to the underlying assumptions of the \textsc{Total} and \textsc{Single} attacks mandate the need for an attack method tailored to multi-task models. 
The adversarial samples constructed from \textsc{Single} attacks are task-specific, and thus are not effective on non-targeted tasks. 
On the other hand, although the adversarial samples created from \textsc{Total} attack are task-agnostic, they are effective on only the tasks whose gradients dominate in MTL. 
An effective multi-task attack should be able to generate task-agnostic adversarial samples that are effective on all tasks in a multi-task model.

\section{The Proposed Method: \pjn{}}
\label{sect:DGBA}
\revise{\textbf{RQ2}: \textit{Can adversarial attacks be designed to simultaneously attack all tasks in a multi-task model (theoretically)?}
Dynamic Gradient Balancing Multi-task Attack (\pjn{}) builds on the success of existing single-task adversarial attacks, while addressing the challenge in attacking multi-task models. 
\pjn{} accomplishes this by dynamically balancing the gradients across tasks, to derive an adversarial perturbation that is effective on all tasks. 
In this section, we first formulate a new attack optimization problem tailored for multi-task models. Since the problem is intractable,
\pjn{} reformulates it to an Integer Linear Programming (ILP) problem, and then generates adversarial samples by solving the ILP problem.}

\subsection{Multi-Task Attack Optimization} 
We first reformulate the original single-task adversarial optimization introduced in Eq.~\ref{equ:op_ori} by decomposing $\delta = \eta \cdot \beta$. Here, $\eta$ represents the magnitude of the perturbation. $\beta$ represents the signed gradient direction vector, with values $\{-1,0,1\}$. 
\begin{equation}\label{equ:op_ori_beta}
\begin{aligned}
    \max_\beta \, &\mathcal{L}(x+\eta \cdot \beta,y) \\
    s.t. \, \lVert \eta \cdot \beta \rVert_p \leq \epsilon, \quad 
    &\forall \beta^{(k)} \in \beta: \beta^{(k)} \in \{-1,0,1\}.
\end{aligned}
\end{equation}

The above formulation is used to attack a single task. 
However, attacking multiple tasks simultaneously in a multi-task model is fundamentally a multi-objective optimization problem. 
Specifically, since there are multiple tasks and each task has its own objective functions, there is no single metric that can measure the multi-task attack success as the attack success rate~\cite{croce2020reliable} used in single-task attack.
Therefore, we formulate the multi-task attack optimization problem with a multi-task--specific objective function that aligns with the standard practice of assessing model performance in MTL.


A multi-task model's performance is typically measured by Average Relative Accuracy (ARA), denoted as $\Delta Acc$, as opposed to using absolute values~\cite{sun2019adashare,zhang2022automtl}.
The ARA metric compares the performance of a given multi-task model $M$ to that of a baseline model $B$:
\begin{equation}
\label{equ:acc}
    \Delta Acc = \frac{1}{N} \sum_{i=1}^N \frac{Acc_{M,t_i} - Acc_{B,t_i}}{Acc_{B,t_i}},
\end{equation}
where $N$ is the number of tasks. 
$\Delta Acc$ is the average difference in accuracy between $Acc_{M,t_i}$ and $Acc_{B,t_i}$ in all tasks $t_i$, normalized by the accuracy of $B$. 
A higher $\Delta Acc$ indicates better model performance compared to the baseline. 

When attacking a multi-task model, the goal is to
substantially reduce the task performance of $M$ relative to $B$, where now $B$ represents the model before the attack and $M$ denotes the model after the attack.
$\Delta Acc$  will be a negative value, and the higher its absolute value is, the more effective the attack is. 
To find a perturbation direction $\beta$ that is the most effective, we reformulate the objective function in Eq.~\ref{equ:op_ori_beta}: 
\begin{equation} 
\label{equ:obj-intractable}
\begin{aligned}
    \beta^\ast &= \arg\max_\beta |\Delta Acc| =  \arg\max_\beta \Delta \mathcal{L} \\
    &= \arg\max_\beta \frac{1}{N} \sum_{i=1}^n \frac{\mathcal{L}_i(x + \eta \cdot \beta,y_i)-\mathcal{L}_i(x,y_i)}{\mathcal{L}_i(x,y_i)},
\end{aligned}
\end{equation}
where $\mathcal{L}_i(x + \eta \cdot \beta,y_i)-\mathcal{L}_i(x,y_i)$ represents the model loss difference for task $t_i$ before and after the attack to substitute the accuracy difference, since task loss is widely used in MTL as an accuracy proxy, to indicate model quality (i.e., higher task loss corresponds to lower task accuracy) \cite{sun2019adashare,zhang2022automtl,zhang2022tree}.
Besides, considering the loss functions for vision tasks used in our work are Cross-Entropy Loss, Mean Squared Error, or Mean Absolute Error, we simply use $\mathcal{L}_i(x,y_i)$ instead of $|\mathcal{L}_i(x,y_i)|$ in the denominator.

The optimization problem in Eq.~\ref{equ:obj-intractable} can identify the optimal signed gradient direction vector $\beta^\ast$ that is effective on all tasks. 
However, this problem is intractable due to the high-dimensional, non-convex nature of the loss landscape across multiple tasks, where finding a global-optimal solution is computationally prohibitive~\cite{kreinovich1996approximate,horavcek2017interval}. 
Thus, we next explain a relaxed ILP formulation to solve the problem. 

\subsection{Optimization Solution via Relaxed LP}
\label{sec:optimization}
We reformulate Eq.~\ref{equ:obj-intractable} to an ILP problem by applying the Taylor Expansion on $\mathcal{L}_i(x + \eta \cdot \beta,y_i)$ (see~\ref{sect:theory-details} for the detailed derivation):
\begin{equation} 
\label{equ:obj_ILP}
\begin{aligned}
    \beta^\ast = \arg\max_\beta \sum_i \beta \cdot \frac{\partial \mathcal{L}_i(x,y_i)}{\partial x} \cdot \frac{1}{\mathcal{L}_i(x,y_i)} \\
    s.t. \quad \forall \beta^{(k)} \in \beta: \beta^{(k)} \in \{-1,0,1\}.
\end{aligned}
\end{equation}
In practice, $\beta$ and $\frac{\partial\mathcal{L}_i(x,y_i)}{\partial x}$ are two matrices the same size as $x$. Their product represents the dot product of the corresponding vectorized matrices. 

\pjn{} identifies $\beta^*$ by first addressing a relaxed Linear Programming (LP) problem based on Eq.~\ref{equ:obj_ILP} and then rounding the resulting solution to obtain an integer solution. 
In the first step, the LP relaxation will remove the requirement of integer values, i.e. $\forall \beta^{(k)} \in \beta: \beta^{(k)} \in \{-1,0,1\}$, allowing them to be any real value instead. 
To solve the relaxed LP, we calculate the derivative of the objective function with respect to the variable $\beta$ as follows,
\begin{equation} \label{equ:de_obj}
\begin{aligned}
    \frac{\partial \Delta \mathcal{L}}{\partial \beta} = 
    \sum_{i=1}^n \frac{\partial \mathcal{L}_i(x,y_i)}{\partial x} \cdot \frac{1}{\mathcal{L}_i(x,y_i)}.
\end{aligned}
\end{equation}
In other words, $\beta$ starts from the original state, that is, a zero matrix, and is updated along the direction of $\frac{\partial \Delta \mathcal{L}}{\partial \beta}$ to maximize $\Delta \mathcal{L}$ in the LP relaxation. 
Then in the second step, \pjn{} reintroduces the integer constraint, and the solution for the original ILP in Eq.~\ref{equ:obj_ILP} can be obtained by performing a rounding operation: $\beta^\ast = \text{sign}(\beta)$.

The optimal $\beta^\ast$ suggests that \textit{the effective attack direction for multi-task models should be the sum of each task's gradients dynamically weighted by its loss value.} 
\pjn{} mitigates the dominating task issue in \textsc{Total} by dynamically balancing the gradients across tasks and avoids the limited transferability problem in \textsc{Single-x} by optimizing over all tasks simultaneously.

\subsection{Integrating \pjn{} with Existing Attacks} 
\pjn{} can be easily integrated with any existing white-box single-task attack. 
The key operation is to substitute the single task gradient, i.e., $\frac{\partial\mathcal{L}}{\partial x}$, with the balanced multi-task counterpart, i.e., $\sum_{i=1}^n \frac{\partial \mathcal{L}_i(x,y_i)}{\partial x} \cdot \frac{1}{\mathcal{L}_i(x,y_i)}$. 

To illustrate. we provide the pseudocode for APGD integrated with \pjn{} in Algorithm~\ref{alg:APGD-DGBA}.
We color the adaptation made by \pjn{} in blue. 
Notice that on top of the key design shown in lines 2 and 11, we also change the absolute loss value used in the original APGD to the relative loss value sum over all the tasks in lines 4 and 14 to align with the MTL scenario. 
\revise{This integration procedure generalizes to any gradient-based white-box single-task attack, including multi-model attacks such as Auto-SAGA. 
We describe the integration of \pjn{} with Auto-SAGA in~\ref{sect:autosaga-dgba}.}

{\medskip
\small 
\begin{breakablealgorithm}
\label{alg:APGD-DGBA}
    \caption{\pjn{}-APGD}
    \begin{algorithmic}[1]
        \Require $x^{(0)}, \HL{ \{y_i\}, \{\mathcal{L}_i\}}, \eta, \alpha, \mathcal{N}_{iter}, \text{attack checkpoints:} W$
        \Ensure $x_{max}$
        \State $\HL{l_i \leftarrow \mathcal{L}_i(x^{(0)}, y_i), \forall i=1, \cdots, n.}$
        \State $\HL{\beta^\ast \leftarrow \text{sign}(\sum_i^n \frac{\partial \mathcal{L}_i(x^{(0)},y_i)}{\partial x^{(0)}} \cdot \frac{1}{\mathcal{L}_i(x^{(0)},y_i)}) }$ 
        
        \State $x^{(1)} \leftarrow P(x^{(0)}+\eta\cdot \HL{\beta^\ast} )$ 
        \State $l_{max} \leftarrow \HL{\max\{\sum_i^n \frac{\mathcal{L}_i(x^{(0)})-l_i}{l_i}, \sum_i^n \frac{\mathcal{L}_i(x^{(1)})-l_i}{l_i}} \}$
        \If{$l_{max} \equiv \HL{\sum_i^n \frac{\mathcal{L}_i(x^{(0)})-l_i}{l_i}}$}
            \State $x_{max} \leftarrow x^{(0)}$
        \Else
            \State $x_{max} \leftarrow x^{(1)}$
        \EndIf
        \For{$k = 1$ to $\mathcal{N}_{iter}-1$}
            \State $\HL{\beta^\ast \leftarrow \text{sign}(\sum_i^n \frac{\partial \mathcal{L}_i(x^{(k)},y_i)}{\partial x^{(k)}} \cdot \frac{1}{\mathcal{L}_i(x^{(k)},y_i)}) }$ 
            \State $z^{k+1} \leftarrow P(x^{(k)}+\eta\cdot \HL{\beta^\ast} )$
            \State $x^{k+1} \leftarrow P(x^{(k)}+\alpha(z^{k+1}-x^{(k)})+(1-\alpha)(x^{(k)}-x^{(k-1)}))$
            \If{$\HL{\sum_i^n \frac{\mathcal{L}_i(x^{(k+1)})-l_i}{l_i}} > l_{max}$}
                \State $x_{max} \leftarrow x^{(k+1)}$
                \State $l_{max} \leftarrow \HL{\sum_i^n \frac{\mathcal{L}_i(x^{(k+1)})-l_i}{l_i}}$
            \EndIf
            \If{$k \in W$}
                \State \textbf{update} $\eta$ and $x^{(k+1)}$
            \EndIf
        \EndFor
    \end{algorithmic}
\end{breakablealgorithm}
\medskip
}

\section{Experiments}\label{sect:evaluation}
\revise{\textbf{RQ2}: \textit{Can adversarial attacks be designed to simultaneously attack all tasks in a multi-task model (empirically)?}
This section answers RQ2 by conducting experiments on different multi-task models with being attacked by \pjn{} to validate its effectiveness on attacking all tasks at once.}

\subsection{Experimental Settings}\label{sect:exp-settings}
\textbf{Datasets and Tasks}: We use two popular datasets in MTL, \textbf{NYUv2} \cite{silberman2012indoor} and \textbf{Tiny-Taskonomy} \cite{zamir2018taskonomy}. 
The NYUv2 dataset consists of RGB-D indoor scenes and three tasks, semantic segmentation, depth estimation, and surface normal prediction. 
Tiny-Taskonomy contains RGB indoor images, and its five representative tasks are semantic segmentation, surface normal prediction, depth estimation, keypoint detection, and edge detection. 

\textbf{Evaluation Metrics and Loss Functions}: 
Semantic segmentation uses a pixel-wise cross-entropy loss for each predicted class label. Surface normal prediction uses the inverse of cosine similarity between the normalized prediction and ground truth. 
All other tasks use the $l_{1}$ loss. 
Detailed metrics for each specific task are given in~\ref{sect:eval-metric}. 
Many tasks have distinct evaluation metrics with various scales. 
Compounding the problem of different scales is the fact that some metrics are higher-the-better (e.g., accuracy, mean of intersection over union), while others are lower-the-better (e.g., distance, error). 
To address the issues and fairly measure the success of attacks, we formulate a multi-task attack metric, \textbf{Average Relative Performance (ARP)}: 
\begin{equation}
\label{equ:arp}
     \frac{1}{N} \sum_{i=1}^{N} \frac{1}{M_i} \sum_{j=1}^{M_{i}} (-1)^{s_{i,j}}(m'_{i,j}-m_{i,j})/m_{i,j} \times 100\%,
\end{equation}
where $m_{i,j}$ and $m'_{i,j}$ represent the values of task $t_i$'s $j$-th metric for the model before and after the attack respectively, and $s_{i,j}$ equals 0 if this metric is lower-the-better and 1 otherwise. $M_i$ denotes the number of metrics for task $t_i$, and $N$ is the number of tasks. 
For each attack, we measure the corresponding ARP. 
\textit{A higher ARP indicates a higher performance drop and thus a more effective attack}.

\textbf{Multi-Task Models}: We evaluate branched multi-task models from TreeMTL \cite{zhang2022tree} using two backbone architectures: Deeplab-ResNet34 \cite{chen2017deeplab} and MobileNetV2 \cite{sandler2018mobilenetv2}. 
We randomly sampled and trained 25 models with Deeplab-ResNet34 and 20 with MobileNetV2 for the NYUv2 dataset. 
For the Tiny-Taskonomy dataset, we sampled 15 models with Deeplab-ResNet34. 
These models cover a range of parameter sharing configurations from the all-shared models to those comprising an ensemble of independent single-task models, represented by various model indexes. We provide the detailed model architectures and their corresponding model indexes in~\ref{sect:multi-task-models}. 

\textbf{Counterparts for Comparison}: 
We compare \pjn{} to three types of baselines. 
(i) The first type is the \textit{adapted multi-task attacks} that repurpose existing single-task white-box attacks to multi-task models. 
It includes \textsc{Total}, \textsc{SignTotal}, and \textsc{Single-x}. 
We compare these baselines with \pjn{} by integrating them with three different single-task white-box attack methods, FGSM \cite{goodfellow2014explaining}, PGD \cite{madry2018towards}, and APGD \cite{croce2020reliable}.  
(ii) The second type of baseline is a multi-model attack method called Auto-SAGA \cite{xu2022securing}, which is designed to attack multiple independent DNNs but can be directly applied to attack multi-task models. 
We integrate \pjn{} with Auto-SAGA to fairly compare with the original Auto-SAGA.  
(iii) The third type is 
a multi-task model attack approach called WGD \cite{ghamizi2022adversarial}. It introduces a task attack rate to optimally weigh each task's gradient following the loss balancing strategy from GradNorm \cite{chen2018gradnorm}, a multi-task optimization method.

\subsection{Attack on Clean Multi-Task Models} \label{sect:exp-overall}

\begin{table}[t]
\caption{ARP (\%, higher-the-better) of 8 multi-task models with diverse sharing patterns trained on \textbf{NYUv2} and attacked by PGD, APGD, and Auto-SAGA variants. The perturbation bound $\epsilon=8$. For brevity, the name of \textsc{Single-x} variants are simplified to the task name only. IND: independent, AS: all-shared.} \label{tab:NYUv2-resnet}
\centering
\scriptsize
\tabcolsep=0.01cm
\begin{tabular}{c|c|cccccc|cccccc|cc}
\toprule
\multirow{2}{*}{\begin{tabular}[c]{@{}c@{}} Model \\ Index\end{tabular}} & \multirow{2}{*}{\begin{tabular}[c]{@{}c@{}}\#Params \\ (M)\end{tabular}} & \multicolumn{6}{c|}{PGD}                                              & \multicolumn{6}{c|}{APGD}                                       & \multicolumn{2}{c}{Auto-SAGA} \\ \cline{3-16} 
                                                                             &                               & Segm  & Norm  & Dept            & Total & SignTotal & \textbf{\pjn{}}           & Segm  & Norm  & Dept            & Total & SignTotal & \textbf{\pjn{}}           & Baseline   & \textbf{\pjn{}}            \\ \hline
IND                                                                         & 63.83                         & 28.18 & 18.23 & 62.14           & 50.52 & 67.85     & \textbf{73.41}  & 30.74 & 23.52 & 75.62           & 58.30 & 69.29     & \textbf{87.58}  & 71.07      & \textbf{88.53}   \\
5                                                                            & 62.48                         & 29.12 & 30.00 & 70.99           & 61.40 & 71.46     & \textbf{79.60}  & 32.06 & 39.25 & 88.03           & 70.34 & 73.05     & \textbf{94.44}  & 84.51      & \textbf{95.70}   \\
35                                                                           & 62.25                         & 36.46 & 32.54 & 98.11           & 68.30 & 91.24     & \textbf{100.53} & 41.01 & 41.71 & 119.52          & 78.10 & 95.22     & \textbf{121.30} & 124.97     & \textbf{131.56}  \\
41                                                                           & 61.13                         & 38.06 & 46.02 & 100.41          & 67.51 & 95.20     & \textbf{103.03} & 42.09 & 59.54 & 120.39          & 75.42 & 99.05     & \textbf{122.50} & 113.15     & \textbf{126.94}  \\
21 & 55.65 & 31.45 & 39.55 & 79.10 & 65.00 & 76.70 & \textbf{84.81} & 34.80 & 57.48 & 96.28 & 74.13 & 78.75 & \textbf{100.94} & 92.91 & \textbf{102.72} \\
39                                                                           & 55.43                         & 36.03 & 45.68 & 83.78           & 67.83 & 90.40     & \textbf{96.63}  & 39.81 & 56.61 & 106.07          & 76.16 & 94.08     & \textbf{115.62} & 108.61     & \textbf{118.71}  \\
26                                                                           & 42.54                         & 34.19 & 45.39 & 80.94           & 69.39 & 88.58     & \textbf{93.88}  & 38.32 & 58.72 & 101.69          & 80.18 & 92.38     & \textbf{115.79} & 107.18     & \textbf{118.67}  \\
AS                                                                   & 21.28                         & 46.65 & 53.36 & \textbf{105.74} & 58.79 & 83.56     & 88.51           & 56.39 & 69.60 & \textbf{133.54} & 67.42 & 88.10     & 108.57          & 112.56     & \textbf{119.22}  \\ \bottomrule
\end{tabular}
\end{table}

\begin{table}[t]
\caption{ARP (\%, higher-the-better) of 8 multi-task models with diverse sharing patterns trained on \textbf{Tiny-Taskonomy} and attacked by PGD, APGD, and Auto-SAGA variants. The perturbation bound $\epsilon=8$.
}
 \label{tab:Taskonomy-resnet1}
\centering
\scriptsize
\tabcolsep=0.01cm
\begin{tabular}{c|c|ccccccc|ccccccc|cc}
\toprule
\multirow{2}{*}{\begin{tabular}[c]{@{}c@{}} Model \\ Index\end{tabular}} & \multirow{2}{*}{\begin{tabular}[c]{@{}c@{}}\#Params \\ (M)\end{tabular}} & \multicolumn{7}{c|}{PGD}                                                    & \multicolumn{7}{c|}{APGD}                                                      & \multicolumn{2}{c}{Auto-SAGA}     \\ \cline{3-18} 
                                                                             &                               & Segm   & Norm  & Dept   & Keyp  & Edge  & Total           & \textbf{\pjn{}}           & Segm            & Norm  & Dept   & Keyp  & Edge  & Total           & \textbf{\pjn{}}           & Baseline        & \textbf{\pjn{}}           \\ \hline
IND                                                                         & 106.38                        & 248.04 & 23.13 & 117.44 & 12.48 & 11.50 & \textbf{282.01} & 274.17          & 245.25          & 23.21 & 117.83 & 12.40 & 11.37 & \textbf{279.28} & 270.25          & \textbf{282.01} & 274.17          \\
190                                                                          & 105.03                        & 190.91 & 49.41 & 138.67 & 14.98 & 14.02 & 232.55          & \textbf{236.63} & 175.02          & 45.06 & 130.69 & 14.26 & 13.64 & 213.47          & \textbf{218.29} & 216.32          & \textbf{218.88} \\
358                                                                          & 103.68                        & 189.90 & 50.42 & 152.70 & 18.14 & 15.53 & 237.12          & \textbf{247.29} & 174.80          & 46.32 & 142.73 & 17.07 & 14.73 & 217.99          & \textbf{227.61} & 220.93          & \textbf{228.97} \\
959                                                                          & 96.86                         & 201.59 & 48.17 & 163.56 & 15.56 & 17.64 & 241.75          & \textbf{256.35} & 183.75          & 44.46 & 150.37 & 14.70 & 17.04 & 221.31          & \textbf{234.58} & 223.54          & \textbf{234.56} \\
1020                                                                         & 83.53                         & 222.92 & 49.98 & 139.97 & 18.25 & 18.01 & 259.27          & \textbf{263.03} & 202.50          & 44.34 & 130.95 & 17.02 & 16.89 & 235.80          & \textbf{240.26} & 238.95          & \textbf{241.14} \\
1043                                                                         & 75.59                         & 205.27 & 44.26 & 145.47 & 23.99 & 38.96 & 246.90          & \textbf{250.79} & 189.32          & 39.97 & 136.29 & 22.55 & 36.50 & 227.23          & \textbf{233.25} & 230.54          & \textbf{234.28} \\
1037                                                                         & 62.48                         & 216.15 & 45.51 & 152.47 & 17.61 & 47.35 & 252.55          & \textbf{261.82} & 197.99          & 41.05 & 141.10 & 16.88 & 40.50 & 230.86          & \textbf{240.70} & 240.00          & \textbf{246.43} \\
AS                                                                   & 21.28                         & 231.23 & 77.53 & 104.10 & 16.02 & 22.40 & \textbf{231.08} & 196.07          & \textbf{227.99} & 76.06 & 103.84 & 15.97 & 21.87 & 226.97          & 192.85          & \textbf{231.07} & 196.07          \\ \bottomrule
\end{tabular}
\end{table}

This subsection compares \pjn{} with baselines on their effectiveness in attacking clean multi-task models.
Tables~\ref{tab:NYUv2-resnet} and \ref{tab:Taskonomy-resnet1} compare the attack performance of the baselines and \pjn{} at the model level on NYUv2 and Tiny-Taskonomy respectively at $\epsilon=8$. 
Overall, \pjn{} integrated with PGD and APGD have the highest ARP in 7 out of the 8 models on NYUv2, and in 6 out of the 8 models on Tiny-Taskonomy. 
Full tables for all models and more results with $\epsilon=4$ are included in~\ref{sect:attack-on-clean-more}.

\pjn{} outperforms baselines \textsc{Total}, \textsc{SignTotal}, and \textsc{Single-x} in attack performance because it alleviates the baselines' limitations discussed in Attack Framework.
As illustrated in Figure \ref{fig:transfer}, \textsc{Single-x} achieves limited attack effectiveness on tasks that are not the attack target \textsc{x} (also discussed in Section \ref{sect:exp-sharing}); \textsc{Total} shares a similar pattern of attack effectiveness across tasks as \textsc{Single-Segm} due to the issue of gradient dominance, making it less effective in attacking all tasks simultaneously. 
In contrast, \pjn{} dynamically balances the attack directions for all tasks, making it more threatening for systems that require high robustness.
This is also the reason why \pjn{} outperforms the multi-model attack approach AutoSAGE --- \pjn{} balances the gradients in AutoSAGE.

\begin{figure}
  \centering
  \begin{subfigure}{0.23\textwidth}
    \includegraphics[width=\linewidth]{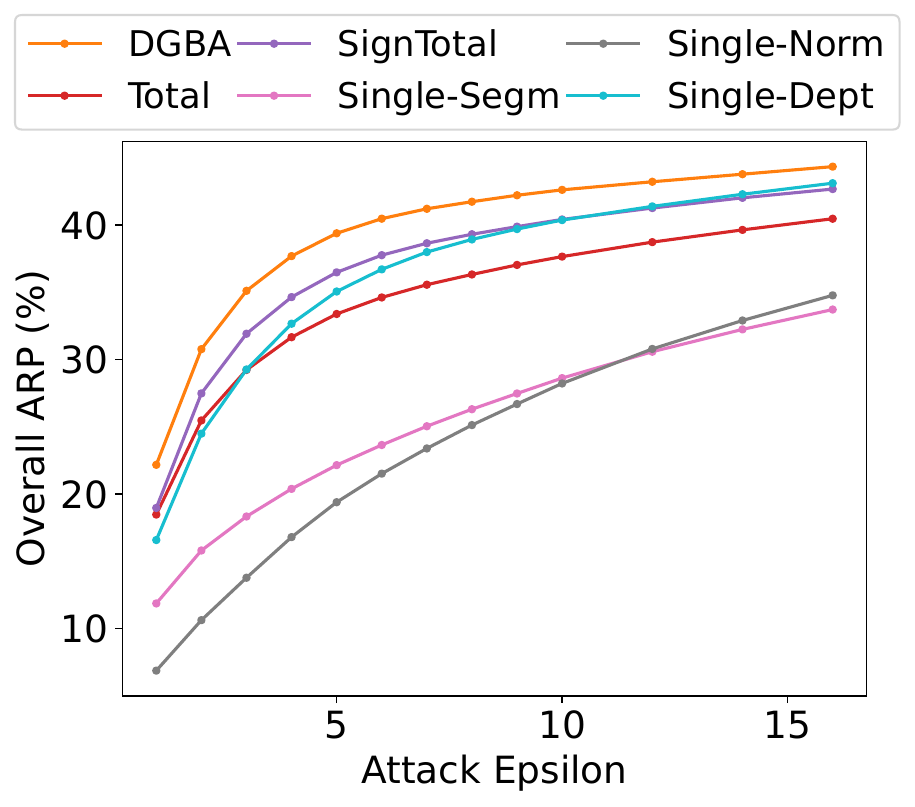}
    \caption{FGSM}
  \end{subfigure}
  \hfill
  \begin{subfigure}{0.23\textwidth}
    \includegraphics[width=\linewidth]{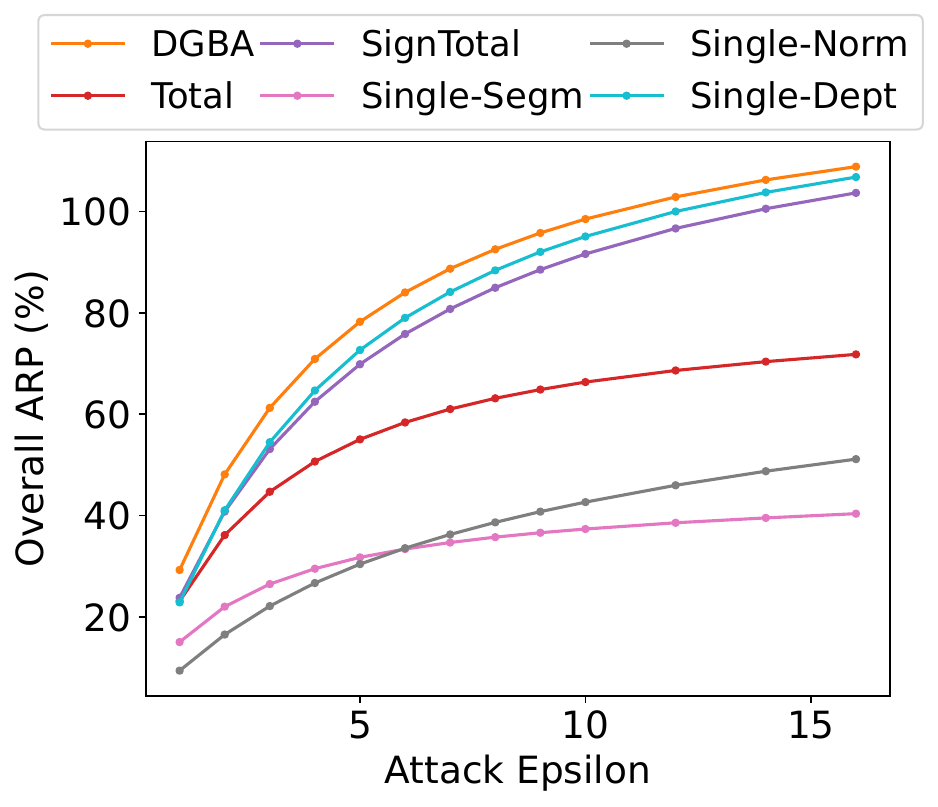}
    \caption{PGD}
  \end{subfigure}
  \hfill
  \begin{subfigure}{0.23\textwidth}
    \includegraphics[width=\linewidth]{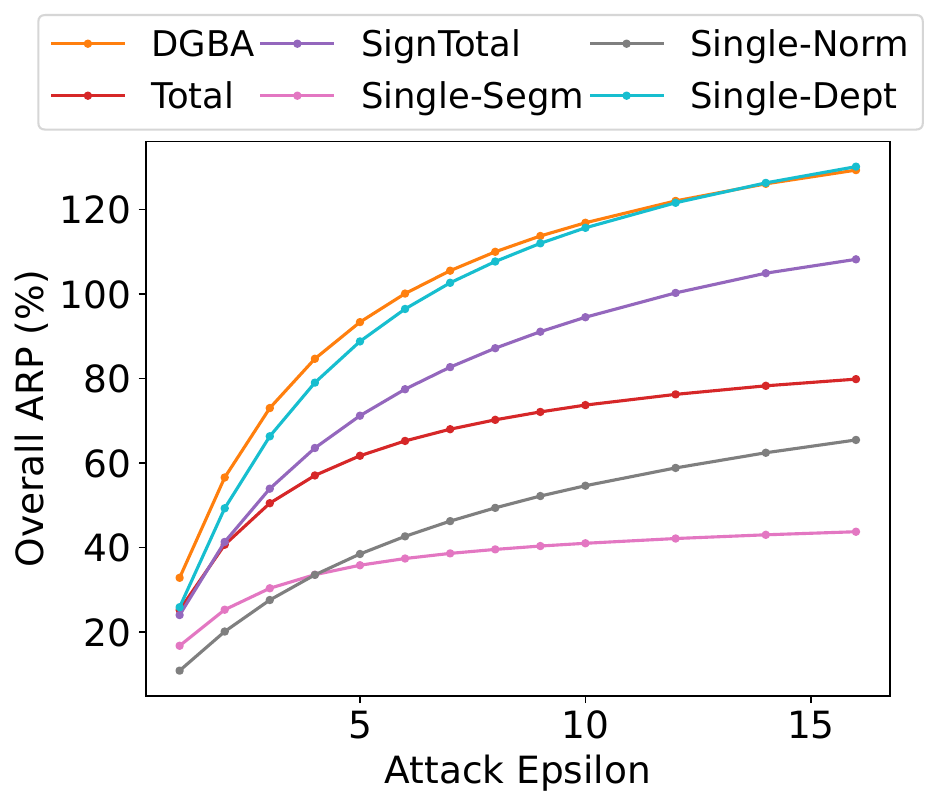}
    \caption{APGD}
  \end{subfigure}
  \hfill
  \begin{subfigure}{0.23\textwidth}
    \includegraphics[width=\linewidth]{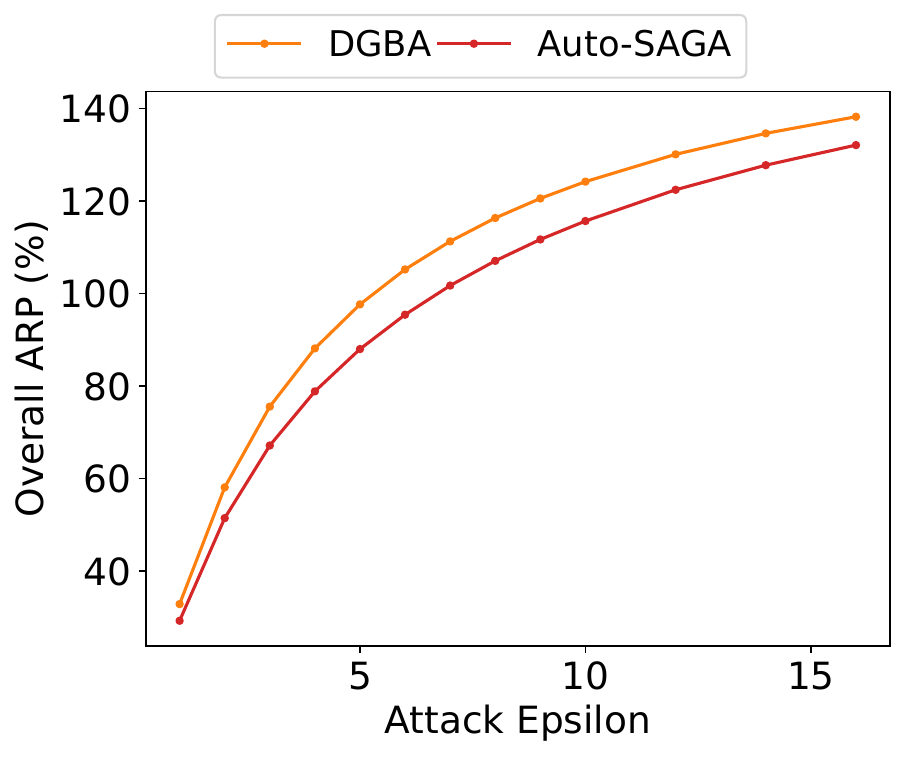}
    \caption{Auto-SAGA}
  \end{subfigure}
  \caption{Attack performance comparisons in terms of ARP averaged over 25 multi-task models trained on NYUv2 with Deeplab-ResNet34. The adapted attacks and \pjn{} variants are built on (a) FGSM, (b) PGD, (c) APGD, and (d) Auto-SAGA. The perturbation bound $\epsilon$ ranges from 1 to 16.} \label{fig:NYUv2-resnet-overall}
\end{figure}

\begin{figure}
  \centering
  \begin{subfigure}{0.23\textwidth}
    \includegraphics[width=\linewidth]{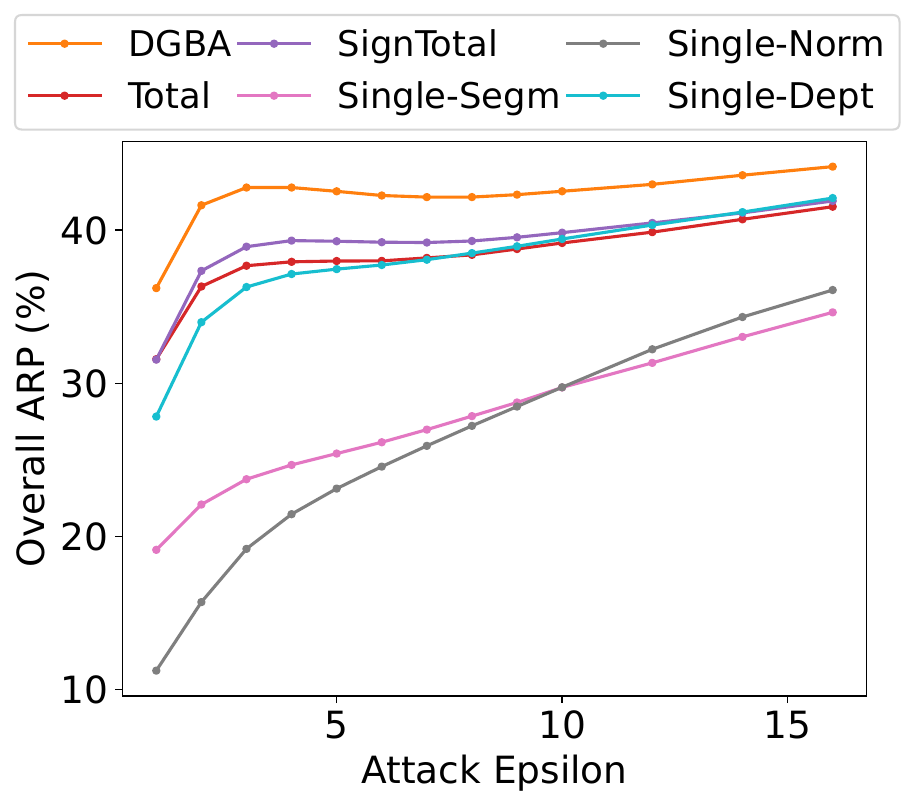}
    \caption{FGSM}
  \end{subfigure}
  \hfill
  \begin{subfigure}{0.23\textwidth}
    \includegraphics[width=\linewidth]{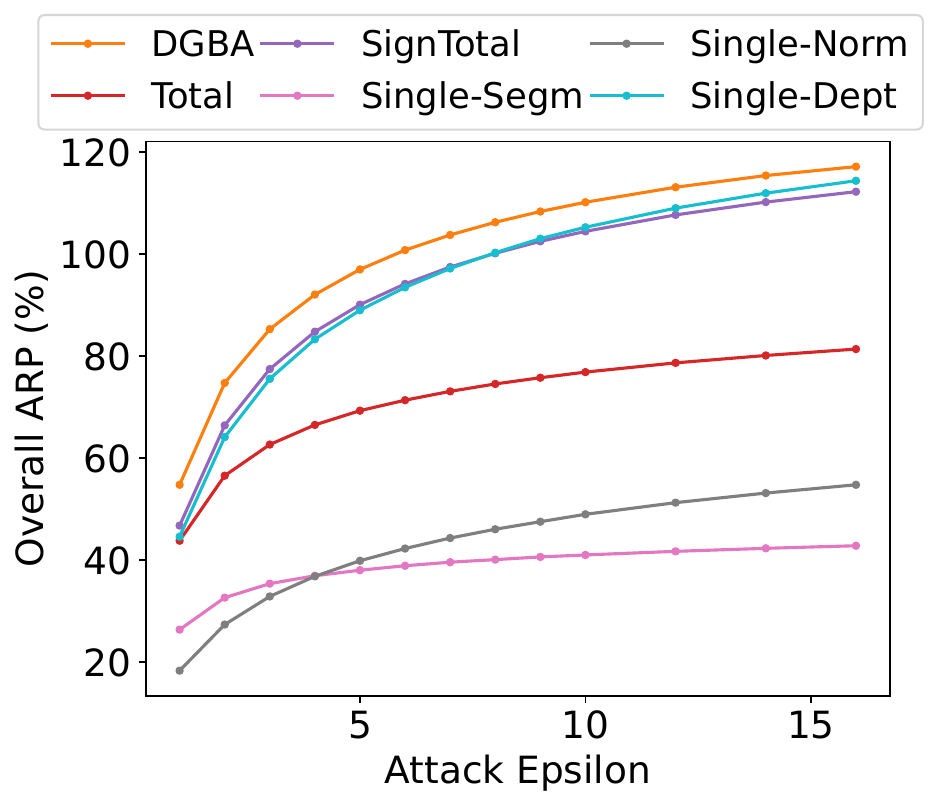}
    \caption{PGD}
  \end{subfigure}
  \hfill
  \begin{subfigure}{0.23\textwidth}
    \includegraphics[width=\linewidth]{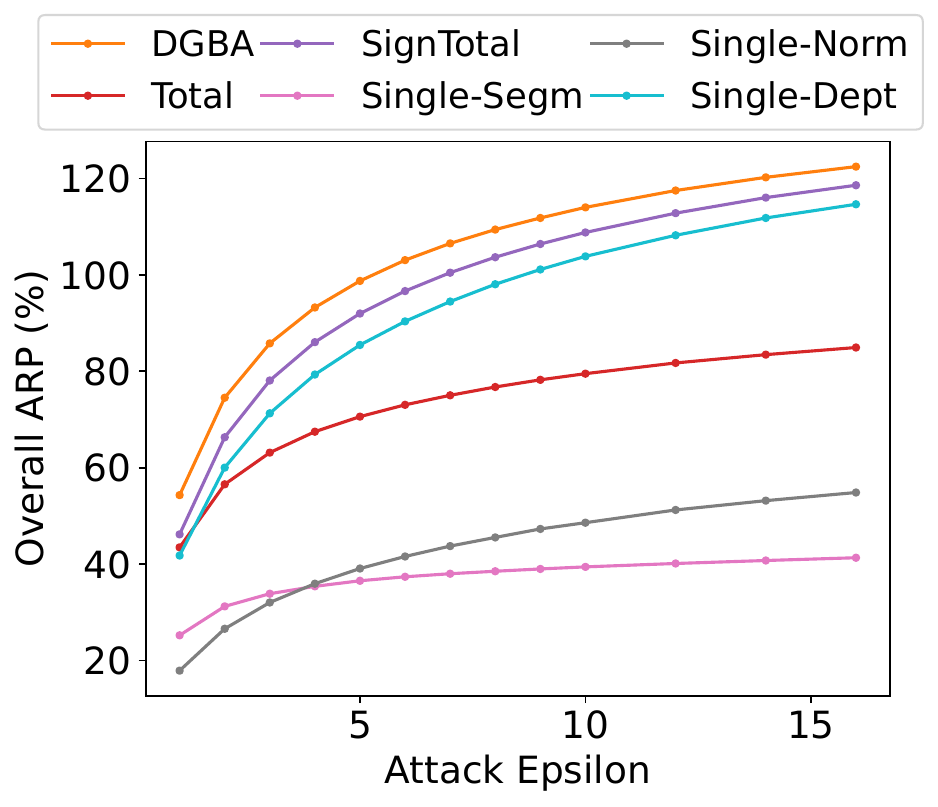}
    \caption{APGD}
  \end{subfigure}
  \hfill
  \begin{subfigure}{0.23\textwidth}
    \includegraphics[width=\linewidth]{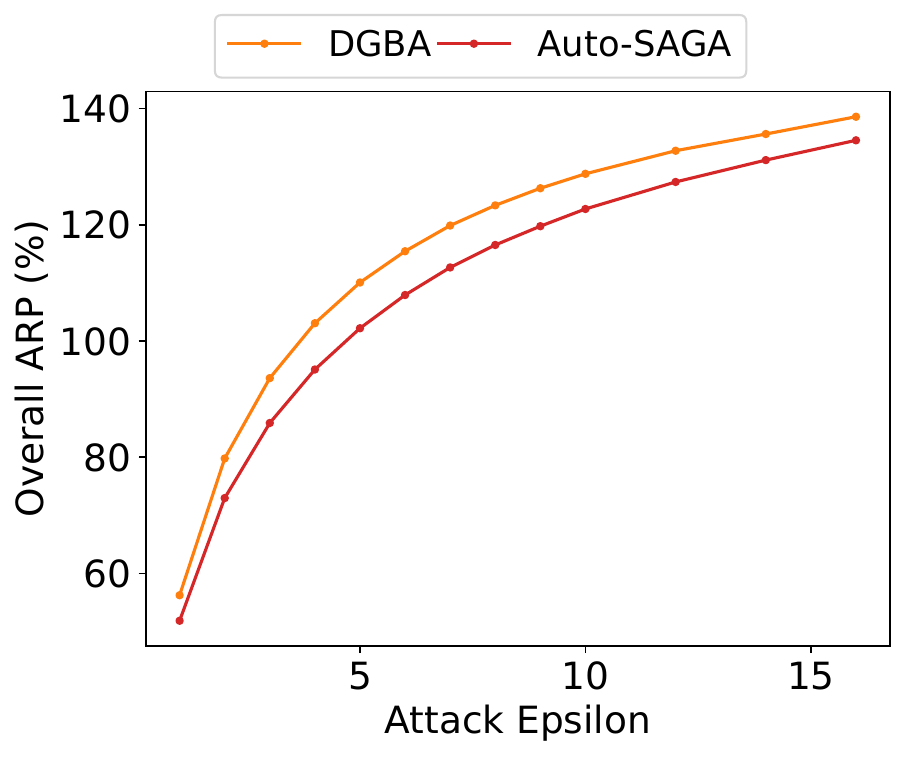}
    \caption{Auto-SAGA}
  \end{subfigure}
  \caption{Attack performance comparisons in terms of ARP on NYUv2 with MobileNetV2 similar to Figure \ref{fig:NYUv2-resnet-overall}.} \label{fig:NYUv2-mobilenet-overall}
\end{figure}

Figures~\ref{fig:NYUv2-resnet-overall} and \ref{fig:NYUv2-mobilenet-overall} further report the influence of the perturbation bounds on the attack performance with different backbone models, Deeplab-ResNet34 and MobileNet respectively. 
We vary the maximum perturbation bound $\epsilon$ from 1 to 16 and report the average of the ARP over all multi-task models with diverse architectures.
We denote the average ARP as the \textbf{overall ARP}. 
DGBA is the best-performing attack in almost all $\epsilon$ and dataset settings, and when integrated with any of the three existing single-task (i.e., FGSM, PGD, APGD) or the multi-model attack (i.e., Auto-SAGA) approaches. 
There are a few cases, where for $\epsilon \geq 10$, DGBA and $\textsc{Total}$ converge or perform almost identically.
This is because at higher $\epsilon$ values, the magnitude of the noise becomes larger (and thus more visible) and all attacks become more effective. We provide visual adversarial samples in~\ref{sect:attack-on-adv-more}.

\begin{figure}[h]
  \centering
  \begin{subfigure}{1\textwidth}
    \includegraphics[width=.4\linewidth]{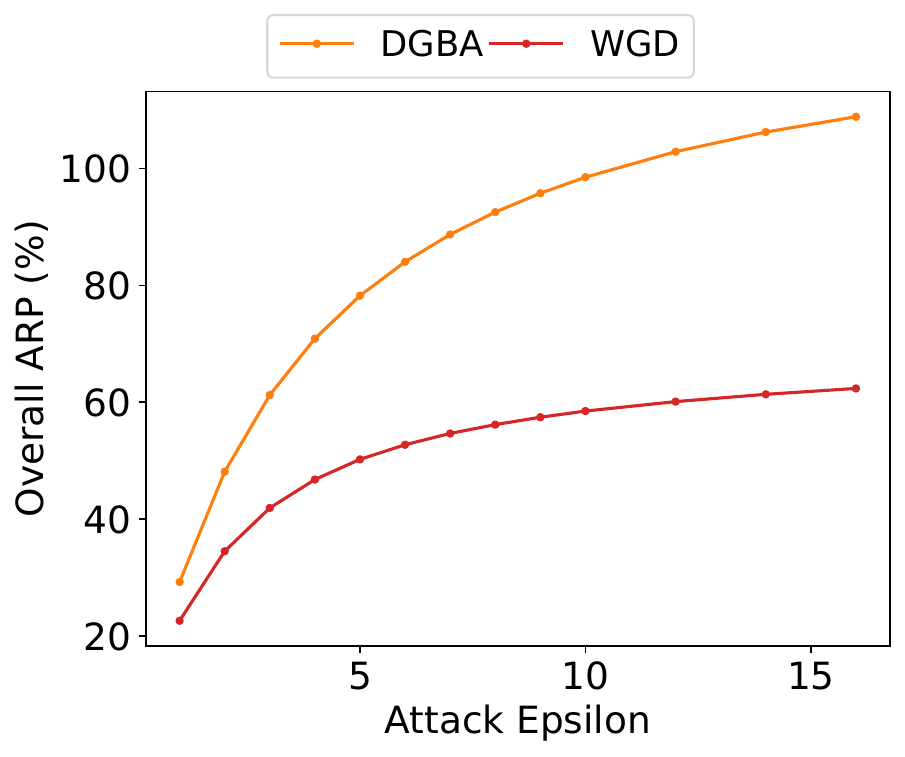}
    \includegraphics[width=.5\linewidth]{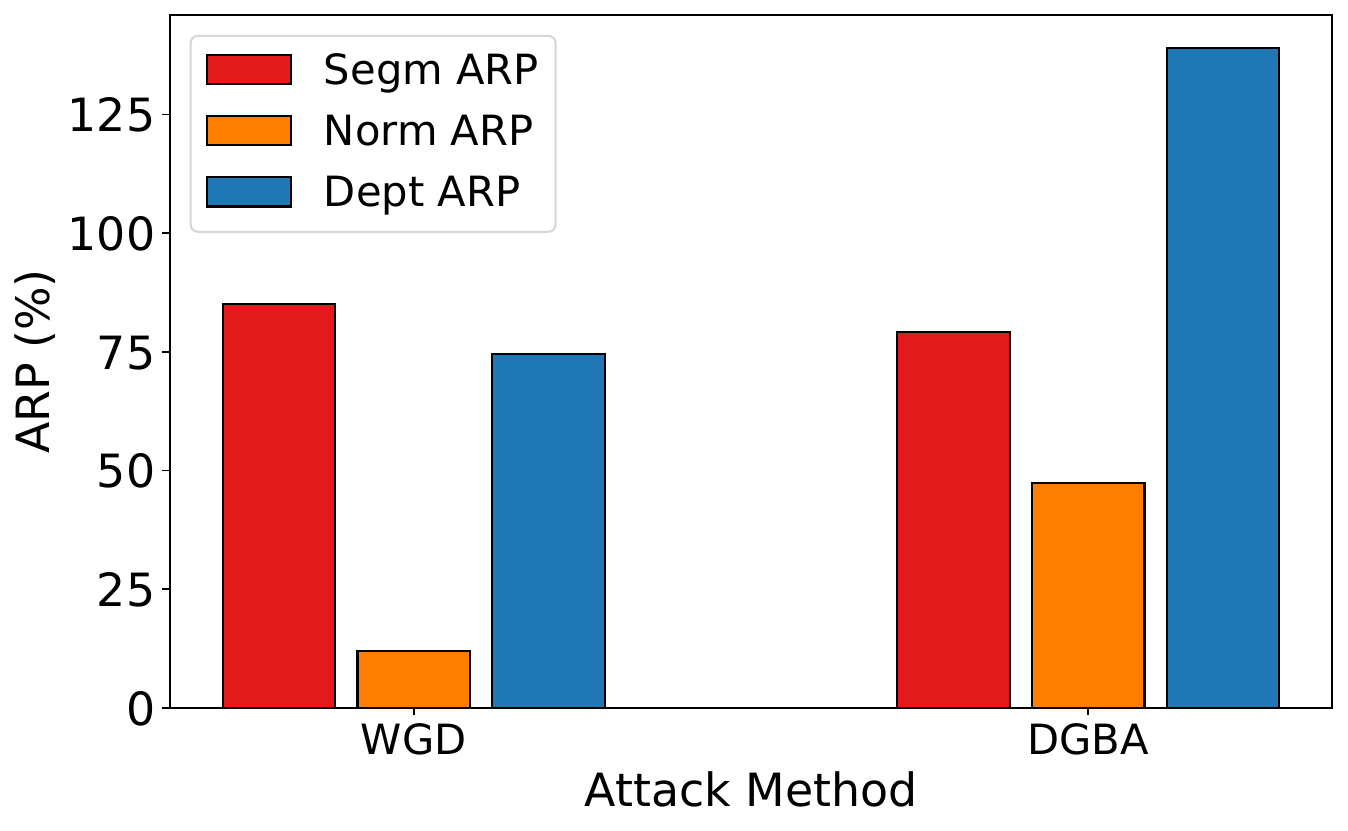}
    \caption{PGD}
  \end{subfigure}
  \begin{subfigure}{1\textwidth}
    \includegraphics[width=.4\linewidth]{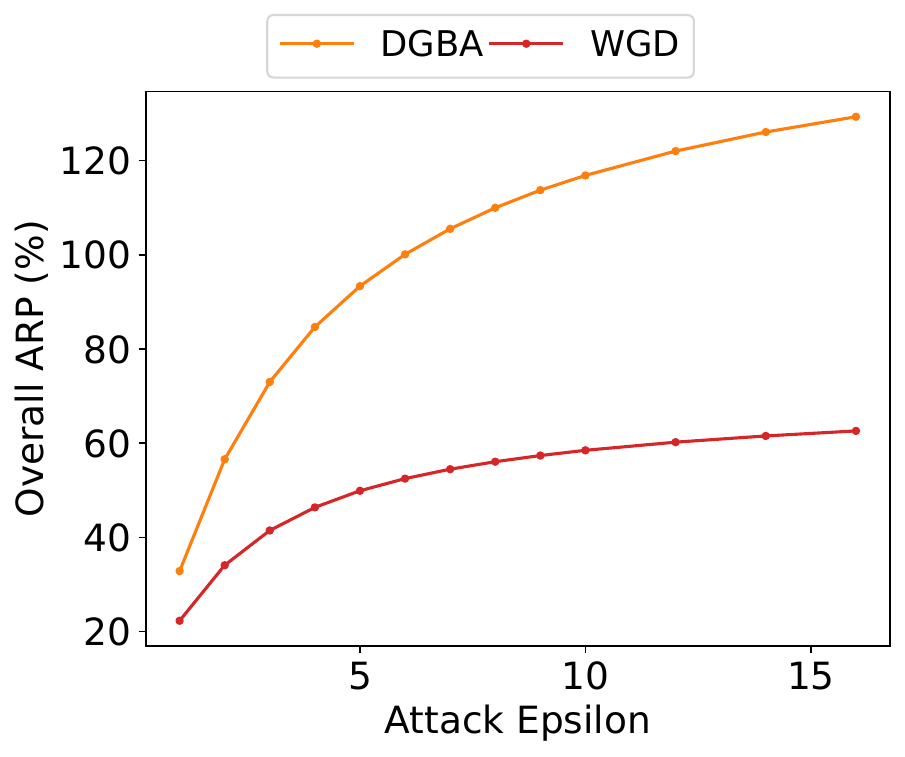}
    \includegraphics[width=.5\linewidth]{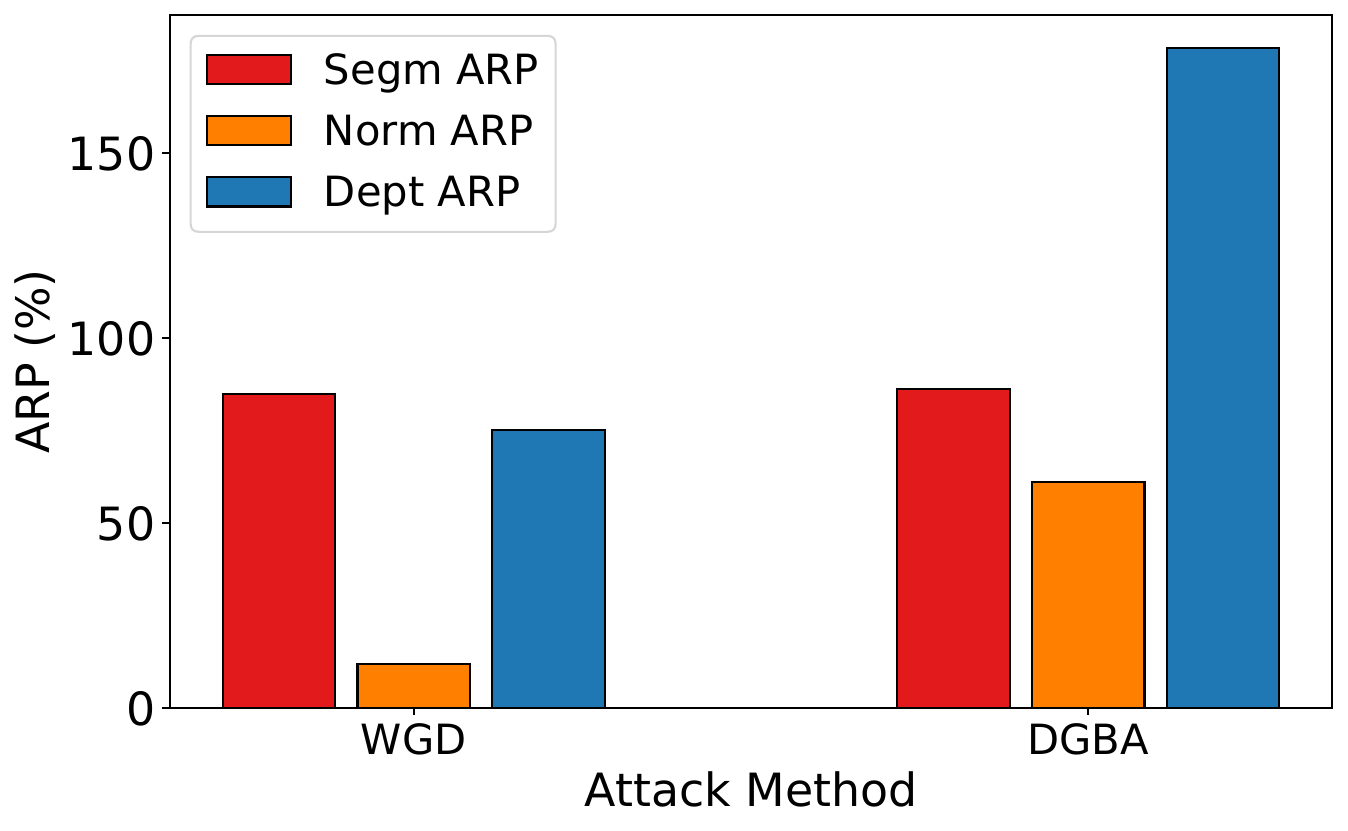}
    \caption{APGD}
  \end{subfigure}
  \caption{Attack performance comparisons between WGD and \pjn{} on NYUv2 with Deeplab-ResNet34. The left figures compare the average ARP over 25 multi-task models while the right ones illustrate the attack effectiveness on each task.} \label{fig:wgd_comparison}
\end{figure}

Figure~\ref{fig:wgd_comparison} presents the attack performance comparisons between \pjn{} and WGD, built on top of PGD and APGD attack methods. The left panels report the average ARP over multi-task models trained with Deeplab-ResNet34 on the NYUv2 dataset, while the right figures illustrate the attack effectiveness on each individual task, similar to Figure~\ref{fig:transfer}.
Overall, \pjn{} significantly outperforms WGD spanning all $\epsilon$ settings and achieves more balanced attack effectiveness across all tasks compared to WGD. When generating perturbations, WGD assigns more importance to tasks experiencing larger perturbations during the attack, which could amplify the task dominance phenomenon and thus lead to worse attack performance compared to \pjn{}.

\subsection{Attack on Adversarially Trained Models}\label{sect:exp-adv}
A common single task strategy to defend against adversarial attacks is to leverage adversarial training \cite{madry2018towards,bai2021recent,zhang2020attacks}. 
This defense involves generating adversarial samples and using them as part of the dataset during training. This typically results in reduced model performance on clean inputs, but increased robustness to adversarial attacks.
\revise{Although adversarial training has been incorporated with multi-task settings in prior work~\cite{mao2020multitask,ghamizi2023gat}, they typically aim to enhance the robustness of a single primary task by introducing auxiliary tasks. It remains unclear whether adversarial training can simultaneously enhance the robustness of all tasks within a multi-task model, especially when incorporated with our proposed balanced adversarial samples.}

\begin{table}[t]
\caption{ARP of multi-task models trained with and without adversarial training and then attacked by multi-task attacks. The multi-task attacks include \pjn{}, \textsc{Single}, and \textsc{Total} with $\epsilon=8$, while the adversarial training is the FAT version of them with $K=\tau=20$. (higher-the-better) } \label{tab:adv_ARP}
\centering
\small
\tabcolsep=0.03cm
\begin{tabular}{c|cccc|c}
\toprule
Adv. Train  & Single-Segm & Single-Norm & Single-Dept & Total & DGBA  \\ \midrule
w/o AT      & 46.65 & 53.36 & \textbf{105.74} & 58.79 & 88.51          \\
Single-Segm & 14.67 & 6.60  & 16.70           & 19.41 & \textbf{19.92} \\
Single-Norm & 20.20 & 10.61 & 25.52           & 25.78 & \textbf{29.26} \\
Single-Dept & 16.60 & 8.41  & 14.12           & 18.48 & \textbf{20.31} \\
Total       & {\color[HTML]{3166FF} 13.21} & 6.26  & 13.16           & 16.03 & {\color[HTML]{3166FF} \textbf{16.62}} \\
\pjn{}      & 13.58 & {\color[HTML]{3166FF} 5.97}  & {\color[HTML]{3166FF} 12.76}           & {\color[HTML]{3166FF} 15.98} & \textbf{16.64} \\ \bottomrule
\end{tabular}
\end{table}

We adopt the single task Friendly Adversarial Training (FAT) \cite{zhang2020attacks} to MTL. 
FAT is a PGD-based adversarial training method. 
In order to apply this defense effectively to MTL, we modify the default PGD to DGBA-PGD to use adversarial examples generated from multiple tasks as opposed to just single task adversarial examples. 
Our new defense is coined \textbf{FAT DGBA} (see~\ref{sect:attack-on-adv-more} for the detailed algorithm). 
We also train multi-task models using FAT with the adapted MTL attacks \textsc{Single} and \textsc{Total}. 
The accuracy of the models after adversarial training are included in the~\ref{sect:attack-on-adv-more}. 
After adversarial training, we re-evaluate the multi-task attacks to assess both model robustness and attack effectiveness. 

Table \ref{tab:adv_ARP} reports the ARP of 30 cases, i.e., (without adversarial training + 5 adversarially trained models) $\times$ 5 attack methods. The multi-task models are trained on NYUv2 with Deeplab-ResNet34 and attacked by PGD-variants.
We make two main observations. 
First, the notable reduction in ARP of attack methods (e.g., from $105.74\%$ to $12.76\% - 25.52\%$ when attacking with \textsc{Single-Dept}) demonstrates that models enhanced with adversarial training are substantially more robust than the model without such training, especially with FAT \textsc{Total} and \pjn{} (in blue).
Second, from the perspective of attack performance, \pjn{} is still the most effective attack method (in bold), consistently outperforming the \textsc{Single-x} and \textsc{Total} baselines by up to $18.65\%$.

\section{Parameter Sharing and Model Robustness}
\label{sect:exp-sharing}
\revise{\textbf{RQ3}: \textit{How does parameter sharing affect the robustness of multi-task models to adversarial attacks?}
We raise this question because} existing literature in MTL determines the appropriate level of parameter sharing with a focus on optimizing task accuracy, while our study reveals the relationship between parameter sharing and model robustness (measured by ARP) with \pjn{}. 
For example, with the level of parameter sharing increases, the ARP increases from 87.58 to 108.57 (see Table \ref{tab:NYUv2-resnet} \pjn{}-APGD). This result indicates that \textbf{a higher degree of parameter sharing is also associated with increased adversarial vulnerability}. 
This increased vulnerability potentially arises from the fundamental trade-off between improved task accuracy from positive task interactions enabled by sharing, and increased susceptibility to adversarial attacks due to \textbf{the greater transferability of perturbations across shared parameters}.

\begin{figure}[htb]
\centering
  \includegraphics[width=.65\linewidth]{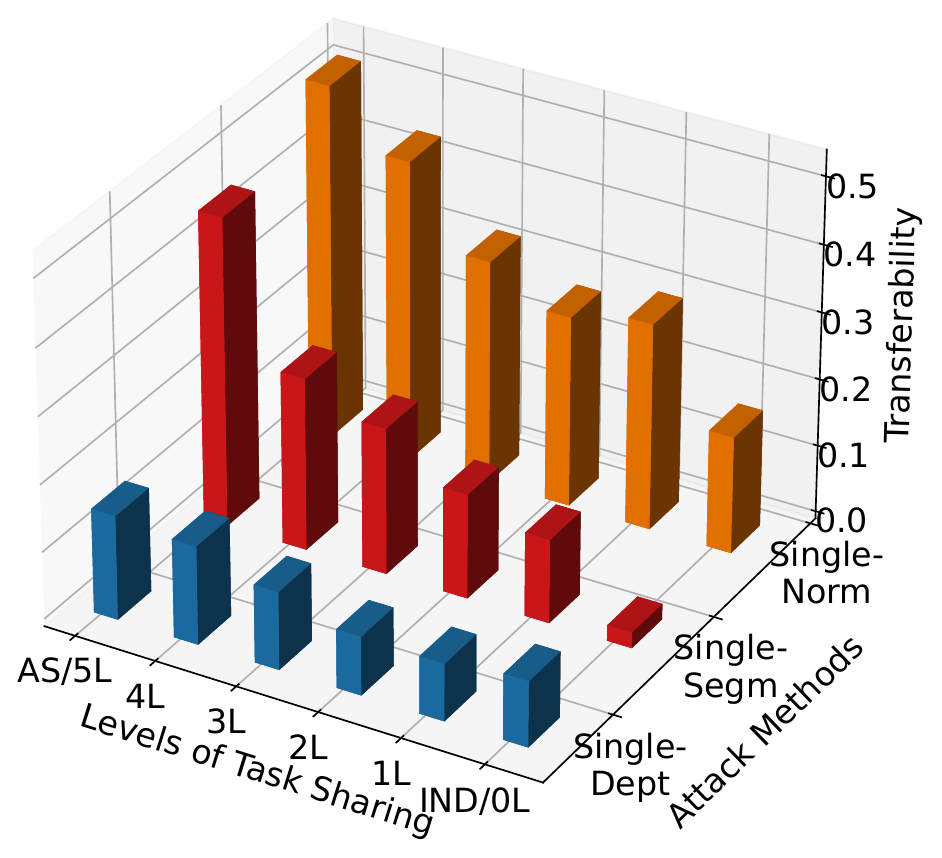}
  \caption{The relationship between the levels of parameter sharing in multi-task models (x-axis) and the attack transferability (z-axis). The y-axis represents APGD variants \textsc{Single-x}.
  } \label{fig:shareVSrobust}
\end{figure}

To verify our hypothesis, we first define \textit{attack transferability} in the MTL context. When using an attack method like \textsc{Single-x}, which attacks only one task in a multi-task model, the rest of the tasks in the multi-task model that are not the target could be affected.
We measure attack transferability as the ratio between the performance of the attack on all non-targeted tasks versus the performance of the attack on the targeted task. 
Mathematically, let \textsc{x} represent a task under attack, and \textsc{y} represent another task. The transferability of attack \textsc{Single-x} is
    $\frac{1}{n-1}\sum_{\text{\textsc{y}} \neq \text{\textsc{x}}} \frac{\text{ARP-\textsc{y}}}{\text{ARP-\textsc{x}}}$,  
where $n$ represents the number of tasks in the multi-task model and ARP-x represents the average relative performance of task x. 
If the ratio is close to zero, this means that attacking task x has minimal influence on the performance of task y.


Figure \ref{fig:shareVSrobust} illustrates the relationship between the levels of parameter sharing and attack transferability, showing the results for six multi-task models attacked by APGD \textsc{Single-x} variants. 
These models represent six levels of parameter sharing, ranging from all-share (AS/5L), where all layers of the backbone model (5L) are shared, to independent models (IND/0L), where no layers are shared.
We observe a positive correlation between the degree of parameter sharing and the transferability of attacks in the three \textsc{Single-x} variants. 
As the level of parameter sharing increases, attack transferability also increases, from $0.08$ to $0.15$ ($1.875\times$) for \textsc{Single-Dept}, $0.02$ to $0.46$ ($23\times$) for \textsc{Single-Segm}, and $0.17$ to $0.53$ ($3.12\times$) for \textsc{Single-Norm}.
These findings suggest a trade-off in multi-task model design: While sharing more parameters among related tasks enhances task accuracy, it also amplifies attack transferability, thereby reducing model robustness, even when facing single task attacks. 

\section{Conclusion}
This work makes significant contributions in understanding the adversarial robustness of multi-task learning (MTL). 
We first analyzed adaptions of single task white-box attacks to the MTL domain and experimentally demonstrated their ineffectiveness. 
We then developed a new attack framework \pjn{} to effectively attack all tasks in multi-task models. 
On models trained on the NYUv2 and Tiny-Taskonomy datasets, \pjn{} achieves the highest overall attack strength and is the strongest attack on 6 out of 8 models for both datasets. 
From the defense side, we also adversarially trained multi-task models called FAT \pjn{}. \pjn{} is the most effective attack, even on these multi-task models. 
We further analyzed the adversarial transferability of MTL adversarial examples and discovered a new phenomenon: Task sharing can lead to increased adversarial transferability. 
By highlighting the adversarial vulnerability, our work advocates for future research into developing robust and secure multi-task vision models for safety-critical applications.
\revise{Another interesting extension of this work is to adapt \pjn{} to attack large-scale multimodal foundation models—such as GPT-4o or Gemini—which typically operate in black-box or API-limited settings. This presents a non-trivial challenge, as it requires extending \pjn{} beyond white-box supervision and designing new strategies for gradient-free multi-task attacks.}

\section{Acknowledgement}
This material is based upon work supported by the National Science Foundation under Grant No. 2312396, 2220211, 2224054, and 2247893. Any opinions, findings, and conclusions or recommendations expressed in this material are those of the author(s) and do not necessarily reflect the views of the National Science Foundation.

\newpage
\appendix
\revise{\section{Notation Table}\label{sect:notation}
To improve readability and clarify our use of variables, we summarize the main mathematical symbols used throughout the paper in Table~\ref{tab:notation}. 
Note that although inputs $x$ and perturbations $\delta$ are matrix-valued (e.g., pixel-level images), we follow the common convention in the adversarial learning literature of using italic symbols for both vectors and matrices. 
This notation is consistent with prior works such as APGD~\cite{croce2020reliable} and Auto-SAGA~\cite{xu2022securing}, and allows for easier comparison with standard formulations in the field.}

\begin{table}[h]
\centering
\tabcolsep=0.03cm
\caption{Summary of Notation} \label{tab:notation}
\begin{tabular}{lll}
\toprule
\textbf{Symbol} & \textbf{Type} & \textbf{Role} \\
\midrule
$ x $           & Matrix       & Input image \\
$ x_{adv} $     & Matrix       & Adversarial image \\
$ y_i $         & Scalar/Matrix        & Ground truth label for task $ i $ \\
$ \mathcal{L}_i $ & Scalar func & Loss function for task $ i $ \\
$ \mathcal{L} $   & Scalar func & Combined multi-task loss \\
$ \delta $         & Matrix       & Adversarial perturbation \\
$ \eta $        & Matrix       & Magnitude for perturbation direction \\
$ \beta $        & Matrix       & Binary mask for perturbation direction \\
$ \Delta\mathcal{L} $ & Scalar    & Normalized loss increase after applying perturbation \\
$ n $            & Scalar       & Number of tasks in the multi-task model \\
$ T $            & Scalar       & Number of PGD iterations \\
\bottomrule
\end{tabular}
\end{table}

\section{Related Works}\label{sect:related}
\textbf{Adversarial Attacks.}
Adversarial attacks fall into two main categories: white-box and black-box attacks~\cite{mahmood2021back}.
In white-box attacks, an attacker has access to the target model's internal information, enabling direct gradient extraction and adversarial example generation~\cite{carlini2019evaluating}. On the contrary, in black-box attacks, an attacker has limited model knowledge and uses alternative information sources \cite{chen2017zoo,zhou2020dast} to create adversarial examples. 
This paper focuses on white-box attacks, as they are generally more effective compared to black-box attacks \cite{croce2020reliable,wang2022iwa}.
%

In recent years, many white-box attack techniques have been developed.
Fast Gradient Sign Method (FGSM)~\cite{goodfellow2014explaining} generates adversarial examples by introducing non-random noise in the gradient direction of the loss function.
Projected Gradient Descent (PGD)~\cite{madry2018towards} and Momentum Iterative Method (MIM) \cite{dong2018boosting} improve FGSM by generating adversarial samples in an iterative process. 
Later, Croce and Hein \cite{croce2020reliable} proposed Auto Projected Gradient Descent (APGD) with adaptive step size and combined it with two complementary attacks \cite{croce2020minimally, andriushchenko2020square}, developing the ensemble method APGD that outperforms existing methods on diverse benchmark datasets. 
In addition to gradient-based strategies, alternative methods have emerged. 
For instance, the Backward Pass Differentiable Approximation (BPDA) \cite{athalye2018obfuscated} accommodates non-differentiable functions, while the Carlini and Wagner ($C\&W$) attack \cite{carlini2017towards} perturbs images with minimal delta to misclassify them. 

\textbf{Multi-Task Learning.}
In Multi-Task Learning (MTL), researchers develop memory and computation-efficient multi-task models that simultaneously address multiple tasks \cite{ruder2017overview,yao2020survey}. 
The main challenge lies in determining the parameters to share across tasks to optimize both resource efficiency and task accuracy. This has led to an abundance of multi-task model architectures designed either manually~\cite{huang2015cross,jou2016deep,dvornik2017blitznet,kokkinos2017ubernet,ranjan2017hyperface} or automatically~\cite{sun2019adashare,zhang2022tree,ahn2019deep,guo2020learning,zhang2022automtl}. 
This paper focuses on attacking branched multi-task models, which are the most representative in the MTL literature~\cite{vandenhende2019branched,guo2020learning,bruggemann2020automated,zhang2022tree}. 
Besides, branched multi-task models have a wide range of sharing patterns across tasks, facilitating the study of the relationship between the robustness of multi-task models and their sharing patterns.

\textbf{Robustness of Multi-Task Models.} Few studies have examined the robustness of the model in MTL settings.
A pioneering study \cite{mao2020multitask} pointed out that the adversarial robustness of deep neural networks increases as the number of tasks increases. 
Subsequent research \cite{ghamizi2022adversarial,li2023sibling} further emphasizes the importance of selecting suitable tasks for joint learning to create more robust models.
While these studies offer intriguing insights, they do not specifically propose adversarial attack methods for general multi-task models.
Regarding attacks on multi-task models, MTA \cite{guo2020multi} tries to develop attacks in the MTL setting, however, the generated adversarial samples are task-specific and thus fail to attack all the tasks simultaneously. 
Some other work \cite{gurulingan2021uninet,sobh2021adversarial} attempted to attack the multi-task model by generating adversarial examples for each image while attacking one task at a time or simply targeting the joint multi-task loss.
\revise{Recent work~\cite{chen2023learning} proposes a task-agnostic adversarial sample generator that can attack multiple independent models trained for different tasks. However, its applicability to the unified multi-task models, where tasks share parameters, is not demonstrated and remains an open question.
While SMTA~\cite{guo2024stealthy} explicitly targets multi-task models, but with a fundamentally different goal: degrading a single target task, while preserving performance on others, i.e., a stealthy attack. 
MTADV~\cite{wang2024multi} introduces a multi-task adversarial attack framework for facial authentication, but it is designed specifically for cross-modal facial datasets and does not generalize to general vision tasks.
Different from existing methods, the proposed \pjn{} aims at jointly degrading all tasks in a multi-task model, regardless of the model’s sharing structure and target vision tasks.}

\section{MTL Attack Optimization Approximation}\label{sect:theory-details}
The optimization problem we formulate for multi-task attack in Section 3 of the main paper is,
\begin{equation} \label{equ:app_obj_ILP}
\begin{aligned}
    \beta^\ast &= \arg\max_\beta \Delta \mathcal{L} \\
    &= \arg\max_\beta \frac{1}{N} \sum_{i=1}^n \frac{\mathcal{L}_i(x + \eta \cdot \beta,y_i)-\mathcal{L}_i(x,y_i)}{\mathcal{L}_i(x,y_i)}.
\end{aligned}
\end{equation}
Here, $\mathcal{L}_i(x + \eta \cdot \beta,y_i)-\mathcal{L}_i(x,y_i)$ represents the model loss difference for task $t_i$ before and after the attack.

As the problem is intractable, we make some approximations and reformulate it to be an Integer Linear Programming (ILP) problem. To do so, we first apply the Taylor expansion on $\mathcal{L}_i(x + \eta \cdot \beta,y_i)$ in the numerator of the objective function at the point of $x$:
\begin{equation} \label{equ:num-taylor}
\begin{aligned}
    &\mathcal{L}_i(x + \eta \cdot \beta,y_i)-\mathcal{L}_i(x,y_i) \\
    &= \mathcal{L}_i(x,y_i) + \eta \cdot \beta \cdot \frac{\partial\mathcal{L}_i(x,y_i)}{\partial x} + \xi - \mathcal{L}_i(x,y_i) \\
    &\approx \eta \cdot \beta \cdot \frac{\partial\mathcal{L}_i(x,y_i)}{\partial x}.
\end{aligned}
\end{equation}
We ignore the remainder $\xi$ because, in the context of adversary attacks, we have $\lVert \eta \cdot \beta \rVert_p \leq \epsilon$, indicating that the change $\eta \cdot \beta$ is sufficiently small.

The proposed approximate optimization problem for multi-task attacks is thus formulated as follows:
\begin{equation} 
\begin{aligned}
    \beta^\ast = \arg\max_\beta \frac{1}{N} \sum_{i=1}^n \eta \cdot \beta \cdot \frac{\partial\mathcal{L}_i(x,y_i)}{\partial x} \cdot \frac{1}{\mathcal{L}_i(x,y_i)} \\
    s.t. \quad \forall \beta^{(k)} \in \beta: \beta^{(k)} \in \{-1,0,1\},
\end{aligned}
\end{equation}
where $\frac{1}{N}$ and $\eta$ are constants that can be ignored when solving the optimization problem. In practice, $\beta$ and $\frac{\partial\mathcal{L}_i(x,y_i)}{\partial x}$ are two matrices with the same size as $x$, thus their product represents the dot product of the corresponding vectorized matrices. 

\revise{\textbf{Why the reminder $\xi$ in Eq.~\ref{equ:num-taylor} is negligible?}}
$\xi$ is mainly the second-order term in Taylor expansion, which is relatively small compared to the first-order term and thus can be ignored.
We provide an analytic upper bound on the second–order remainder and quantify its size relative to the first–order term.

Formally, the second-order term is,
\begin{equation} \label{equ:remainder}
\begin{aligned}
\xi = \tfrac12\delta^\top H_{i}(x + \delta)\delta
\end{aligned}
\end{equation}
where $H_{i}=\nabla^{2}_{x}\mathcal L_{i}$ is the Hessian.  
Since $\delta=\eta\cdot\beta$ with the element-wise mask $\beta\!\in\!\{-1,0,1\}^{m}$ and the $\ell_{\infty}$ budget $\lVert\eta\rVert_{\infty}\leq\epsilon$, we bound the quadratic remainder as,
\begin{equation} \label{equ:remainder_bound}
\begin{aligned}
\left|\tfrac12\,\delta^{\top}H_{i}(\cdot)\,\delta\right|
  &\le \tfrac12\,\varepsilon^{2}\,\lVert H_{i}\rVert_{\max}\,\lVert\beta\rVert_{1},
\end{aligned}
\end{equation}
where $\lVert H_{i}\rVert_{\max}\!=\!\max_{p,q}|(H_{i})_{p,q}|$.

Then the relative ratio between the second-order and first-order term would be,
\begin{equation} \label{equ:ratio_def}
\begin{aligned}
\mathcal R_i
  :=\frac{\bigl|\text{remainder}\bigr|}
          {\bigl|\text{first-order term}\bigr|}
  \leq\frac{\tfrac12\,\epsilon^{2}\lVert H_{i}\rVert_{\max}\lVert\beta\rVert_{1}}
         {\epsilon\,\lVert\nabla_{x}\mathcal L_{i}(x)\rVert_{1}}
  =\frac{\epsilon}{2}
      \frac{\lVert H_{i}\rVert_{\max}}
           {\lVert\nabla_{x}\mathcal L_{i}(x)\rVert_{1}\big/\lVert\beta\rVert_{1}}.
\end{aligned}
\end{equation}

Empirically, for the all-shared multi-task models with Deeplab-ResNet34 and over 100 images in the NYUv2 dataset used in Section 4, we measure $\frac{\lVert H_{i}\rVert_{\max}}{\lVert\nabla_{x}\mathcal L_{i}(x)\rVert_{1}\big/\lVert\beta\rVert_{1}}$ in Table~\ref{tab:taylor-quantites}.

\begin{table}[htb]
\caption{The ratio in Eq.~\ref{equ:ratio_def} measured in practice over 100 images in NYUv2 dataset on all-shared multi-task models with Deeplab-ResNet34.}\label{tab:taylor-quantites}
\centering
\tabcolsep=0.2cm
\begin{tabular}{c|ccc}
\toprule
Ratio & \textbf{Mean $\pm$ Std} & Min & Max \\ \midrule
$\frac{\lVert H_{i}\rVert_{\max}}{\lVert\nabla_{x}\mathcal L_{i}(x)\rVert_{1}\big/\lVert\beta\rVert_{1}}$         &  $\mathbf{9.61 \pm 3.66e-04}$ &  $4.18e-04$   &  $1.63e-03$       \\ \bottomrule
\end{tabular}
\end{table}
Therefore taking the mean of the empirical results and the common setting of $\epsilon = 8$, the ratio $\mathcal R_i \le \mathrm{3.84e-03}$, which means the second–order term contributes less than $0.38\%$ of the linear term, confirming that the linearization error is negligible.

\revise{\textbf{Why we can solve Eq.~\ref{equ:app_obj_ILP} with LP relaxation?}}
As a commonly-used technique in solving ILP, LP relaxation and rounding do not introduce approximation error for our attack optimization problem~\cite{williamson2011design}.

Notice that both $\beta$ and $\frac{\partial\mathcal{L}_i(x,y_i)}{\partial x}$ in Eq.~\ref{equ:app_obj_ILP} are pixel-level matrices as the input image $x$, we can decompose it into pixel-level maximization in pixel $k$.
\begin{equation} 
\label{equ:app_obj_ILP_pixel}
\begin{aligned}
\beta^{\star}= \arg\max_{\beta^{(k)}\in\{-1,0,1\}} \sum_{k=1}^{m} c^{(k)}\,\beta^{(k)},
\qquad
c^{(k)} = \sum_{i}\frac{\partial\mathcal L_{i}(x,y_{i})}{\partial x^{(k)}}
            \,\frac{1}{\mathcal L_{i}(x,y_{i})},
\end{aligned}
\end{equation} 
where $m$ is the total number of pixels in $x$.
Because the objective is \textbf{linear and separable}, the global maximizer is obtained \emph{coordinate-wise}:
\begin{equation} \label{equ:best_beta}
\begin{aligned}
\beta^{(k)\star} =
  \begin{cases}
  +1,&\text{if } c^{(k)}>0,\\
  -1,&\text{if } c^{(k)}<0,\\
   0,&\text{if } c^{(k)}=0.
  \end{cases}
\end{aligned}
\end{equation} 

This analytical maximizer matches precisely the algorithm we employ in practice.  
Starting from an all-zero mask, we iteratively update $\beta$ along the gradient direction $\partial\Delta\mathcal L/\partial\beta$, i.e., $\sum_{i=1}^n \frac{\partial \mathcal{L}_i(x,y_i)}{\partial x} \cdot \frac{1}{\mathcal{L}_i(x,y_i)}$. This ascent drives every active coordinate to its extreme value, i.e., $\pm1$.  
The integer solution is then obtained by a simple sign operation, $\beta^{\star}=\operatorname{sign}(\beta)$.  
Accordingly, the LP relaxation followed by sign rounding is \textbf{exact} for \pjn{}, introducing no discretization error.  

\section{Auto-SAGA and \pjn{}-Auto-SAGA}\label{sect:autosaga-dgba}
As presented in Section 3 in the main paper, integrating \pjn{} with any existing attack can be easily accomplished by substituting the single-task gradient, i.e., $\frac{\partial\mathcal{L}(x)}{\partial x}$, with the balanced multi-task counterpart, i.e., $\sum_{i=1}^n \frac{\partial\mathcal{L}_i(x,y_i)}{\partial x} \cdot \frac{1}{\mathcal{L}_i(x,y_i)}$.
In addition to Algorithm 1 in the main paper, we further illustrate how to integrate \pjn{} into the multi-model attack algorithm Auto-SAGA.  

For Auto-SAGA \cite{rathbun2022game}, the original attack formulation is
\begin{equation}
    {
    \begin{array}{c}
         \displaystyle G_{blend}(x_{adv}^{(k)}) = \gamma G_{blend}(x_{adv}^{(k-1)}) + \sum_{k \in D \backslash R} \alpha^{(k)}_{k} \phi^{(k)}_{k} \odot \frac{\partial \mathcal{L}_{s}}{\partial x_{adv}^{(k)}}  \\
         \displaystyle + \sum_{r \in R} \alpha^{(k)}_{r} \phi^{(k)}_{r} \odot (\mathbb{E}_{t \sim T}[\frac{\partial \mathcal{L}_{r}}{\partial t(x^{(k)}_{adv})}]).
    \end{array}}
    \label{equ:ae}
\end{equation}

To integrate \pjn{} with Auto-SAGA, we normalize the gradient of the objective function $\mathcal{L}_{s}$ and $\mathcal{L}_{r}$ with the objective function value. The updated attack will be,
\begin{equation}
    {
    \begin{array}{c}
         \displaystyle G_{blend}(x_{adv}^{(k)}) = \gamma G_{blend}(x_{adv}^{(i-1)}) + \sum_{k \in D \backslash R} \alpha^{(k)}_{k} \phi^{(k)}_{k} \odot (\frac{\partial \mathcal{L}_{s}}{\partial x_{adv}^{(k)}} \HL{\cdot \frac{1}{\mathcal{L}_{s}}} )  \\
         \displaystyle + \sum_{r \in R} \alpha^{(k)}_{r} \phi^{(k)}_{r} \odot (\mathbb{E}_{t \sim T}[\frac{\partial \mathcal{L}_{r}}{\partial t(x^{(k)}_{adv})} \HL{\cdot \frac{1}{\mathcal{L}_{r}}}]).
    \end{array}}
    \label{equ:ae-MTL}
\end{equation}

\section{More Experimental Results}\label{sect:more-exp}
This section reports evaluation metrics and more experimental results that are omitted from the main paper. 

\subsection{Evaluation Metrics}\label{sect:eval-metric}
\textit{Semantic segmentation} is evaluated using mean Intersection over Union and Pixel Accuracy (mIoU and Pixel Acc, the higher the better) in NYUv2. 

\textit{Surface normal prediction} is evaluated using mean and median angle distances between the prediction and the ground truth (the lower the better), and the percentage of pixels whose prediction is within the angles of $11.25\degree$, $22.5\degree$ and $30\degree$ to the ground truth (the higher the better). 

\textit{Depth estimation} uses the absolute and relative errors between prediction and ground truth (the lower the better). 
Furthermore, the percentage of pixels whose prediction is within the thresholds of $1.25, 1.25^2, 1.25^3$ to the ground truth, i.e. $\delta=\max\{\frac{p_{pred}}{p_{gt}}, \frac{p_{gt}}{p_{pred}}\}<thr$, is used (the higher the better). 

Tiny-Taskonomy is evaluated using the task-specific loss of each task directly.

\subsection{Attack on Clean Multi-Task Models}\label{sect:attack-on-clean-more}
\begin{figure}[htb]
  \centering
  \begin{subfigure}{0.45\textwidth}
    \includegraphics[width=\linewidth]{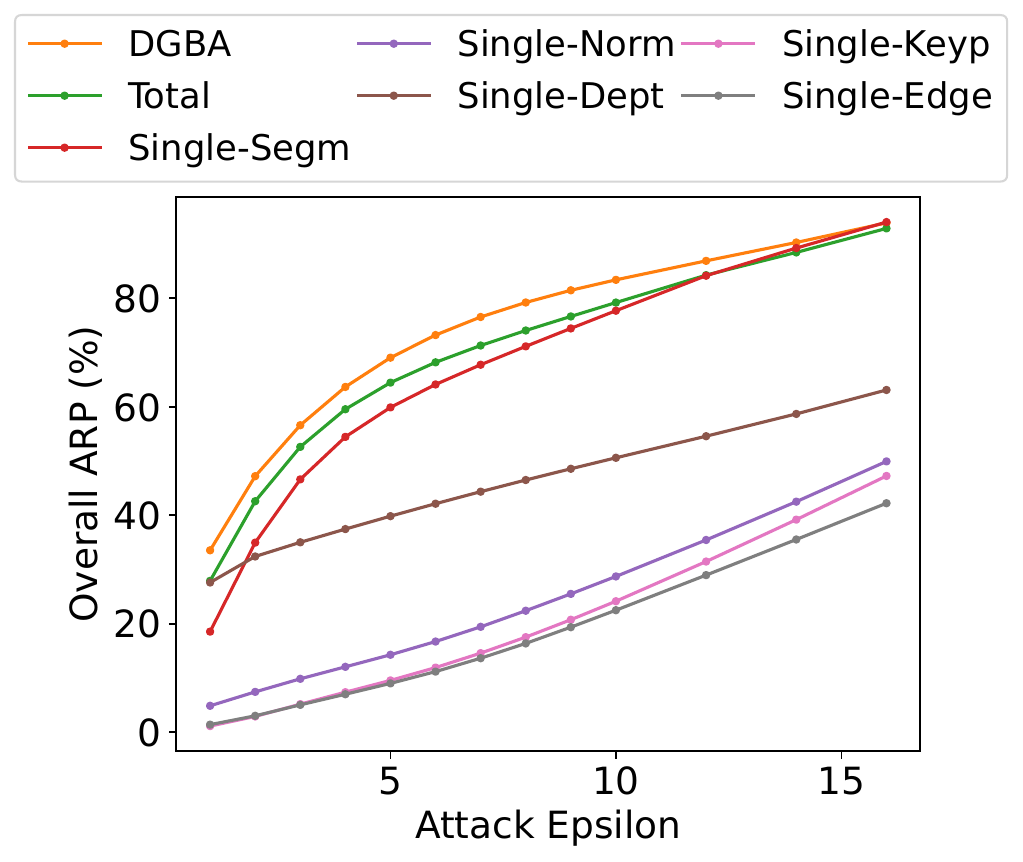}
    \caption{FGSM}
  \end{subfigure}
  \hfill
  \begin{subfigure}{0.45\textwidth}
    \includegraphics[width=\linewidth]{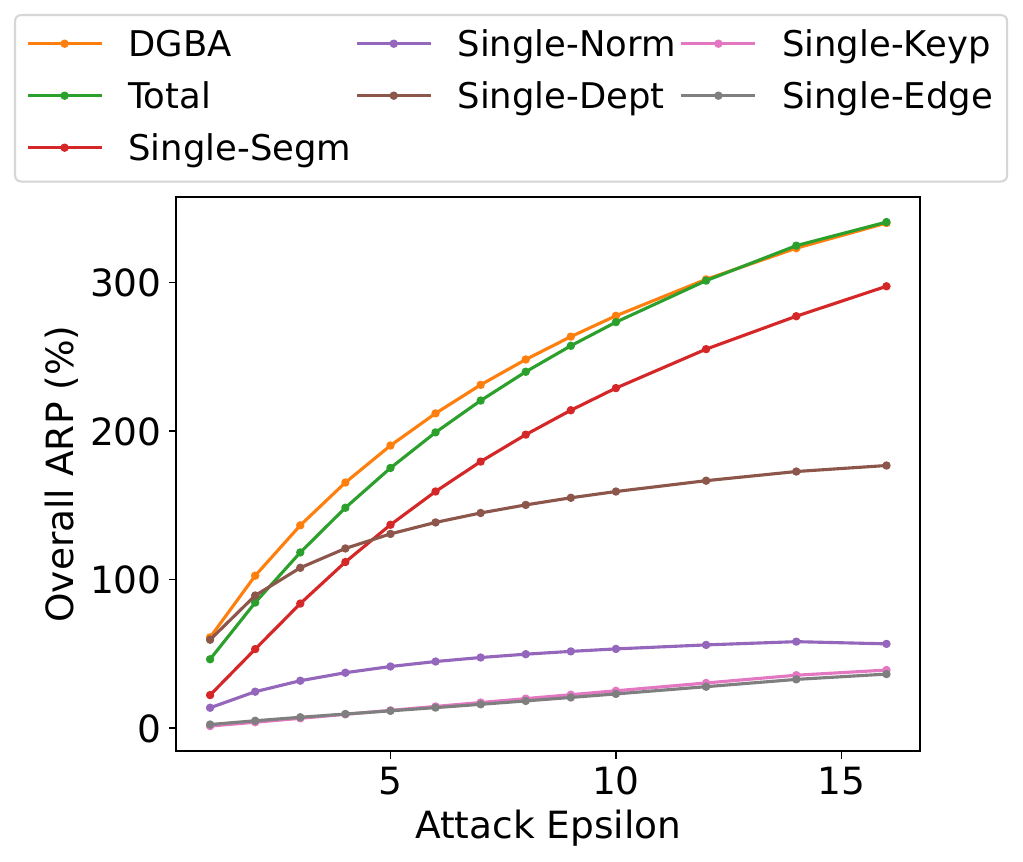}
    \caption{PGD}
  \end{subfigure}
  \\
  \begin{subfigure}{0.45\textwidth}
    \includegraphics[width=\linewidth]{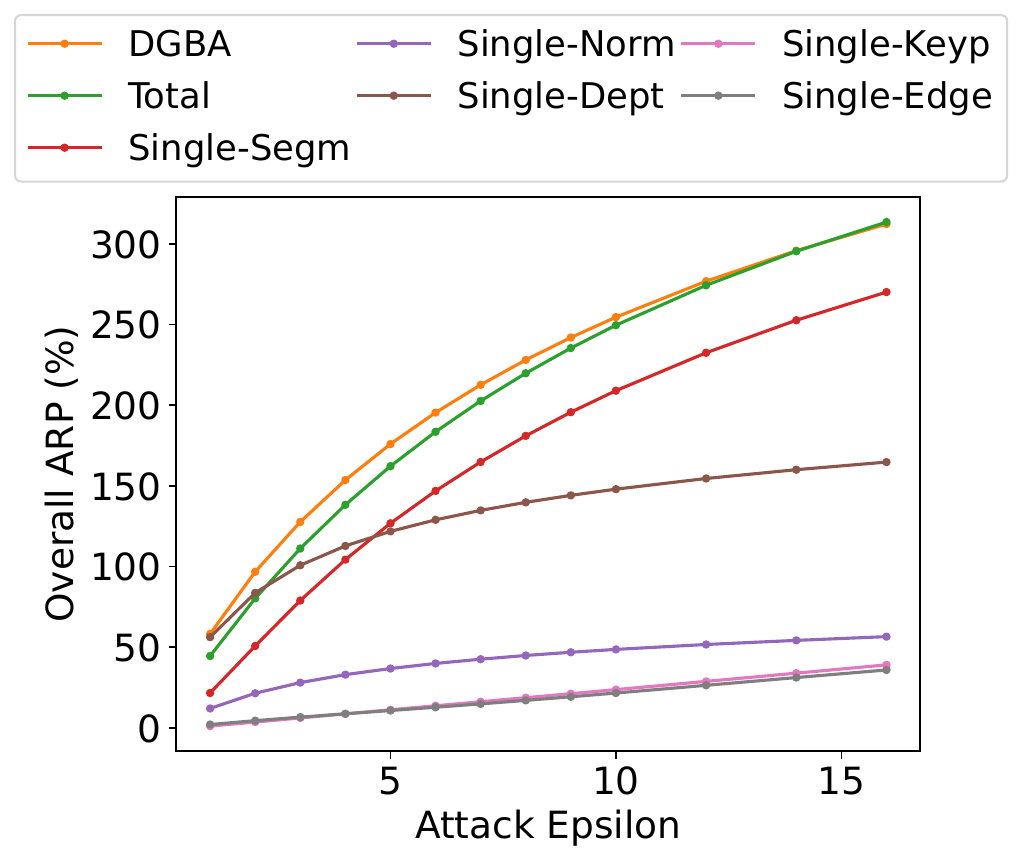}
    \caption{APGD}
  \end{subfigure}
  \hfill
  \begin{subfigure}{0.45\textwidth}
    \includegraphics[width=\linewidth]{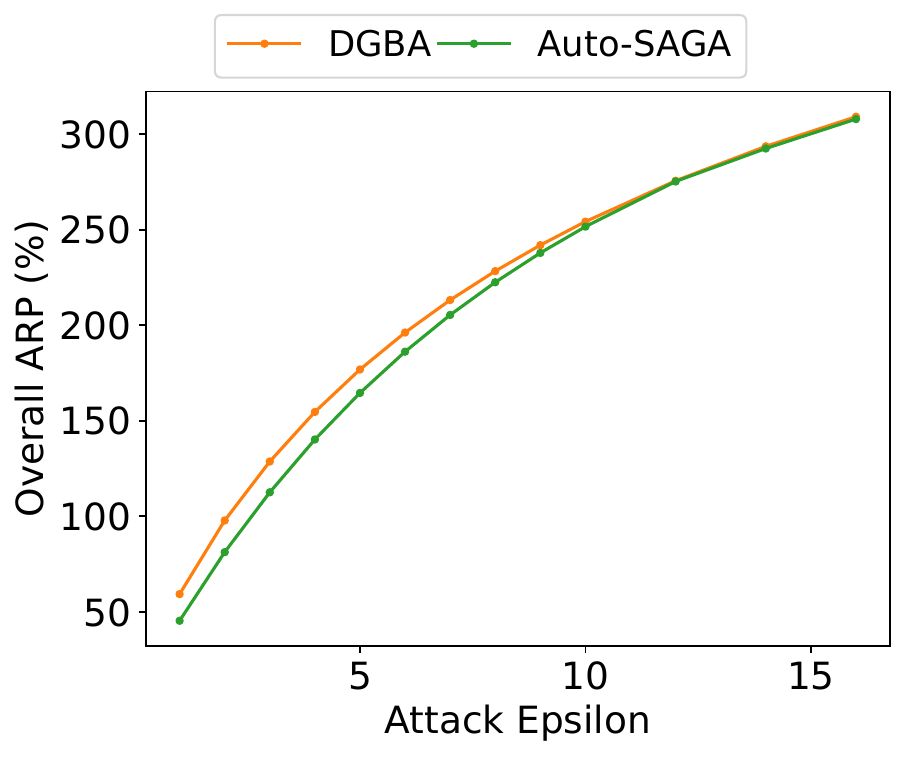}
    \caption{Auto-SAGA}
  \end{subfigure}
  \caption{Attack performance comparisons in terms of ARP averaged over 15 multi-task models trained on Tiny-Taskonomy with Deeplab-ResNet34.} \label{fig:Taskonomy-resnet-overall}
\end{figure}

Figure \ref{fig:Taskonomy-resnet-overall} illustrates the attack performance on Tiny-Taskonomy with Deeplab-ResNet34. 
To be consistent with the main paper, we conduct the experiments using different variants of \pjn{} and the adapted multi-task attacks. 
The x-axis represents the perturbation bound $\epsilon$ ranging from 1 to 16, while the y-axis displays the overall Average Relative Performance (ARP, higher-the-better). 
Overall, \pjn{} achieves competitive results compared with baselines and always outperforms baselines when the perturbation is small.

We present the full tables of ARP after the attack of all 25 multi-task models trained on NYUv2 with Deeplab-ResNet34 for perturbation bound $\epsilon=8$ in Tables \ref{tab:NYUv2-resnet1-full} and \ref{tab:NYUv2-resnet2-full}. 
Similarly, Tables \ref{tab:Taskonomy-resnet1-full} and \ref{tab:Taskonomy-resnet2-full} show the attack results for all 15 multi-task models for Tiny-Taskonomy.
Overall, \pjn{} achieves 80\% first place (80 out of 100 cases) on NYUv2 and 88.33\% (53 out of 60) on Tiny-Taskonomy, demonstrating the effectiveness of adversarial samples from \pjn{}. 

We also show the evaluation results with perturbation bound $\epsilon=4$ in Tables~\ref{tab:NYUv2-resnet1-full-4} and \ref{tab:NYUv2-resnet2-full-4} for NYUv2 and Tables~\ref{tab:Taskonomy-resnet1-full-4} and \ref{tab:Taskonomy-resnet2-full-4} for Taskonomy.

We further compare \pjn{} with a variant of \textsc{Total} called \textsc{WeightedTOTAL}, which uses grid search to determine optimal task-specific weights when generating adversarial samples. 
Specifically, each task-specific gradient is assigned a weight selected from the range 0.1 to 0.9 in increments of 0.1, with the sum of the weights constrained to 1. \textsc{WeightedTOTAL} is integrated with PGD and APGD for comparison against \pjn{}.

Table \ref{tab:NYUv2-resnet-WeightTotal} reports the optimal weights and the ARP achieved by the \textsc{WeightedTOTAL} attack on all 25 multi-task models trained on NYUv2 with Deeplab-ResNet34, using a perturbation bound of $\epsilon=8$. 
Overall, \pjn{} achieves a higher ARP in 49 out of 50 cases compared to \textsc{WeightedTOTAL}. 
Additionally, it is noteworthy that \textsc{WeightedTOTAL} incurs significant computational overhead due to its grid search process, while \pjn{} introduces negligible overhead compared to the original single-task attacks, as it retains the core attack mechanisms with only additional dynamic gradient weighting.

\revise{\subsection{Black-Box Attack on Clean Multi-Task Models}}
We further evaluate the effectiveness of \pjn{} under the black-box scenarios to verify whether adversarial examples generated from a local surrogate model can be effectively transferred to a target black-box model.
Specifically, the experiment is conducted on multi-task models trained on the NYUv2 dataset under the classic transfer-based black-box attack setting:
\begin{itemize}
    \item \textbf{Surrogate model.} We generate adversarial examples using white-box attacks on the all-shared model (i.e., AS) with a fixed perturbation budget of $\epsilon=8$.
    \item \textbf{Target models.} These adversarial samples are then evaluated, \emph{without} any access to gradients, on 7 additional multi-task networks that vary in their parameter-sharing strategies, ranging from fully independent task models (i.e., IND) to partially shared configurations.
\end{itemize}

The results are summarized in Table~\ref{tab:NYUv2-resnet-black}.
We highlight three key observations. 
First, adversarial examples generated via \pjn{} on the surrogate model demonstrate transferability across models with different sharing structures.
In particular, \pjn{} achieves the highest ARP in 18 out of the 21 evaluated settings.
Second, the degree of transferability is closely tied to the architectural similarity between the surrogate and target models. The lowest ARP occurs when transferring to the fully independent models (i.e., IND), which has no shared parameters and is structurally furthest from the all-shared surrogate.
Finally, as expected, we observe a natural performance gap between black-box and white-box scenarios. For instance, the last row of Table~\ref{tab:NYUv2-resnet-black}, corresponding to the white-box attack on the surrogate model itself, shows significantly higher ARP values than the black-box transfer cases. Similar comparisons can be made between Table~\ref{tab:NYUv2-resnet-black} and the Table~\ref{tab:NYUv2-resnet} in the main paper.

\begin{table}[htb]
\caption{ARP (\%, higher-the-better) of 7 multi-task models with diverse sharing patterns trained on \textbf{NYUv2} and attacked by PGD, APGD, and Auto-SAGA variants under \textbf{black-box attack}. The adversarial samples are generated by attacking the all-shared model with perturbation bound $\epsilon=8$. For brevity, the name of \textsc{Single-x} variants are simplified to the task name only. IND: independent, AS: all-shared, BB vs WB: black-box vs white-box} \label{tab:NYUv2-resnet-black}
\centering
\scriptsize
\tabcolsep=0.01cm
\begin{tabular}{cc|c|cccccc|cccccc|cc}
\toprule
\multicolumn{2}{c|}{\multirow{2}{*}{Model}} & \multirow{2}{*}{\#Params (M)} & \multicolumn{6}{c|}{PGD}                                                     & \multicolumn{6}{c|}{APGD}                                                    & \multicolumn{2}{c}{Auto-SAGA}   \\ \cline{4-17} 
\multicolumn{2}{c|}{}                       &                               & Segm  & Norm  & Dept           & Total          & SignTotal & \pjn{}           & Segm  & Norm  & Dept           & Total          & SignTotal & \pjn{}           & Baseline       & \pjn{}           \\ \hline
\multirow{7}{*}{BB}       & IND      & 63.83                         & 23.52 & 11.21 & 20.83          & \textbf{25.57} & 22.85     & 25.14          & 23.55 & 11.38 & 21.23          & \textbf{25.65} & 22.84     & 25.32          & \textbf{25.56} & 25.14          \\
                                 & 5        & 62.48                         & 24.80 & 13.07 & 24.41          & 27.06          & 24.78     & \textbf{27.21} & 24.82 & 13.16 & 24.90          & 27.25          & 24.81     & \textbf{27.46} & 27.10          & \textbf{27.19} \\
                                 & 35       & 62.25                         & 28.71 & 16.92 & 44.55          & 34.36          & 37.06     & \textbf{41.52} & 28.73 & 16.94 & 44.96          & 34.48          & 37.08     & \textbf{41.87} & 34.34          & \textbf{41.56} \\
                                 & 41       & 61.13                         & 30.62 & 18.46 & 44.77          & 36.32          & 39.01     & \textbf{43.44} & 30.77 & 18.57 & 45.35          & 36.51          & 38.92     & \textbf{43.78} & 36.24          & \textbf{43.44} \\
                                 & 21       & 55.65                         & 26.10 & 13.29 & 25.50          & 28.52          & 26.40     & \textbf{28.91} & 26.06 & 13.38 & 25.97          & 28.65          & 26.43     & \textbf{29.27} & 28.56          & \textbf{28.92} \\
                                 & 39       & 55.43                         & 26.47 & 15.63 & 28.32          & 29.38          & 28.16     & \textbf{31.45} & 26.54 & 15.73 & 28.90          & 29.63          & 28.24     & \textbf{31.83} & 29.45          & \textbf{31.43} \\
                                 & 26       & 42.54                         & 28.69 & 19.80 & 30.70          & 32.07          & 32.80     & \textbf{36.43} & 28.70 & 20.03 & 31.19          & 32.15          & 32.74     & \textbf{36.64} & 32.06          & \textbf{36.46} \\ \hline
WB                        & AS       & 21.28                         & 37.80 & 37.87 & \textbf{78.58} & 49.49          & 64.36     & 69.60          & 44.80 & 49.65 & \textbf{99.32} & 56.27          & 66.05     & 84.44          & 79.27          & \textbf{91.01} \\ \bottomrule
\end{tabular}
\end{table}

\subsection{Attack on Adversarially Trained Models}\label{sect:attack-on-adv-more}
As introduced in Section 4.3 in the main paper, we adopt the single task Friendly Adversarial Training (FAT) \cite{zhang2020attacks} in the MTL context. Specifically, we modify the underlying adversarial samples generation process to the multi-task attacks we investigated in this paper, including naive multi-task adaptations \textsc{Single} and \textsc{Total}, and the proposed \pjn{}. 
The detailed adversarial training algorithm is described in Algorithm \ref{alg:FAT-MTL}, where for each mini-batch, we generate the adversarial samples with \pjn{}-PGD to train the model.

\medskip
\begin{breakablealgorithm}
\label{alg:FAT-MTL}
    \caption{FAT \pjn{}}
    \begin{algorithmic}[1]
        \Require network $f_\theta$, training dataset $S=\{(x^j, \{y^j_i\}_{i=1}^n)\}$
        \Ensure adversarially robust multi-task model
        \For{epoch $= 1, \cdots, T$}
            \For{mini-batch $= 1, \cdots, M$}
                \State Sample a mini-batch $\{(x^j, \{y^j_i\})\}$ from $S$
                \For{$j=1,\cdots,b$}
                    \State Obtain adversarial data $\hat{x}^j$ of $x^j$ by \pjn{}-PGD
                \EndFor
            \EndFor
            \State Update model $\theta$ with the adversarial samples
        \EndFor
    \end{algorithmic}
\end{breakablealgorithm}
\medskip

Table \ref{tab:adv_clean_PGD_no_ori} reports the accuracy of multi-task models trained on NYUv2 with adversarial training. 
It includes metrics for all tasks and performance degradation (shown in columns containing ARP). 
We employ PGD-based adapted multi-task attacks \textsc{Single-x} and \textit{Total} as well as \pjn{} with $\epsilon = 8$ to generate different adversarial data.
It can be seen that after adversarial training, the average accuracy of the multi-task model is dropped by $13.75\%\sim16.53\%$ compared with training with clean data only (the ``w/o AT'' row).

The same phenomenon of decreased model accuracy and increased model robustness can also be observed in Tables \ref{tab:adv_clean_FGSM} and \ref{tab:adv_ARP_FGSM}, where FGSM-based multi-task attack variants are utilized when generating adversarial samples in adversarial training.
Table \ref{tab:adv_clean_FGSM} reports the accuracy of the adversarially trained multi-task models similar to Table \ref{tab:adv_clean_PGD_no_ori}, while Table \ref{tab:adv_ARP_FGSM} shows the ARP of multi-task models trained with and without FAT in Table \ref{tab:adv_clean_FGSM} and attacked by multi-task attacks including FGSM variants of \pjn{}, \textsc{Single}, and \textsc{Total} with $\epsilon=8$.

We first observe a $3.58\%\sim5.54\%$ accuracy drop with adversarial training from Table \ref{tab:adv_clean_FGSM}, that is, decreased model performance.
Then we observe an increase in model robustness after adversarial training from the lower ARP after the attack reported in Table \ref{tab:adv_ARP_FGSM}. For instance, when attacked by \pjn{}, ARP decreased from $39.40\%$ to $9.65\%\sim17.09\%$.
Furthermore, \pjn{} remains the most effective attack method, consistently outperforming the \textsc{Single-x} and \textsc{Total} baselines by up to 6.87\%.

\subsection{Parameter Sharing and Model Robustness}\label{sect:param-sharing-more}
Table \ref{tab:shareVSrobust} reports the numerical results for Figure 5 in the main paper, including the results for six multi-task models attacked by APGD \textsc{Single-x} variants.
These models represent six levels of parameter sharing, ranging from all-share (AS/5L), where all five layers of the backbone model (5L) are shared, to independent models (IND/0L), where no layers are shared.
We observe a distinct trend where the attack transferability decreases along with the reduction in levels of parameter sharing, irrespective of the specific attack method utilized.

\begin{table}[htb]
    \caption{The numerical results for the attack transferability of six multi-task models with various levels of parameter sharing attacked by APGD \textsc{Single-x} variants.} \label{tab:shareVSrobust}
    \centering
    \tabcolsep=0.05cm
    \begin{tabular}{c|ccc}
    \toprule
           & Single-Segm & Single-Norm & Single-Dept \\ \midrule
    AS/5L  & 0.46 & 0.53 & 0.15 \\
    4L     & 0.26 & 0.45 & 0.15 \\
    3L     & 0.22 & 0.33 & 0.12 \\
    2L     & 0.16 & 0.28 & 0.09 \\
    1L     & 0.12 & 0.31 & 0.08 \\
    IND/0L & 0.02 & 0.17 & 0.1    \\   \bottomrule
    \end{tabular}
\end{table}

\subsection{Visualization Results}\label{sect:vis}
\begin{figure}[htb]
  \centering
  \begin{subfigure}{0.95\textwidth}
  \includegraphics[width=\linewidth]{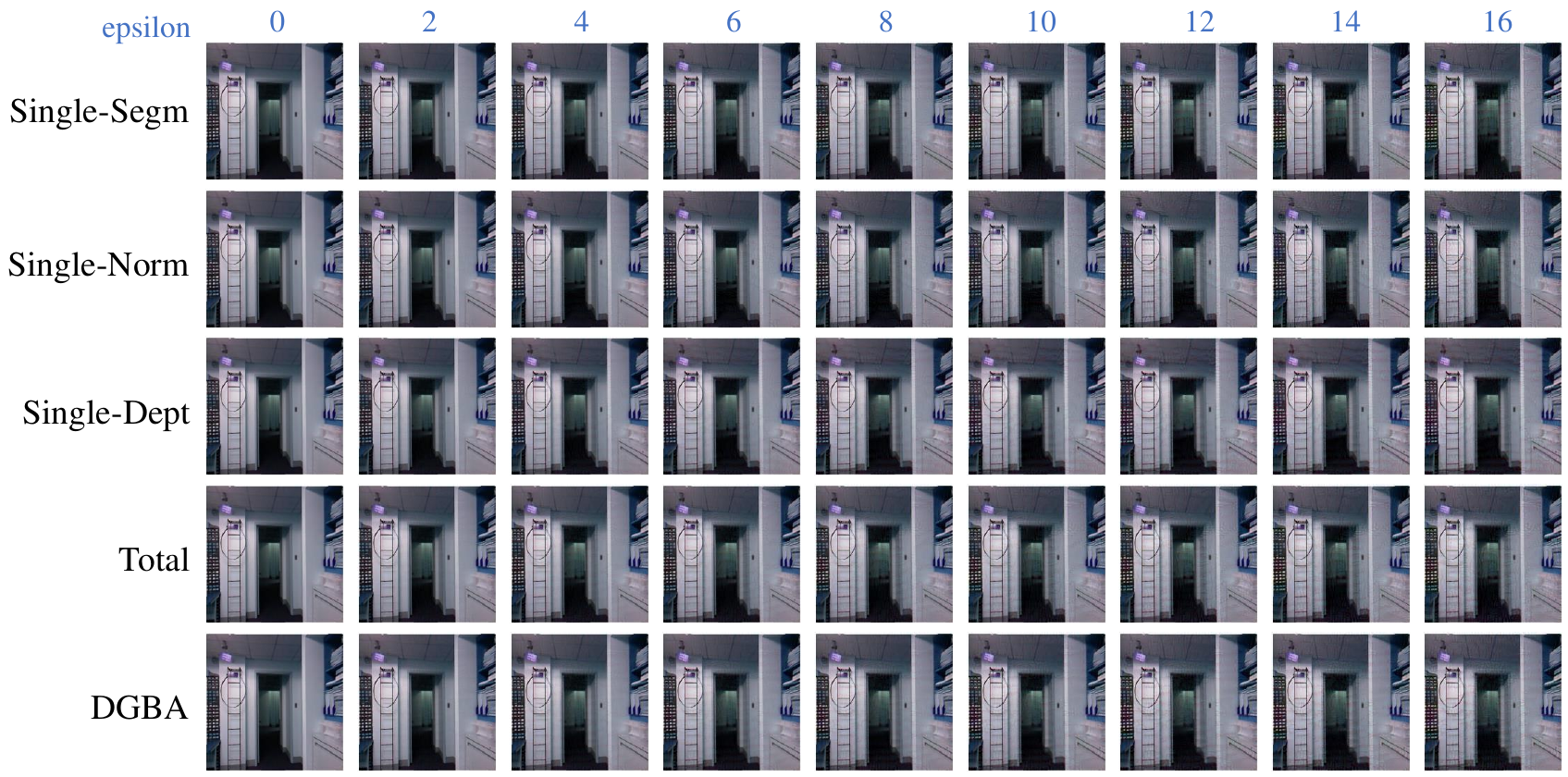}
  \caption{NYUv2}
  \end{subfigure}
  \hfill
  \begin{subfigure}{0.95\textwidth}
  \includegraphics[width=\linewidth]{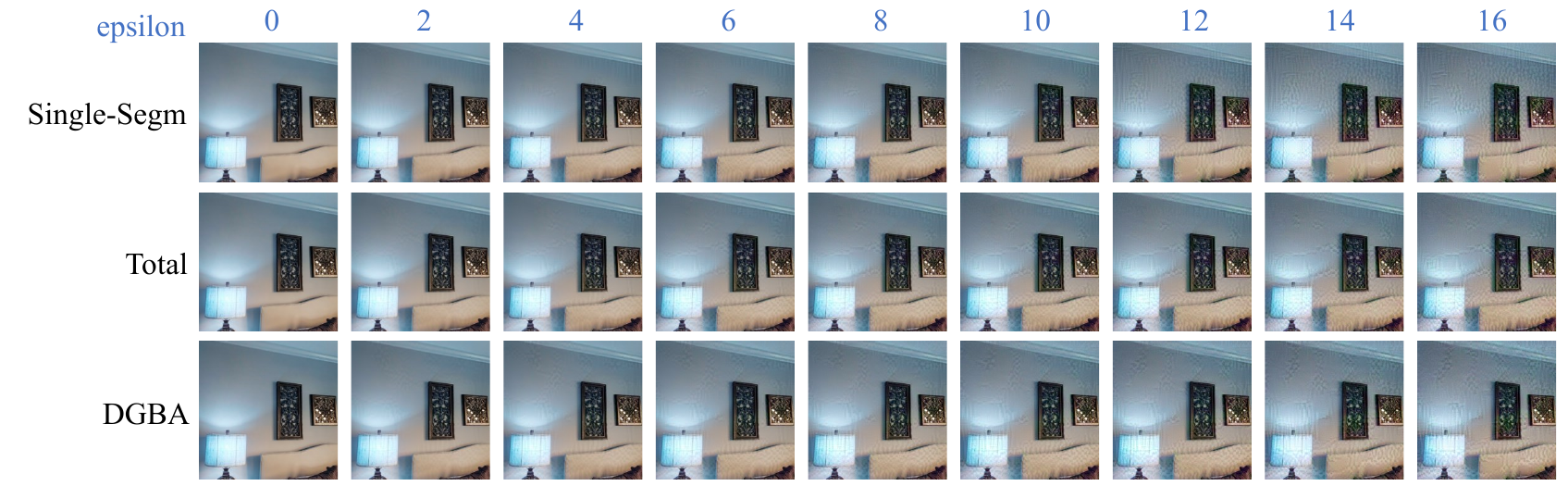}
  \caption{Taskonomy}
  \end{subfigure}
  \caption{Visualization for adversarial samples of one image from (a) NYUv2 and (b) Taskonomy with different attack methods including \textsc{Single-x} attack, \textsc{total} attack, and the proposed \pjn{} and various attack strength $\epsilon$ from 0 to 16.} \label{fig:visulization}
\end{figure}

Figure \ref{fig:visulization} visualizes adversarial samples generated given an image from (a) NYUv2 and (b) Taskonomy with different attack methods including \textsc{Single-x} attack, \textsc{total} attack, and the proposed \pjn{} and various attack strengths $\epsilon$ from 0 to 16. 
We show that along with the increase in the perturbation bound $\epsilon$, the magnitude of the noise becomes larger and more visible. All attacks become more effective and thus lead to similar attack performance as shown in Figures 1 and 2 of the main paper.

\subsection{Multi-Task Architectures with Model Index}\label{sect:multi-task-models}
We show the effectiveness of the proposed multi-task attack \pjn{} by conducting attack experiments on multi-task models with different sharing levels in the main paper.
Tables \ref{tab:NYUv2-layouts} and \ref{tab:Taskonomy-layouts} present the multi-task model architectures in the Layout format proposed by TreeMTL \cite{zhang2022tree}. 
A layout is a symbolized representation of a tree-structured multi-task architecture. 
For $T$ tasks and a backbone model with $B$ branching points, a layout $\mathbf{L}=[L_1,L_2,\cdots,L_B]$,
where $L_i$ is a list of task sets at the $i$-th branching point. 
Task sets in $L_i = [L_i^1, L_i^2, \cdots]$ are subsets of tasks $\mathcal{T}$ and a task set $L_i^p$ means the set of tasks in $ L_i^p$ sharing the $i$-th block. 

\begin{table}[t]
\caption{ARP of all 25 multi-task models with diverse sharing patterns trained on NYUv2 and attacked by FGSM and PGD variants with perturbation bound $\epsilon=8$. For brevity, the name of \textsc{Single-x} variants are simplified to the task name only. IND: independent, AS: all-shared.} \label{tab:NYUv2-resnet1-full}
\centering
\scriptsize
\tabcolsep=0.02cm
\begin{tabular}{c|c|cccccc|cccccc}
\toprule
\multirow{2}{*}{\begin{tabular}[c]{@{}c@{}}Model \\ Index\end{tabular}} & \multirow{2}{*}{\#Params (M)} & \multicolumn{6}{c|}{FGSM}                                              & \multicolumn{6}{c}{PGD}                 \\ \cline{3-14}
                                                                             &                               & Segm  & Norm  & Dept            & Total & SignTotal & \pjn{}           & Segm  & Norm  & Dept            & Total & SignTotal & \pjn{}             \\ \hline
IND                    & 63.83                          & 21.56    & 16.97    & 35.24                                 & 33.47 & { 38.1}  & { \textbf{39.07}} & 28.18    & 18.23    & 62.14                                  & 50.52 & { 67.85}  & { \textbf{73.41}}  \\
14                      & 63.6                           & 21.72    & 27.07    & 35.96                                 & 38.51 & { 46.54} & { \textbf{46.69}} & 29.54    & 44.67    & 63.07                                  & 63.09 & { 92.95}  & { \textbf{98.9}}   \\
4                       & 63.6                           & 21.12    & 17.09    & 39.01                                 & 38.32 & { 43.45} & { \textbf{44.22}} & 27.92    & 21.74    & 63.88                                  & 59.03 & { 72.88}  & { \textbf{79.05}}  \\
9                       & 63.6                           & 26.2     & 21.63    & 38.79                                 & 37.72 & { 40.91} & { \textbf{43.21}} & 33.56    & 19.35    & 83.54                                  & 70.83 & { 85.73}  & { \textbf{92.65}}  \\
23                      & 63.59                          & 26.18    & 23.64    & 40.69                                 & 37.52 & { 41.52} & { \textbf{43.8}}  & 33.2     & 26.47    & 80.66                                  & 65.38 & { 84.67}  & { \textbf{90.43}}  \\
5                       & 62.48                          & 23.19    & 22.05    & 39.14                                 & 38.54 & { 39.67} & { \textbf{42.99}} & 29.12    & 30       & 70.99                                  & 61.4  & { 71.46}  & { \textbf{79.6}}   \\
10                      & 62.48                          & 26.1     & 19.72    & 39.59                                 & 34.76 & { 39.99} & { \textbf{41.57}} & 37.65    & 21.7     & { \textbf{97.24}}  & 60.91 & 86.69                         & { 93.55}           \\
28                      & 62.47                          & 26.12    & 28.27    & 40.01                                 & 38.67 & { 42.65} & { \textbf{44.93}} & 33.36    & 49.64    & 89.3                                   & 72.06 & { 101.31} & { \textbf{107.59}} \\
35                      & 62.25                          & 25.12    & 24.02    & { 37.27}          & 33.53 & 36.93                        & { \textbf{39.73}} & 36.46    & 32.54    & { 98.11}           & 68.3  & 91.24                         & { \textbf{100.53}} \\
38                      & 62.25                          & 25.94    & 26.21    & 35.87                                 & 34.49 & { 37.22} & { \textbf{39.52}} & 34.42    & 43.02    & 83.04                                  & 63.46 & { 87.27}  & { \textbf{93.76}}  \\
41                      & 61.13                          & 26.08    & 26.38    & { 40.88}          & 36.12 & 39.79                        & { \textbf{42.64}} & 38.06    & 46.02    & { 100.41}          & 67.51 & 95.2                          & { \textbf{103.03}} \\
11                      & 55.66                          & 28.75    & 20.65    & { 39.53}          & 35.69 & 39.38                        & { \textbf{41.37}} & 38.52    & 18.99    & { \textbf{93.81}}  & 57.74 & 82.58                         & { 89.68}           \\
21                      & 55.65                          & 25.27    & 23.64    & { 42.66}          & 39.93 & 40.26                        & { \textbf{44.65}} & 31.45    & 39.55    & { 79.1}            & 65    & 76.7                          & { \textbf{84.81}}  \\
33                      & 55.43                          & 27.67    & 25.99    & { 38.89}          & 36.26 & 36.67                        & { \textbf{40.35}} & 33.67    & 35.08    & { \textbf{87.47}}  & 65.02 & 77.33                         & { 87.08}           \\
36                      & 55.43                          & 27.8     & 29.84    & 37.76                                 & 34.08 & { 39.49} & { \textbf{41.36}} & 38.36    & 40.88    & { 96.69}           & 57.78 & 91.97                         & { \textbf{99.72}}  \\
39                      & 55.43                          & 29.97    & 30.95    & 40.68                                 & 39.95 & { 41.62} & { \textbf{44.76}} & 36.03    & 45.68    & 83.78                                  & 67.83 & { 90.4}   & { \textbf{96.63}}  \\
44                      & 54.32                          & 25.99    & 25.9     & { 39.41}          & 34.13 & 37.99                        & { \textbf{40.86}} & 40.48    & 43.64    & { \textbf{104.05}} & 63.19 & 93.5                          & { 100.56}          \\
42                      & 54.32                          & 28.27    & 29.47    & 38.55                                 & 36.15 & { 39.21} & { \textbf{41.42}} & 36.9     & 52.1     & { 91.33}           & 60.92 & 89.17                         & { \textbf{95.79}}  \\
48                      & 47.5                           & 29.01    & 28.54    & { 38.05}          & 34.44 & 36.5                         & { \textbf{38.38}} & 39.88    & 48.02    & { \textbf{92.03}}  & 58.34 & 84.81                         & { 90.34}           \\
26                      & 42.54                          & 25.47    & 23.76    & 38.8                                  & 37.75 & { 38.87} & { \textbf{41.97}} & 34.19    & 45.39    & 80.94                                  & 69.39 & { 88.58}  & { \textbf{93.88}}  \\
17                      & 42.54                          & 25.5     & 25.78    & { 41.79}          & 39.93 & 39.61                        & { \textbf{44.12}} & 32.21    & 41.36    & { 79.58}           & 66.8  & 74.34                         & { \textbf{84.75}}  \\
34                      & 42.32                          & 27.11    & 24.73    & { 35.06}          & 31.75 & 34.69                        & { \textbf{36.48}} & 41.27    & 30.49    & { \textbf{86.88}}  & 52.77 & 77                            & { 82.03}           \\
43                      & 41.21                          & 28.1     & 27.35    & { 39.21}          & 34.18 & 37.55                        & { \textbf{39.68}} & 42.74    & 44.65    & { \textbf{104.83}} & 58.03 & 90.72                         & { 95.97}           \\
49                      & 34.39                          & 28.23    & 28.02    & { \textbf{39.85}} & 34.31 & 37.66                        & { 39.55}          & 44.96    & 60.65    & { \textbf{119.52}} & 59.28 & 98.39                         & { 106.54}          \\
AS             & 21.28                          & 29.5     & 28.31    & { 38.91}  & 34.71 & 36.66                        & { \textbf{39.4}}          & 46.65    & 53.36    & { \textbf{105.74}} & 58.79 & 83.56                         & { 88.51}           \\ \bottomrule
\end{tabular}
\end{table}

\begin{table}[htb]
\caption{ARP of all 25 multi-task models similar as Table \ref{tab:NYUv2-resnet1-full} for \pjn{} and the two baselines, APGD and Auto-SAGA.
} \label{tab:NYUv2-resnet2-full}
\centering 
\scriptsize
\tabcolsep=0.02cm
\begin{tabular}{c|c|cccccc|cc}
\toprule
\multirow{2}{*}{\begin{tabular}[c]{@{}c@{}}Model \\ Index\end{tabular}} & \multirow{2}{*}{\#Params (M)} & \multicolumn{6}{c|}{APGD}                                              & \multicolumn{2}{c}{Auto-SAGA}                 \\ \cline{3-10}
                                                                             &                               & Segm  & Norm  & Dept            & Total & SignTotal & \pjn{}           & Baseline & \pjn{}             \\ \hline
IND                    & 63.83                          & 30.74    & 23.52    & { 75.62}           & 58.3  & 69.29                        & { \textbf{87.58}}  & 71.07    & { \textbf{88.53}}  \\
14                      & 63.6                           & 32.43    & 52.73    & 80.01                                  & 73.77 & { 95.69} & { \textbf{118.63}} & 94.14    & { \textbf{118.97}} \\
4                       & 63.6                           & 30.62    & 30.53    & { 76.81}           & 67.7  & 73.99                        & { \textbf{93.73}}  & 77.63    & { \textbf{94.41}}  \\
9                       & 63.6                           & 37.53    & 25.86    & { 100.49}          & 80.34 & 88.5                         & { \textbf{108.6}}  & 102.27   & { \textbf{110.24}} \\
23                      & 63.59                          & 37       & 32.9     & { 98.26}           & 73.23 & 87.13                        & { \textbf{107.08}} & 102.43   & { \textbf{110.87}} \\
5                       & 62.48                          & 32.06    & 39.25    & { 88.03}           & 70.34 & 73.05                        & { \textbf{94.44}}  & 84.51    & { \textbf{95.7}}   \\
10                      & 62.48                          & 41.62    & 28.69    & { \textbf{122.36}} & 67.27 & 89.72                        & { 113.78}          & 110.51   & { \textbf{119.37}} \\
28                      & 62.47                          & 37.28    & 59.37    & { 113.04}          & 83.65 & 105.58                       & { \textbf{130.93}} & 118.33   & { \textbf{133.16}} \\
35                      & 62.25                          & 41.01    & 41.71    & { 119.52}          & 78.1  & 95.22                        & { \textbf{121.3}}  & 124.97   & { \textbf{131.56}} \\
38                      & 62.25                          & 38.3     & 54.74    & { 106.5}           & 71.85 & 90.27                        & { \textbf{113.04}} & 105.82   & { \textbf{117.23}} \\
41                      & 61.13                          & 42.09    & 59.54    & { 120.39}          & 75.42 & 99.05                        & { \textbf{122.5}}  & 113.15   & { \textbf{126.94}} \\
11                      & 55.66                          & 43.1     & 24.76    & { \textbf{112.77}} & 63.12 & 84.97                        & { 106.21}          & 107.85   & { \textbf{115.3}}  \\
21                      & 55.65                          & 34.8     & 57.48    & { 96.28}           & 74.13 & 78.75                        & { \textbf{100.94}} & 92.91    & { \textbf{102.72}} \\
33                      & 55.43                          & 37.37    & 44.63    & { \textbf{110.04}} & 74.07 & 79.81                        & { 104.87}          & 109.74   & { \textbf{112.81}} \\
36                      & 55.43                          & 43       & 50.84    & { 117.13}          & 63.6  & 95.66                        & { \textbf{119.98}} & 112.12   & { \textbf{127.46}} \\
39                      & 55.43                          & 39.81    & 56.61    & { 106.07}          & 76.16 & 94.08                        & { \textbf{115.62}} & 108.61   & { \textbf{118.71}} \\
44                      & 54.32                          & 45.29    & 54.94    & { \textbf{128.01}} & 68.93 & 92.54                        & { 120.52}          & 125.6    & { \textbf{130.93}} \\
42                      & 54.32                          & 41.34    & 67.03    & { \textbf{116.01}} & 67.46 & 92.41                        & { 114.74}          & 113.28   & { \textbf{120.97}} \\
48                      & 47.5                           & 44.42    & 61.56    & { \textbf{111.9}}  & 62.92 & 84.19                        & { 106.66}          & 107.63   & { \textbf{114.49}} \\
26                      & 42.54                          & 38.32    & 58.72    & { 101.69}          & 80.18 & 92.38                        & { \textbf{115.79}} & 107.18   & { \textbf{118.67}} \\
17                      & 42.54                          & 35.91    & 51.57    & { 98.39}           & 75.5  & 76.19                        & { \textbf{98.4}}   & 100.51   & { \textbf{103.25}} \\
34                      & 42.32                          & 45.56    & 38.38    & { \textbf{108.03}} & 55.82 & 79.52                        & { 96.55}           & 95.69    & { \textbf{105.38}} \\
43                      & 41.21                          & 47.44    & 59.7     & { \textbf{131.97}} & 61.11 & 89.67                        & { 115.1}           & 122.87   & { \textbf{128.99}} \\
49                      & 34.39                          & 48.4     & 70.91    & { \textbf{136.53}} & 61.73 & 92.81                        & { 119.81}          & 125.91   & { \textbf{130.8}}  \\
AS             & 21.28                          & 56.39    & 69.6     & { \textbf{133.54}} & 67.42 & 88.1                         & { 108.57}          & 112.56   & { \textbf{119.22}} \\ \bottomrule
\end{tabular}
\end{table}

\begin{table}[htb]
\caption{ARP of all 15 multi-task models with diverse sharing patterns trained on Tiny-Taskonomy and attacked by FGSM and PGD variants with perturbation bound $\epsilon=8$. For brevity, the name of \textsc{Single-x} variants are simplified to the task name only. IND: independent, AS: all-shared.} \label{tab:Taskonomy-resnet1-full}
\centering
\scriptsize
\tabcolsep=0.02cm
\begin{tabular}{c|c|ccccccc|ccccccc}
\toprule
\multirow{2}{*}{\begin{tabular}[c]{@{}c@{}}Model \\ Index\end{tabular}} & \multirow{2}{*}{\#Params (M)} & \multicolumn{7}{c|}{FGSM}                                               & \multicolumn{7}{c}{PGD}                                                     \\ \cline{3-16} 
                                                                             &                               & Segm           & Norm  & Dept  & Keyp  & Edge  & Total & \pjn{}          & Segm   & Norm  & Dept   & Keyp  & Edge  & Total           & \pjn{}           \\ \hline
IND                                                                         & 106.38                        & 85.42          & 18.71 & 51.99 & 11.09 & 11.96 & 88.26 & \textbf{95.10} & 248.04 & 23.13 & 117.44 & 12.48 & 11.50 & \textbf{282.01} & 274.17          \\
190                                                                          & 105.03                        & 73.19          & 23.13 & 42.67 & 13.75 & 16.55 & 72.87 & \textbf{74.15} & 190.91 & 49.41 & 138.67 & 14.98 & 14.02 & 232.55          & \textbf{236.63} \\
348                                                                          & 105.02                        & 67.60          & 23.05 & 49.61 & 14.74 & 14.86 & 70.57 & \textbf{78.14} & 202.55 & 51.07 & 147.26 & 19.62 & 14.74 & 246.88          & \textbf{249.54} \\
200                                                                          & 104.81                        & 67.83          & 19.17 & 44.49 & 12.39 & 14.55 & 71.01 & \textbf{76.78} & 182.76 & 50.02 & 139.60 & 13.13 & 16.56 & 226.76          & \textbf{233.80} \\
352                                                                          & 104.80                        & 67.17          & 22.59 & 44.80 & 15.19 & 14.85 & 68.88 & \textbf{74.70} & 178.64 & 48.81 & 143.03 & 17.71 & 13.70 & 224.02          & \textbf{234.34} \\
469                                                                          & 104.57                        & 72.08          & 22.52 & 49.17 & 17.73 & 15.17 & 73.25 & \textbf{79.48} & 182.49 & 49.82 & 155.29 & 27.55 & 18.20 & 228.44          & \textbf{240.80} \\
358                                                                          & 103.68                        & 69.37          & 23.91 & 46.00 & 15.62 & 15.60 & 69.91 & \textbf{74.85} & 189.90 & 50.42 & 152.70 & 18.14 & 15.53 & 237.12          & \textbf{247.29} \\
481                                                                          & 103.46                        & 72.21          & 22.81 & 46.38 & 21.68 & 16.24 & 74.59 & \textbf{81.94} & 212.18 & 51.14 & 157.02 & 38.94 & 24.78 & 256.71          & \textbf{266.46} \\
191                                                                          & 98.21                         & 68.01          & 20.88 & 40.81 & 12.19 & 13.79 & 68.19 & \textbf{69.76} & 190.28 & 52.63 & 153.36 & 13.94 & 12.94 & 226.54          & \textbf{231.65} \\
959                                                                          & 96.86                         & 69.48          & 23.28 & 44.76 & 21.98 & 18.80 & 73.29 & \textbf{76.47} & 201.59 & 48.17 & 163.56 & 15.56 & 17.64 & 241.75          & \textbf{256.35} \\
958                                                                          & 83.75                         & 74.68          & 23.46 & 47.23 & 22.66 & 16.49 & 80.55 & \textbf{84.18} & 210.33 & 50.29 & 166.25 & 15.22 & 13.20 & 250.88          & \textbf{266.38} \\
1020                                                                         & 83.53                         & 74.85          & 22.43 & 48.32 & 20.34 & 16.31 & 80.78 & \textbf{86.30} & 222.92 & 49.98 & 139.97 & 18.25 & 18.01 & 259.27          & \textbf{263.03} \\
1043                                                                         & 75.59                         & 77.23          & 21.55 & 53.74 & 22.06 & 23.22 & 84.89 & \textbf{94.04} & 205.27 & 44.26 & 145.47 & 23.99 & 38.96 & 246.90          & \textbf{250.79} \\
1037                                                                         & 62.48                         & 72.67          & 22.63 & 48.99 & 22.62 & 17.82 & 77.62 & \textbf{84.68} & 216.15 & 45.51 & 152.47 & 17.61 & 47.35 & 252.55          & \textbf{261.82} \\
AS                                                                  & 21.28                         & \textbf{76.86} & 45.37 & 54.92 & 21.83 & 21.60 & 76.24 & 76.03          & 231.23 & 77.53 & 104.10 & 16.02 & 22.40 & \textbf{231.08} & 196.07          \\ \bottomrule
\end{tabular}
\end{table}

\begin{table}[htb]
\caption{ARP of all 15 multi-task models similar as Table \ref{tab:Taskonomy-resnet1-full}. The fundamental attack methods are APGD and Auto-SAGA.} \label{tab:Taskonomy-resnet2-full}
\centering
\scriptsize
\tabcolsep=0.05cm
\begin{tabular}{c|c|ccccccc|cc}
\toprule
\multirow{2}{*}{Model Index} & \multirow{2}{*}{\#Params (M)} & \multicolumn{7}{c|}{APGD}                                                      & \multicolumn{2}{c}{Auto-SAGA}     \\ \cline{3-11} 
                                  &                               & Segm            & Norm  & Dept   & Keyp  & Edge  & Total           & \pjn{}           & Baseline        & \pjn{}           \\ \hline
IND                              & 106.38                        & 245.25          & 23.21 & 117.83 & 12.40 & 11.37 & \textbf{279.28} & 270.25 & \textbf{282.01} & 274.17          \\
190                               & 105.03                        & 175.02          & 45.06 & 130.69 & 14.26 & 13.64 & 213.47          & \textbf{218.29} & 216.32          & \textbf{218.88} \\
348                               & 105.02                        & 184.25          & 46.19 & 137.71 & 18.80 & 13.90 & 225.45          & \textbf{229.03} & 228.10          & \textbf{229.64} \\
200                               & 104.81                        & 168.93          & 44.77 & 130.06 & 12.55 & 15.74 & 208.48          & \textbf{215.48} & 210.54          & \textbf{216.37} \\
352                               & 104.80                        & 164.72          & 44.09 & 134.15 & 16.84 & 13.15 & 205.59          & \textbf{216.01} & 207.59          & \textbf{216.50} \\
469                               & 104.57                        & 168.58          & 44.55 & 143.19 & 26.30 & 17.31 & 210.28          & \textbf{220.91} & 212.21          & \textbf{221.93} \\
358                               & 103.68                        & 174.80          & 46.32 & 142.73 & 17.07 & 14.73 & 217.99          & \textbf{227.61} & 220.93          & \textbf{228.97} \\
481                               & 103.46                        & 193.75          & 45.87 & 146.31 & 37.71 & 21.32 & 234.60          & \textbf{244.78} & 236.99          & \textbf{246.22} \\
191                               & 98.21                         & 174.37          & 47.96 & 142.62 & 13.05 & 12.25 & 208.24          & \textbf{213.53} & 211.71          & \textbf{214.30} \\
959                               & 96.86                         & 183.75          & 44.46 & 150.37 & 14.70 & 17.04 & 221.31          & \textbf{234.58} & 223.54          & \textbf{234.56} \\
958                               & 83.75                         & 192.26          & 45.68 & 152.92 & 14.37 & 12.80 & 229.57          & \textbf{242.97} & 232.22          & \textbf{243.31} \\
1020                              & 83.53                         & 202.50          & 44.34 & 130.95 & 17.02 & 16.89 & 235.80          & \textbf{240.26} & 238.95          & \textbf{241.14} \\
1043                              & 75.59                         & 189.32          & 39.97 & 136.29 & 22.55 & 36.50 & 227.23          & \textbf{233.25} & 230.54          & \textbf{234.28} \\
1037                              & 62.48                         & 197.99          & 41.05 & 141.10 & 16.88 & 40.50 & 230.86          & \textbf{240.70} & 240.00          & \textbf{246.43} \\
AS                       & 21.28                         & \textbf{227.99} & 76.06 & 103.84 & 15.97 & 21.87 & 226.97          & 192.85          & \textbf{231.07} & 196.07          \\ \bottomrule
\end{tabular}
\end{table}

\begin{table}[htb]
\caption{ARP of all 25 multi-task models with diverse sharing patterns trained on \textbf{NYUv2} and attacked by FGSM and PGD variants with perturbation bound $\epsilon=4$. For brevity, the name of \textsc{Single-x} variants are simplified to the task name only. IND: independent, AS: all-shared.} \label{tab:NYUv2-resnet1-full-4}
\centering  
\scriptsize
\tabcolsep=0.05cm
\begin{tabular}{c|c|cccccc|cccccc}
\toprule
\multirow{2}{*}{\begin{tabular}[c]{@{}c@{}}Model \\ Index\end{tabular}} & \multirow{2}{*}{\#Params (M)} & \multicolumn{6}{c|}{FGSM}                                              & \multicolumn{6}{c}{PGD}                 \\ \cline{3-14}
                                                                             &                               & Segm  & Norm  & Dept            & Total & SignTotal & \pjn{}           & Segm  & Norm  & Dept            & Total & SignTotal & \pjn{}             \\ \hline
IND & 63.83 & 17.09 & 9.31  & 25.73 & 27.45 & 31.47 & \textbf{33.15} & 24.07 & 12.34 & 39.67          & 37.64 & 46.06 & \textbf{52.51} \\
14  & 63.60 & 19.73 & 10.82 & 33.63 & 33.88 & 36.30 & \textbf{39.51} & 27.67 & 12.33 & 64.64          & 57.08 & 63.67 & \textbf{73.42} \\
4   & 63.60 & 17.57 & 20.65 & 27.17 & 33.05 & 40.21 & \textbf{41.11} & 25.28 & 35.17 & 42.23          & 47.76 & 66.83 & \textbf{73.81} \\
9   & 63.60 & 16.99 & 9.37  & 29.26 & 32.13 & 35.28 & \textbf{37.64} & 23.91 & 13.30 & 44.03          & 44.93 & 51.26 & \textbf{58.82} \\
23  & 63.59 & 19.59 & 13.92 & 32.88 & 32.01 & 35.67 & \textbf{38.01} & 27.56 & 17.54 & 61.19          & 52.80 & 62.37 & \textbf{70.10} \\
5   & 62.48 & 17.74 & 12.82 & 30.67 & 32.45 & 33.11 & \textbf{37.15} & 24.51 & 17.33 & 48.98          & 46.82 & 51.18 & \textbf{59.36} \\
10  & 62.48 & 21.43 & 10.35 & 34.04 & 30.76 & 35.13 & \textbf{37.51} & 31.32 & 13.71 & 73.02          & 50.90 & 64.28 & \textbf{73.08} \\
28  & 62.47 & 20.33 & 20.16 & 33.82 & 35.17 & 39.46 & \textbf{41.95} & 27.68 & 35.74 & 63.97          & 56.20 & 73.24 & \textbf{80.54} \\
35  & 62.25 & 20.25 & 18.54 & 30.89 & 30.84 & 34.10 & \textbf{37.00} & 27.91 & 29.93 & 58.75          & 50.02 & 64.14 & \textbf{71.00} \\
38  & 62.25 & 19.56 & 15.72 & 33.75 & 31.01 & 34.50 & \textbf{37.92} & 29.23 & 22.82 & 74.39          & 54.53 & 66.51 & \textbf{76.85} \\
41  & 61.13 & 21.00 & 19.33 & 34.84 & 32.15 & 36.69 & \textbf{39.67} & 31.10 & 32.23 & 75.81          & 54.53 & 71.75 & \textbf{79.93} \\
11  & 55.66 & 22.12 & 10.87 & 33.89 & 30.51 & 33.73 & \textbf{36.88} & 31.73 & 12.76 & 70.28          & 47.48 & 60.24 & \textbf{68.96} \\
21  & 55.65 & 18.96 & 15.58 & 34.51 & 34.17 & 34.54 & \textbf{39.08} & 26.53 & 23.81 & 56.20          & 50.44 & 56.26 & \textbf{64.85} \\
33  & 55.43 & 21.55 & 18.91 & 31.83 & 29.54 & 34.62 & \textbf{37.13} & 31.01 & 28.48 & 72.41          & 47.68 & 66.77 & \textbf{74.71} \\
36  & 55.43 & 21.64 & 21.18 & 33.84 & 34.08 & 36.15 & \textbf{39.67} & 29.49 & 32.03 & 62.09          & 54.67 & 67.51 & \textbf{74.64} \\
39  & 55.43 & 19.99 & 16.21 & 31.60 & 30.32 & 31.06 & \textbf{35.11} & 27.57 & 24.14 & 62.35          & 50.84 & 57.22 & \textbf{66.48} \\
44  & 54.32 & 20.61 & 19.00 & 34.79 & 30.57 & 35.03 & \textbf{38.05} & 33.02 & 30.36 & \textbf{76.81} & 52.04 & 69.01 & 76.70          \\
42  & 54.32 & 20.61 & 20.92 & 32.00 & 30.52 & 34.49 & \textbf{37.13} & 30.17 & 36.93 & 66.13          & 49.23 & 66.79 & \textbf{73.41} \\
48  & 47.50 & 22.34 & 20.88 & 33.26 & 29.65 & 33.05 & \textbf{35.71} & 32.75 & 33.81 & 70.45          & 48.91 & 65.49 & \textbf{71.34} \\
26  & 42.54 & 19.16 & 17.23 & 32.22 & 33.51 & 35.20 & \textbf{38.65} & 28.21 & 32.11 & 58.06          & 54.17 & 63.98 & \textbf{71.52} \\
17  & 42.54 & 19.70 & 17.81 & 34.21 & 35.10 & 34.35 & \textbf{39.69} & 26.89 & 28.06 & 57.05          & 52.25 & 56.21 & \textbf{65.35} \\
34  & 42.32 & 22.45 & 15.94 & 30.89 & 28.23 & 31.25 & \textbf{33.59} & 33.91 & 21.11 & \textbf{64.29} & 44.38 & 57.97 & 63.58          \\
43  & 41.21 & 22.92 & 19.65 & 34.96 & 30.41 & 34.54 & \textbf{37.20} & 35.07 & 30.38 & \textbf{78.53} & 49.62 & 68.22 & 74.39          \\
49  & 34.39 & 22.75 & 21.15 & 35.46 & 30.18 & 34.34 & \textbf{36.90} & 37.63 & 43.02 & \textbf{86.58} & 51.52 & 73.98 & 81.24          \\
AS  & 21.28 & 24.21 & 20.95 & 35.75 & 30.65 & 33.22 & \textbf{35.97} & 37.80 & 37.87 & \textbf{78.58} & 49.49 & 64.36 & 69.60           \\ \bottomrule
\end{tabular}
\end{table}

\begin{table}[htb]
\caption{ARP of all 25 multi-task models similar as Table \ref{tab:NYUv2-resnet1-full-4}. The attack methods are APGD and Auto-SAGA variants.} \label{tab:NYUv2-resnet2-full-4}
\centering   
\scriptsize
\tabcolsep=0.05cm
\begin{tabular}{c|c|cccccc|cc}
\toprule
\multirow{2}{*}{\begin{tabular}[c]{@{}c@{}}Model \\ Index\end{tabular}} & \multirow{2}{*}{\#Params (M)} & \multicolumn{6}{c|}{APGD}                                              & \multicolumn{2}{c}{Auto-SAGA}                 \\ \cline{3-10}
                                                                             &                               & Segm  & Norm  & Dept            & Total & SignTotal & \pjn{}           & Baseline & \pjn{}             \\ \hline
IND & 63.83 & 27.16 & 15.28 & 48.07           & 43.27 & 46.68 & \textbf{62.72} & 50.60 & \textbf{63.78}  \\
14  & 63.60 & 31.53 & 15.75 & 77.52           & 65.34 & 64.94 & \textbf{86.42} & 80.49 & \textbf{88.43}  \\
4   & 63.60 & 28.65 & 41.38 & 51.30           & 54.93 & 67.88 & \textbf{88.39} & 66.46 & \textbf{87.21}  \\
9   & 63.60 & 27.10 & 17.16 & 52.03           & 51.10 & 51.84 & \textbf{69.28} & 58.22 & \textbf{70.20}  \\
23  & 63.59 & 31.31 & 21.70 & 74.76           & 59.86 & 63.49 & \textbf{83.53} & 78.71 & \textbf{85.70}  \\
5   & 62.48 & 27.91 & 21.99 & 59.31           & 54.01 & 51.87 & \textbf{70.46} & 62.59 & \textbf{71.73}  \\
10  & 62.48 & 35.73 & 17.45 & \textbf{91.60}  & 57.20 & 65.46 & 88.04          & 81.73 & \textbf{92.48}  \\
28  & 62.47 & 31.55 & 43.78 & 80.85           & 65.45 & 74.95 & \textbf{98.42} & 84.99 & \textbf{97.51}  \\
35  & 62.25 & 32.18 & 37.56 & 72.78           & 57.23 & 65.31 & \textbf{85.37} & 74.98 & \textbf{85.89}  \\
38  & 62.25 & 34.16 & 28.68 & 92.92           & 63.05 & 68.05 & \textbf{94.19} & 92.69 & \textbf{99.99}  \\
41  & 61.13 & 35.40 & 40.66 & 93.92           & 62.34 & 73.43 & \textbf{96.80} & 86.64 & \textbf{100.73} \\
11  & 55.66 & 36.47 & 16.01 & \textbf{87.30}  & 53.81 & 61.46 & 83.26          & 80.89 & \textbf{88.73}  \\
21  & 55.65 & 29.81 & 32.22 & 68.68           & 57.18 & 57.15 & \textbf{76.41} & 69.18 & \textbf{78.34}  \\
33  & 55.43 & 35.83 & 35.63 & 90.57           & 54.05 & 68.32 & \textbf{92.08} & 83.96 & \textbf{96.00}  \\
36  & 55.43 & 33.51 & 39.54 & 77.22           & 61.78 & 68.97 & \textbf{88.82} & 80.06 & \textbf{89.48}  \\
39  & 55.43 & 31.24 & 29.83 & 78.41           & 58.07 & 58.41 & \textbf{79.38} & 78.94 & \textbf{83.61}  \\
44  & 54.32 & 38.22 & 37.76 & \textbf{97.08}  & 58.90 & 67.84 & 94.07          & 90.86 & \textbf{98.39}  \\
42  & 54.32 & 34.23 & 45.68 & 84.02           & 55.78 & 68.21 & \textbf{88.75} & 80.95 & \textbf{89.80}  \\
48  & 47.50 & 37.40 & 43.53 & \textbf{86.64}  & 54.54 & 64.54 & 85.45          & 81.61 & \textbf{89.60}  \\
26  & 42.54 & 32.45 & 41.99 & 71.78           & 62.53 & 65.45 & \textbf{86.41} & 77.90 & \textbf{87.52}  \\
17  & 42.54 & 30.39 & 34.78 & 68.70           & 58.85 & 57.18 & \textbf{76.24} & 74.42 & \textbf{79.05}  \\
34  & 42.32 & 38.77 & 25.92 & \textbf{79.02}  & 49.26 & 59.24 & 75.29          & 70.50 & \textbf{79.46}  \\
43  & 41.21 & 40.38 & 39.16 & \textbf{99.14}  & 54.95 & 67.08 & 90.60          & 88.50 & \textbf{97.06}  \\
49  & 34.39 & 41.65 & 51.01 & \textbf{103.47} & 55.19 & 70.24 & 94.25          & 91.96 & \textbf{99.57}  \\
AS  & 21.28 & 44.80 & 49.65 & \textbf{99.32}  & 56.27 & 66.05 & 84.44          & 79.27 & \textbf{91.01} \\ \bottomrule
\end{tabular}
\end{table}

\begin{table}[htb]
\caption{ARP of all 15 multi-task models with diverse sharing patterns trained on \textbf{Tiny-Taskonomy} and attacked by FGSM and PGD variants with perturbation bound $\epsilon=4$. For brevity, the name of \textsc{Single-x} variants are simplified to the task name only. IND: independent, AS: all-shared.} \label{tab:Taskonomy-resnet1-full-4}
\centering
\scriptsize
\tabcolsep=0.06cm
\begin{tabular}{c|c|ccccccc|ccccccc}
\toprule
\multirow{2}{*}{\begin{tabular}[c]{@{}c@{}}Model \\ Index\end{tabular}} & \multirow{2}{*}{\#Params (M)} & \multicolumn{7}{c|}{FGSM}                                               & \multicolumn{7}{c}{PGD}                                                     \\ \cline{3-16} 
                                                                             &                               & Segm           & Norm  & Dept  & Keyp  & Edge  & Total & \pjn{}          & Segm   & Norm  & Dept   & Keyp  & Edge  & Total           & \pjn{}           \\ \hline
IND  & 106.38 & 66.78          & 7.88  & 41.82 & 3.91  & 3.76  & 73.20 & \textbf{79.15} & 153.94          & 16.57 & 94.23  & 5.47  & 4.40  & 178.51 & \textbf{179.32} \\
190  & 105.03 & 53.91          & 11.82 & 34.50 & 5.91  & 6.65  & 56.94 & \textbf{58.89} & 105.63          & 37.71 & 116.13 & 6.74  & 6.36  & 141.97 & \textbf{157.73} \\
348  & 105.02 & 53.04          & 12.33 & 40.33 & 6.35  & 5.90  & 59.07 & \textbf{64.92} & 115.81          & 38.49 & 120.48 & 9.71  & 6.22  & 153.14 & \textbf{166.49} \\
200  & 104.81 & 53.40          & 11.13 & 36.40 & 5.03  & 6.38  & 56.96 & \textbf{61.64} & 104.28          & 36.37 & 113.03 & 5.74  & 7.49  & 140.94 & \textbf{155.51} \\
352  & 104.80 & 51.95          & 11.90 & 35.91 & 6.08  & 5.80  & 55.51 & \textbf{59.68} & 102.05          & 36.28 & 116.72 & 8.21  & 5.88  & 138.80 & \textbf{156.63} \\
469  & 104.57 & 53.07          & 12.95 & 40.12 & 9.45  & 7.10  & 58.58 & \textbf{64.07} & 105.87          & 37.13 & 122.31 & 16.36 & 9.68  & 144.30 & \textbf{160.39} \\
358  & 103.68 & 53.60          & 12.46 & 36.65 & 6.14  & 5.72  & 56.95 & \textbf{60.86} & 107.96          & 38.20 & 123.47 & 8.41  & 6.16  & 148.42 & \textbf{164.93} \\
481  & 103.46 & 57.14          & 13.82 & 37.24 & 11.58 & 7.93  & 60.54 & \textbf{65.79} & 121.96          & 36.53 & 125.33 & 21.67 & 14.79 & 158.55 & \textbf{175.44} \\
191  & 98.21  & 51.06          & 11.31 & 33.85 & 4.79  & 5.30  & 54.51 & \textbf{56.15} & 101.20          & 41.67 & 121.75 & 5.97  & 5.29  & 136.27 & \textbf{151.80} \\
959  & 96.86  & 51.77          & 12.40 & 37.55 & 8.17  & 8.17  & 59.28 & \textbf{63.09} & 113.43          & 37.06 & 127.30 & 5.54  & 7.52  & 149.44 & \textbf{170.95} \\
958  & 83.75  & 56.76          & 11.56 & 38.07 & 8.34  & 6.16  & 63.94 & \textbf{66.90} & 116.17          & 37.81 & 132.04 & 6.21  & 5.85  & 153.90 & \textbf{177.06} \\
1020 & 83.53  & 57.77          & 11.82 & 37.90 & 8.13  & 6.36  & 64.77 & \textbf{68.43} & 126.77          & 36.15 & 115.46 & 6.72  & 7.59  & 158.30 & \textbf{175.65} \\
1043 & 75.59  & 59.98          & 11.12 & 40.90 & 8.55  & 12.49 & 67.54 & \textbf{73.56} & 119.75          & 32.93 & 116.46 & 8.93  & 29.73 & 155.25 & \textbf{170.59} \\
1037 & 62.48  & 55.33          & 12.16 & 36.44 & 8.69  & 8.28  & 61.76 & \textbf{66.87} & 123.41          & 32.09 & 118.24 & 7.56  & 33.09 & 155.66 & \textbf{172.77} \\
AS   & 21.28  & \textbf{59.65} & 26.28 & 36.46 & 8.89  & 8.81  & 59.55 & 59.14          & \textbf{139.16} & 46.94 & 69.82  & 7.34  & 10.84 & 138.78 & 124.39         \\ \bottomrule
\end{tabular}
\end{table}

\begin{table}[htb]
\caption{ARP of all 15 multi-task models similar as Table \ref{tab:Taskonomy-resnet1-full-4}. The attack methods are APGD and Auto-SAGA variants.} \label{tab:Taskonomy-resnet2-full-4}
\centering
\scriptsize
\tabcolsep=0.05cm
\begin{tabular}{c|c|ccccccc|cc}
\toprule
\multirow{2}{*}{Model Index} & \multirow{2}{*}{\#Params (M)} & \multicolumn{7}{c|}{APGD}                                                      & \multicolumn{2}{c}{Auto-SAGA}     \\ \cline{3-11} 
                                  &                               & Segm            & Norm  & Dept   & Keyp  & Edge  & Total           & \pjn{}           & Baseline        & \pjn{}           \\ \hline
IND  & 106.38 & 150.34          & 16.32 & 93.49  & 5.40  & 4.33  & 175.00 & \textbf{175.98} & 178.51          & \textbf{179.32} \\
190  & 105.03 & 98.82           & 33.70 & 109.26 & 6.39  & 6.12  & 132.72 & \textbf{146.86} & 134.60          & \textbf{147.75} \\
348  & 105.02 & 107.45          & 34.00 & 112.95 & 9.31  & 5.93  & 142.54 & \textbf{154.78} & 144.55          & \textbf{155.85} \\
200  & 104.81 & 97.78           & 32.08 & 105.60 & 5.39  & 7.20  & 131.47 & \textbf{145.08} & 133.21          & \textbf{145.95} \\
352  & 104.80 & 95.66           & 32.25 & 109.25 & 7.75  & 5.62  & 129.44 & \textbf{145.90} & 131.41          & \textbf{147.07} \\
469  & 104.57 & 98.82           & 33.00 & 113.63 & 15.70 & 9.14  & 134.73 & \textbf{149.15} & 136.58          & \textbf{150.36} \\
358  & 103.68 & 100.97          & 34.09 & 115.14 & 7.88  & 5.74  & 138.28 & \textbf{153.40} & 140.26          & \textbf{154.84} \\
481  & 103.46 & 113.34          & 32.11 & 116.28 & 20.99 & 12.67 & 147.42 & \textbf{162.81} & 149.73          & \textbf{164.18} \\
191  & 98.21  & 94.75           & 37.13 & 113.48 & 5.59  & 5.13  & 127.89 & \textbf{141.74} & 129.68          & \textbf{142.58} \\
959  & 96.86  & 105.62          & 33.55 & 118.21 & 5.25  & 7.28  & 139.53 & \textbf{158.45} & 141.08          & \textbf{159.39} \\
958  & 83.75  & 108.44          & 33.77 & 122.67 & 5.97  & 5.76  & 143.50 & \textbf{163.92} & 145.27          & \textbf{164.72} \\
1020 & 83.53  & 117.91          & 31.86 & 107.94 & 6.30  & 7.10  & 147.29 & \textbf{162.43} & 149.50          & \textbf{163.90} \\
1043 & 75.59  & 112.46          & 29.12 & 109.43 & 8.45  & 27.97 & 145.09 & \textbf{159.55} & 147.15          & \textbf{160.82} \\
1037 & 62.48  & 115.08          & 28.25 & 110.43 & 7.28  & 29.00 & 145.04 & \textbf{160.81} & 151.71          & \textbf{165.31} \\
AS   & 21.28  & \textbf{136.60} & 46.05 & 69.10  & 7.37  & 10.73 & 135.97 & 122.22          & \textbf{138.78} & 124.39           \\ \bottomrule
\end{tabular}
\end{table}

\begin{table}[htb]
\caption{\textsc{WeightTotal} \textit{VS} \pjn{}: ARP of all 25 multi-task models with diverse sharing patterns trained on NYUv2 and attacked by PGD and APGD variants with perturbation bound $\epsilon=8$. The optimal task-specific weights of \textsc{WeightTotal} when generating adversarial samples are reported in the format of \textit{Segm:Norm:Dept}. IND: independent, AS: all-shared.} \label{tab:NYUv2-resnet-WeightTotal}
\centering
\scriptsize
\tabcolsep=0.05cm
\begin{tabular}{c|ccc|ccc}
\toprule
\multirow{3}{*}{Model Index} & \multicolumn{3}{c|}{PGD}                                             & \multicolumn{3}{c}{APGD}                                                      \\ \cline{2-7} 
                       & \multicolumn{2}{c|}{WeightTotal}             & \multirow{2}{*}{\pjn{}} & \multicolumn{2}{c|}{WeightTotal}                       & \multirow{2}{*}{\pjn{}} \\ \cline{2-3} \cline{5-6}
                       & Optimal Weights & \multicolumn{1}{c|}{ARP}   &                       & Optimal Weights & \multicolumn{1}{c|}{ARP}             &                       \\ \hline
IND                   & 0.1:0.1:0.8     & \multicolumn{1}{c|}{61.04} & \textbf{73.41}        & 0.1:0.1:0.8     & \multicolumn{1}{c|}{72.48}           & \textbf{87.58}        \\
14                     & 0.1:0.5:0.4     & \multicolumn{1}{c|}{72.06} & \textbf{98.9}         & 0.1:0.5:0.4     & \multicolumn{1}{c|}{89.7}            & \textbf{118.63}       \\
4                      & 0.1:0.1:0.8     & \multicolumn{1}{c|}{66.38} & \textbf{79.05}        & 0.1:0.1:0.8     & \multicolumn{1}{c|}{77.56}           & \textbf{93.73}        \\
9                      & 0.1:0.1:0.8     & \multicolumn{1}{c|}{75.24} & \textbf{92.65}        & 0.1:0.1:0.8     & \multicolumn{1}{c|}{98.47}           & \textbf{108.6}        \\
23                     & 0.1:0.1:0.8     & \multicolumn{1}{c|}{73.93} & \textbf{90.43}        & 0.1:0.1:0.8     & \multicolumn{1}{c|}{93.85}           & \textbf{107.08}       \\
5                      & 0.1:0.1:0.8     & \multicolumn{1}{c|}{68.4}  & \textbf{79.6}         & 0.1:0.1:0.8     & \multicolumn{1}{c|}{83.58}           & \textbf{94.44}        \\
10                     & 0.1:0.1:0.8     & \multicolumn{1}{c|}{74.04} & \textbf{93.55}        & 0.1:0.1:0.8     & \multicolumn{1}{c|}{104.93}          & \textbf{113.78}       \\
28                     & 0.1:0.4:0.5     & \multicolumn{1}{c|}{77.73} & \textbf{107.59}       & 0.1:0.3:0.6     & \multicolumn{1}{c|}{105.34}          & \textbf{130.93}       \\
35                     & 0.1:0.2:0.7     & \multicolumn{1}{c|}{77.91} & \textbf{100.53}       & 0.1:0.1:0.8     & \multicolumn{1}{c|}{112.66}          & \textbf{121.3}        \\
38                     & 0.1:0.1:0.8     & \multicolumn{1}{c|}{71.22} & \textbf{93.76}        & 0.1:0.1:0.8     & \multicolumn{1}{c|}{94.42}           & \textbf{113.04}       \\
41                     & 0.1:0.1:0.8     & \multicolumn{1}{c|}{79.97} & \textbf{103.03}       & 0.1:0.1:0.8     & \multicolumn{1}{c|}{109.95}          & \textbf{122.5}        \\
11                     & 0.1:0.1:0.8     & \multicolumn{1}{c|}{73.5}  & \textbf{89.68}        & 0.1:0.1:0.8     & \multicolumn{1}{c|}{97.93}           & \textbf{106.21}       \\
21                     & 0.1:0.1:0.8     & \multicolumn{1}{c|}{74.19} & \textbf{84.81}        & 0.1:0.1:0.8     & \multicolumn{1}{c|}{90.52}           & \textbf{100.94}       \\
33                     & 0.1:0.1:0.8     & \multicolumn{1}{c|}{74.3}  & \textbf{87.08}        & 0.1:0.1:0.8     & \multicolumn{1}{c|}{99.51}           & \textbf{104.87}       \\
36                     & 0.1:0.1:0.8     & \multicolumn{1}{c|}{72.77} & \textbf{99.72}        & 0.1:0.1:0.8     & \multicolumn{1}{c|}{102.56}          & \textbf{119.98}       \\
39                     & 0.1:0.3:0.6     & \multicolumn{1}{c|}{73.64} & \textbf{96.63}        & 0.1:0.1:0.8     & \multicolumn{1}{c|}{98.17}           & \textbf{115.62}       \\
44                     & 0.1:0.1:0.8     & \multicolumn{1}{c|}{79.1}  & \textbf{100.56}       & 0.1:0.1:0.8     & \multicolumn{1}{c|}{110.91}          & \textbf{120.52}       \\
42                     & 0.1:0.2:0.7     & \multicolumn{1}{c|}{73.34} & \textbf{95.79}        & 0.1:0.1:0.8     & \multicolumn{1}{c|}{100.08}          & \textbf{114.74}       \\
48                     & 0.1:0.1:0.8     & \multicolumn{1}{c|}{72.1}  & \textbf{90.34}        & 0.1:0.1:0.8     & \multicolumn{1}{c|}{98.32}           & \textbf{106.66}       \\
26                     & 0.1:0.1:0.8     & \multicolumn{1}{c|}{74.19} & \textbf{93.88}        & 0.1:0.1:0.8     & \multicolumn{1}{c|}{96.45}           & \textbf{115.79}       \\
17                     & 0.1:0.1:0.8     & \multicolumn{1}{c|}{74.82} & \textbf{84.75}        & 0.1:0.1:0.8     & \multicolumn{1}{c|}{92.67}           & \textbf{98.4}         \\
34                     & 0.1:0.1:0.8     & \multicolumn{1}{c|}{66.07} & \textbf{82.03}        & 0.1:0.1:0.8     & \multicolumn{1}{c|}{88.98}           & \textbf{96.55}        \\
43                     & 0.1:0.1:0.8     & \multicolumn{1}{c|}{77.64} & \textbf{95.97}        & 0.1:0.1:0.8     & \multicolumn{1}{c|}{109.13}          & \textbf{115.1}        \\
49                     & 0.1:0.1:0.8     & \multicolumn{1}{c|}{78.76} & \textbf{106.54}       & 0.1:0.1:0.8     & \multicolumn{1}{c|}{113.27}          & \textbf{119.81}       \\
AS             & 0.1:0.1:0.8     & \multicolumn{1}{c|}{75.88} & \textbf{88.51}        & 0.1:0.1:0.8     & \multicolumn{1}{c|}{\textbf{109.36}} & 108.57                \\ \bottomrule
\end{tabular}
\end{table}

\begin{table}[htb]
\caption{The accuracy of six multi-task models with and without adversarial training on NYUv2. The adversarial samples are generated by five PGD-based adversarial attack methods with $\epsilon = 8$.} \label{tab:adv_clean_PGD_no_ori}
\centering
\scriptsize
\tabcolsep=0.01cm
\begin{tabular}{c|ccc|cccccc|cccccc|c}
\hline
\multirow{3}{*}{Adv. Train} & \multicolumn{3}{c|}{Semantic Seg.}                                                                                                             & \multicolumn{6}{c|}{Surface Normal   Prediction}                                                              & \multicolumn{6}{c|}{Depth Estimation}                                                                                                                                      & \multirow{3}{*}{ARP} \\ \cline{2-16}
                               & \multirow{2}{*}{mIoU $\uparrow$} & \multicolumn{1}{c|}{\multirow{2}{*}{\begin{tabular}[c]{@{}c@{}}Pixel\\ Acc $\uparrow$\end{tabular}}} & \multirow{2}{*}{ARP $t_1 $} & \multicolumn{2}{c|}{Error $\downarrow$}          & \multicolumn{3}{c|}{$\theta$, within $\uparrow$}                & \multirow{2}{*}{ARP $t_2 $} & \multicolumn{2}{c|}{Error $\downarrow$}               & \multicolumn{3}{c|}{$\sigma$, within $\uparrow$}                                                                        & \multirow{2}{*}{ARP $t_3 $} &                            \\ \cline{5-9} \cline{11-15}
                               &                       & \multicolumn{1}{c|}{}                                                                     &                            & Mean  & \multicolumn{1}{c|}{Median} & 11.25$\degree$ & 22.5$\degree$  & \multicolumn{1}{c|}{30$\degree$}    &                            & Abs. & \multicolumn{1}{c|}{Rel.} & 1.25 & $1.25^2$ & \multicolumn{1}{c|}{$1.25^3$} &                            &                            \\ \hline
w/o AT      & 25.88 & \multicolumn{1}{c|}{58.05} & -     & 17.28 & \multicolumn{1}{c|}{15.11} & 36.45 & 71.32 & \multicolumn{1}{c|}{84.91} & -    & 0.55 & 0.21 & 64.61 & 89.95 & \multicolumn{1}{c|}{97.39} & -     & -     \\
Single-Segm & 14.53 & \multicolumn{1}{c|}{46.89} & 31.54 & 17.59 & \multicolumn{1}{c|}{16.18} & 29.37 & 73.14 & \multicolumn{1}{c|}{87.22} & 4.60 & 0.68 & 0.25 & 53.84 & 83.77 & \multicolumn{1}{c|}{95.00} & 13.46 & 16.53 \\
Single-Norm & 17.22 & \multicolumn{1}{c|}{48.22} & 25.20 & 17.96 & \multicolumn{1}{c|}{16.07} & 27.33 & 74.48 & \multicolumn{1}{c|}{87.42} & 5.58 & 0.66 & 0.26 & 56.18 & 84.59 & \multicolumn{1}{c|}{95.16} & 12.30 & 14.36 \\
Single-Dept & 18.10 & \multicolumn{1}{c|}{49.33} & 22.54 & 17.72 & \multicolumn{1}{c|}{16.06} & 30.74 & 72.32 & \multicolumn{1}{c|}{86.66} & 4.20 & 0.66 & 0.28 & 55.82 & 83.80 & \multicolumn{1}{c|}{94.67} & 14.51 & 13.75 \\
Total       & 14.99 & \multicolumn{1}{c|}{47.50} & 30.13 & 17.72 & \multicolumn{1}{c|}{15.99} & 30.26 & 72.98 & \multicolumn{1}{c|}{86.55} & 4.21 & 0.66 & 0.28 & 56.23 & 84.16 & \multicolumn{1}{c|}{94.74} & 13.86 & 16.07 \\
DGBA        & 15.67 & \multicolumn{1}{c|}{47.67} & 28.67 & 17.54 & \multicolumn{1}{c|}{15.93} & 29.93 & 74.11 & \multicolumn{1}{c|}{87.18} & 3.64 & 0.67 & 0.28 & 55.46 & 83.40 & \multicolumn{1}{c|}{94.39} & 15.64 & 15.98         \\ \hline
\end{tabular}
\end{table}

\begin{table}[htb]
\caption{The accuracy of the adversarially trained multi-task models similar to Table \ref{tab:adv_clean_PGD_no_ori}. The underlying adversarial sample generation methods are changed to FGSM variants in adversarial training.} \label{tab:adv_clean_FGSM}
\centering
\scriptsize
\tabcolsep=0.01cm
\begin{tabular}{c|ccc|cccccc|cccccc|c}
\hline
\multirow{3}{*}{Adv. Train} & \multicolumn{3}{c|}{Semantic Seg.}                                                                                                             & \multicolumn{6}{c|}{Surface Normal   Prediction}                                                              & \multicolumn{6}{c|}{Depth Estimation}                                                                                                                                      & \multirow{3}{*}{ARP} \\ \cline{2-16}
                               & \multirow{2}{*}{mIoU $\uparrow$} & \multicolumn{1}{c|}{\multirow{2}{*}{\begin{tabular}[c]{@{}c@{}}Pixel\\ Acc $\uparrow$\end{tabular}}} & \multirow{2}{*}{ARP $t_1 $} & \multicolumn{2}{c|}{Error $\downarrow$}          & \multicolumn{3}{c|}{$\theta$, within $\uparrow$}                & \multirow{2}{*}{ARP $t_2 $} & \multicolumn{2}{c|}{Error $\downarrow$}               & \multicolumn{3}{c|}{$\sigma$, within $\uparrow$}                                                                        & \multirow{2}{*}{ARP $t_3 $} &                            \\ \cline{5-9} \cline{11-15}
                               &                       & \multicolumn{1}{c|}{}                                                                     &                            & Mean  & \multicolumn{1}{c|}{Median} & 11.25$\degree$ & 22.5$\degree$  & \multicolumn{1}{c|}{30$\degree$}    &                            & Abs. & \multicolumn{1}{c|}{Rel.} & 1.25 & $1.25^2$ & \multicolumn{1}{c|}{$1.25^3$} &                            &                            \\ \hline
w/o AT                   & 25.88                 & \multicolumn{1}{c|}{58.05}                                                                & -                          & 17.28 & 15.11                       & 36.45 & 71.32 & \multicolumn{1}{c|}{84.91} & -                          & 0.55     & 0.21                          & 64.61       & 89.95                          & \multicolumn{1}{c|}{97.39}                          & -                          & -                          \\
Single-Segm                    & 22.99                 & \multicolumn{1}{c|}{55.72}                                                                & 7.59                      & 18.11 & 15.49                       & 36.36 & 68.22 & \multicolumn{1}{c|}{81.60} & 3.15                      & 0.59     & 0.25                          & 61.23       & 88.08                          & \multicolumn{1}{c|}{96.51}                          & 5.88                      & 5.54                      \\
Single-Norm                    & 23.22                 & \multicolumn{1}{c|}{56.66}                                                                & 6.34                      & 17.53 & 14.85                       & 38.18 & 70.32 & \multicolumn{1}{c|}{83.15} & 0.32                       & 0.58     & 0.24                          & 61.62       & 88.58                          & \multicolumn{1}{c|}{96.80}                          & 4.74                      & 3.58                      \\
Single-Dept                    & 23.41                 & \multicolumn{1}{c|}{55.53}                                                                & 6.94                      & 17.80 & 15.12                       & 36.90 & 69.86 & \multicolumn{1}{c|}{82.77} & 1.27                      & 0.58     & 0.24                          & 62.16       & 88.70                          & \multicolumn{1}{c|}{96.77}                          & 4.17                      & 4.13                      \\
Total                          & 23.00                 & \multicolumn{1}{c|}{55.42}                                                                & 7.83                      & 17.66 & 14.91                       & 37.83 & 70.06 & \multicolumn{1}{c|}{82.75} & 0.27                      & 0.60     & 0.25                          & 60.54       & 88.03                          & \multicolumn{1}{c|}{96.55}                          & 6.43                      & 4.84                      \\
\pjn{}                           & 23.36                 & \multicolumn{1}{c|}{56.12}                                                                & 6.53                      & 18.03 & 15.27                       & 37.07 & 68.78 & \multicolumn{1}{c|}{81.66} & 2.22                      & 0.59     & 0.25                          & 60.57       & 88.04                          & \multicolumn{1}{c|}{96.54}                          & 6.30                      & 5.02                      \\ \hline
\end{tabular}
\end{table}

\begin{table}[htb]
    \caption{ARP of multi-task models trained with and without FAT in Table \ref{tab:adv_clean_FGSM} and attacked by multi-task attacks including FGSM variants of \pjn{}, \textsc{Single}, and \textsc{Total} with $\epsilon=8$.} \label{tab:adv_ARP_FGSM}
    \scriptsize
    \centering
    \tabcolsep=0.05cm
    \begin{tabular}{c|cccc|c}
    \toprule
    Adv. Train  & Single-Segm & Single-Norm & Single-Dept & Total & DGBA  \\ \midrule
    w/o AT      & 29.50        & 28.31       & 38.91       & 34.71 & \textbf{39.40}  \\
    Single-Segm & 6.49        & 5.03        & 7.20         & 7.95  & \textbf{8.83}  \\
    Single-Norm & 12.56       & 10.22       & 14.67       & 14.97 & \textbf{17.09} \\
    Single-Dept & 10.75       & 8.98        & 11.59       & 12.51 & \textbf{13.91} \\
    Total       & 7.39        & 5.62        & 8.04        & 8.79  & \textbf{9.65}  \\
    \pjn{}        & 8.64        & 6.94        & 10.35       & 10.57 & \textbf{11.91} \\ \bottomrule
    \end{tabular}
\end{table}

\begin{table}[htb]
\caption{The model structures (layouts) of multi-task models for NYUv2 with Deeplab-ResNet34. IND: independent, AS: all-shared.} \label{tab:NYUv2-layouts}
\centering
\scriptsize
\tabcolsep=0.01cm
\begin{tabular}{c|c|c}
\toprule
Model Index & \#Params (M) & Layout          \\ \midrule
IND           & 63.83                 & {[}{[}\{0\}, \{2\}, \{1\}{]}, {[}\{0\}, \{2\}, \{1\}{]}, {[}\{0\}, \{2\}, \{1\}{]}, {[}\{0\}, \{2\},   \{1\}{]}, {[}\{0\}, \{2\}, \{1\}{]}{]} \\
9              & 63.6                  & {[}{[}\{1\}, \{0, 2\}{]}, {[}\{1\}, \{0, 2\}{]}, {[}\{1\}, \{2\}, \{0\}{]}, {[}\{1\}, \{2\},   \{0\}{]}, {[}\{1\}, \{2\}, \{0\}{]}{]}         \\
14             & 63.6                  & {[}{[}\{2\}, \{0, 1\}{]}, {[}\{2\}, \{0, 1\}{]}, {[}\{2\}, \{1\}, \{0\}{]}, {[}\{2\}, \{1\},   \{0\}{]}, {[}\{2\}, \{1\}, \{0\}{]}{]}         \\
4              & 63.6                  & {[}{[}\{1, 2\}, \{0\}{]}, {[}\{1, 2\}, \{0\}{]}, {[}\{0\}, \{2\}, \{1\}{]}, {[}\{0\}, \{2\},   \{1\}{]}, {[}\{0\}, \{2\}, \{1\}{]}{]}         \\
23             & 63.59                 & {[}{[}\{0, 1, 2\}{]}, {[}\{1\}, \{0, 2\}{]}, {[}\{1\}, \{2\}, \{0\}{]}, {[}\{1\}, \{2\}, \{0\}{]},   {[}\{1\}, \{2\}, \{0\}{]}{]}             \\
10             & 62.48                 & {[}{[}\{1\}, \{0, 2\}{]}, {[}\{1\}, \{0, 2\}{]}, {[}\{1\}, \{0, 2\}{]}, {[}\{1\}, \{2\}, \{0\}{]},   {[}\{1\}, \{2\}, \{0\}{]}{]}             \\
5              & 62.48                 & {[}{[}\{1, 2\}, \{0\}{]}, {[}\{1, 2\}, \{0\}{]}, {[}\{1, 2\}, \{0\}{]}, {[}\{0\}, \{2\}, \{1\}{]},   {[}\{0\}, \{2\}, \{1\}{]}{]}             \\
28             & 62.47                 & {[}{[}\{0, 1, 2\}{]}, {[}\{2\}, \{0, 1\}{]}, {[}\{2\}, \{0, 1\}{]}, {[}\{2\}, \{1\}, \{0\}{]},   {[}\{2\}, \{1\}, \{0\}{]}{]}                 \\
38             & 62.25                 & {[}{[}\{0, 1, 2\}{]}, {[}\{0, 1, 2\}{]}, {[}\{2\}, \{0, 1\}{]}, {[}\{2\}, \{1\}, \{0\}{]},   {[}\{2\}, \{1\}, \{0\}{]}{]}                     \\
41             & 61.13                 & {[}{[}\{0, 1, 2\}{]}, {[}\{0, 1, 2\}{]}, {[}\{0, 1, 2\}{]}, {[}\{0\}, \{2\}, \{1\}{]}, {[}\{0\},   \{2\}, \{1\}{]}{]}                         \\
33             & 55.43                 & {[}{[}\{0, 1, 2\}{]}, {[}\{0, 1, 2\}{]}, {[}\{1, 2\}, \{0\}{]}, {[}\{1, 2\}, \{0\}{]}, {[}\{0\},   \{2\}, \{1\}{]}{]}                         \\
39             & 55.43                 & {[}{[}\{0, 1, 2\}{]}, {[}\{0, 1, 2\}{]}, {[}\{2\}, \{0, 1\}{]}, {[}\{2\}, \{0, 1\}{]}, {[}\{2\},   \{1\}, \{0\}{]}{]}                         \\
42             & 54.32                 & {[}{[}\{0, 1, 2\}{]}, {[}\{0, 1, 2\}{]}, {[}\{0, 1, 2\}{]}, {[}\{1, 2\}, \{0\}{]}, {[}\{0\},   \{2\}, \{1\}{]}{]}                             \\
44             & 54.32                 & {[}{[}\{0, 1, 2\}{]}, {[}\{0, 1, 2\}{]}, {[}\{0, 1, 2\}{]}, {[}\{1\}, \{0, 2\}{]}, {[}\{1\},   \{2\}, \{0\}{]}{]}                             \\
48             & 47.5                  & {[}{[}\{0, 1, 2\}{]}, {[}\{0, 1, 2\}{]}, {[}\{0, 1, 2\}{]}, {[}\{0, 1, 2\}{]}, {[}\{0\},   \{2\}, \{1\}{]}{]}                                 \\
26             & 42.54                 & {[}{[}\{0, 1, 2\}{]}, {[}\{2\}, \{0, 1\}{]}, {[}\{2\}, \{0, 1\}{]}, {[}\{2\}, \{0, 1\}{]},   {[}\{2\}, \{0, 1\}{]}{]}                         \\
17             & 42.54                 & {[}{[}\{0, 1, 2\}{]}, {[}\{1, 2\}, \{0\}{]}, {[}\{1, 2\}, \{0\}{]}, {[}\{1, 2\}, \{0\}{]},   {[}\{1, 2\}, \{0\}{]}{]}                         \\
34             & 42.32                 & {[}{[}\{0, 1, 2\}{]}, {[}\{0, 1, 2\}{]}, {[}\{1\}, \{0, 2\}{]}, {[}\{1\}, \{0, 2\}{]}, {[}\{1\},   \{0, 2\}{]}{]}                             \\
43             & 41.21                 & {[}{[}\{0, 1, 2\}{]}, {[}\{0, 1, 2\}{]}, {[}\{0, 1, 2\}{]}, {[}\{1\}, \{0, 2\}{]}, {[}\{1\},   \{0, 2\}{]}{]}                                 \\
49             & 34.39                 & {[}{[}\{0, 1, 2\}{]}, {[}\{0, 1, 2\}{]}, {[}\{0, 1, 2\}{]}, {[}\{0, 1, 2\}{]}, {[}\{1\}, \{0,   2\}{]}{]}                                     \\
AS    & 21.28                 & {[}{[}\{0, 1, 2\}{]}, {[}\{0, 1, 2\}{]}, {[}\{0, 1, 2\}{]}, {[}\{0, 1, 2\}{]}, {[}\{0, 1,   2\}{]}{]}  \\ \bottomrule
\end{tabular}
\end{table}

\begin{sidewaystable}[htb]
\caption{The model structures (layouts) of multi-task models for Tiny-Taskonomy with Deeplab-ResNet34. IND: independent, AS: all-shared.} \label{tab:Taskonomy-layouts}
\centering
\scriptsize  
\tabcolsep=0.01cm
\begin{tabular}{c|c|c}
\toprule
\multicolumn{1}{c|}{\begin{tabular}[c]{@{}c@{}}Model\\ Index\end{tabular}}   & \multicolumn{1}{c|}{\begin{tabular}[c]{@{}c@{}}\#Params\\ (M)\end{tabular}}  & Layout         \\ \midrule
IND           & 106.38                & {[}{[}\{0\}, \{1\}, \{2\}, \{4\}, \{3\}{]}, {[}\{0\}, \{1\}, \{2\}, \{4\}, \{3\}{]}, {[}\{0\},   \{1\}, \{2\}, \{4\}, \{3\}{]}, {[}\{0\}, \{1\}, \{2\}, \{4\}, \{3\}{]}, {[}\{0\}, \{1\}, \{2\}, \{4\}, \{3\}{]}{]} \\
190            & 105.03                & {[}{[}\{0\}, \{3\}, \{1, 2, 4\}{]}, {[}\{0\}, \{3\}, \{4\}, \{1, 2\}{]}, {[}\{0\}, \{3\},   \{4\}, \{1, 2\}{]}, {[}\{0\}, \{3\}, \{4\}, \{2\}, \{1\}{]}, {[}\{0\}, \{3\}, \{4\}, \{2\}, \{1\}{]}{]}                 \\
348            & 105.02                & {[}{[}\{1, 2, 3, 4\}, \{0\}{]}, {[}\{0\}, \{1, 2\}, \{4\}, \{3\}{]}, {[}\{0\}, \{1, 2\},   \{4\}, \{3\}{]}, {[}\{0\}, \{4\}, \{3\}, \{2\}, \{1\}{]}, {[}\{0\}, \{4\}, \{3\}, \{2\}, \{1\}{]}{]}                     \\
200            & 104.81                & {[}{[}\{0\}, \{3\}, \{1, 2, 4\}{]}, {[}\{0\}, \{3\}, \{1, 2, 4\}{]}, {[}\{0\}, \{3\}, \{4\},   \{1, 2\}{]}, {[}\{0\}, \{3\}, \{4\}, \{2\}, \{1\}{]}, {[}\{0\}, \{3\}, \{4\}, \{2\}, \{1\}{]}{]}                     \\
469            & 104.57                & {[}{[}\{1, 2, 3, 4\}, \{0\}{]}, {[}\{1, 2, 3, 4\}, \{0\}{]}, {[}\{0\}, \{1, 2\}, \{4\},   \{3\}{]}, {[}\{0\}, \{4\}, \{3\}, \{2\}, \{1\}{]}, {[}\{0\}, \{4\}, \{3\}, \{2\}, \{1\}{]}{]}                             \\
358            & 103.68                & {[}{[}\{1, 2, 3, 4\}, \{0\}{]}, {[}\{0\}, \{3, 4\}, \{1, 2\}{]}, {[}\{0\}, \{3, 4\}, \{1,   2\}{]}, {[}\{0\}, \{4\}, \{3\}, \{2\}, \{1\}{]}, {[}\{0\}, \{4\}, \{3\}, \{2\}, \{1\}{]}{]}                             \\
481            & 103.46                & {[}{[}\{1, 2, 3, 4\}, \{0\}{]}, {[}\{1, 2, 3, 4\}, \{0\}{]}, {[}\{0\}, \{3\}, \{1, 2,   4\}{]}, {[}\{0\}, \{3\}, \{1\}, \{4\}, \{2\}{]}, {[}\{0\}, \{3\}, \{1\}, \{4\}, \{2\}{]}{]}                                 \\
191            & 98.21                 & {[}{[}\{0\}, \{3\}, \{1, 2, 4\}{]}, {[}\{0\}, \{3\}, \{4\}, \{1, 2\}{]}, {[}\{0\}, \{3\},   \{4\}, \{1, 2\}{]}, {[}\{0\}, \{3\}, \{4\}, \{1, 2\}{]}, {[}\{0\}, \{3\}, \{4\}, \{2\}, \{1\}{]}{]}                     \\
959            & 96.86                 & {[}{[}\{1, 2, 4\}, \{0, 3\}{]}, {[}\{0, 3\}, \{4\}, \{1, 2\}{]}, {[}\{0, 3\}, \{4\}, \{1,   2\}{]}, {[}\{0, 3\}, \{4\}, \{2\}, \{1\}{]}, {[}\{4\}, \{2\}, \{1\}, \{3\}, \{0\}{]}{]}                                 \\
958            & 83.75                 & {[}{[}\{1, 2, 4\}, \{0, 3\}{]}, {[}\{0, 3\}, \{4\}, \{1, 2\}{]}, {[}\{0, 3\}, \{4\}, \{1,   2\}{]}, {[}\{0, 3\}, \{4\}, \{2\}, \{1\}{]}, {[}\{0, 3\}, \{4\}, \{2\}, \{1\}{]}{]}                                     \\
1020           & 83.53                 & {[}{[}\{1, 2, 4\}, \{0, 3\}{]}, {[}\{1, 2, 4\}, \{0, 3\}{]}, {[}\{0, 3\}, \{4\}, \{1,   2\}{]}, {[}\{0, 3\}, \{4\}, \{2\}, \{1\}{]}, {[}\{0, 3\}, \{4\}, \{2\}, \{1\}{]}{]}                                         \\
1043           & 75.59                 & {[}{[}\{1, 2, 4\}, \{0, 3\}{]}, {[}\{1, 2, 4\}, \{0, 3\}{]}, {[}\{1, 2, 4\}, \{0,   3\}{]}, {[}\{0, 3\}, \{2, 4\}, \{1\}{]}, {[}\{2, 4\}, \{1\}, \{3\}, \{0\}{]}{]}                                                 \\
1037           & 62.48                 & {[}{[}\{1, 2, 4\}, \{0, 3\}{]}, {[}\{1, 2, 4\}, \{0, 3\}{]}, {[}\{1, 2, 4\}, \{0,   3\}{]}, {[}\{0, 3\}, \{2, 4\}, \{1\}{]}, {[}\{0, 3\}, \{2, 4\}, \{1\}{]}{]}                                                     \\
AS    & 21.28                 & {[}{[}\{0, 1, 2, 3, 4\}{]}, {[}\{0, 1, 2, 3, 4\}{]}, {[}\{0, 1, 2, 3, 4\}{]}, {[}\{0,   1, 2, 3, 4\}{]}, {[}\{0, 1, 2, 3, 4\}{]}{]}     \\ \bottomrule                                                                           
\end{tabular}
\end{sidewaystable}

\clearpage
{\small
\bibliographystyle{elsarticle-num}
\bibliography{reference}
}

\end{document}